# The History Began from AlexNet: A Comprehensive Survey on Deep Learning Approaches

Md Zahangir Alom[1], Tarek M. Taha[1], Chris Yakopcic[1], Stefan Westberg[1], Paheding Sidike[2], Mst Shamima Nasrin[1], Brian C Van Essen[3], Abdul A S. Awwal[3], and Vijayan K. Asari[1]

*Abstract*—In recent years, deep learning has garnered tremendous success in a variety of application domains. This new field of machine learning has been growing rapidly, and has been applied to most traditional application domains, as well as some new areas that present more opportunities. Different methods have been proposed based on different categories of learning, including supervised, semi-supervised, and un-supervised learning. Experimental results show state-of-the-art performance using deep learning when compared to traditional machine learning approaches in the fields of image processing, computer vision, speech recognition, machine translation, art, medical imaging, medical information processing, robotics and control, bio-informatics, natural language processing (NLP), cybersecurity, and many others.

This report presents a brief survey on the advances that have occurred in the area of DL, starting with the Deep Neural Network (DNN). The survey goes on to cover the Convolutional Neural Network (CNN), the Recurrent Neural Network (RNN) including Long Short Term Memory (LSTM) and Gated Recurrent Units (GRU), the Auto-Encoder (AE), the Deep Belief Network (DBN), the Generative Adversarial Network (GAN), and Deep Reinforcement Learning (DRL). Additionally, we have included recent developments such as advanced variant DL techniques based on these DL approaches. This work considers most of the papers published after 2012 from when the history of deep learning began. Furthermore, DL approaches that have been explored and evaluated in different application domains are also included in this survey. We also included recently developed frameworks, SDKs, and benchmark datasets that are used for implementing and evaluating deep learning approaches. There are some surveys that have been published on Deep Learning using Neural Networks [1, 38] and a survey on RL [234]. However, those papers have not discussed the individual advanced techniques for training large scale deep learning models and the recently developed method of generative models [1].

*Index Terms*—**Deep Learning, Convolutional Neural Network (CNN), Recurrent Neural Network (RNN), Auto-Encoder (AE), Restricted Boltzmann Machine (RBM), Deep Belief Network (DBN), Generative Adversarial Network (GAN), Deep Reinforcement Learning (DRL), Transfer Learning.**

## I. INTRODUCTION

Since the 1950s, a small subset of Artificial Intelligence (AI), often called Machine Learning (ML), has revolutionized several fields in the last few decades. Neural Networks (NN) are a subfield of ML, and it was this subfield that spawned Deep Learning (DL). Since its inception DL has been creating ever larger disruptions, showing outstanding success in almost every application domain. Fig. 1 shows, the taxonomy of AI. DL (using either deep architecture of learning or hierarchical learning approaches) is a class of ML developed largely from 2006 onward. Learning is a procedure consisting of estimating the model parameters so that the learned model (algorithm) can perform a specific task. For example, in Artificial Neural Networks (ANN), the parameters are the weight matrices ($w_{i,j}'s$). DL on the other hand consists of several layers in between the input and output layer which allows for many stages of non-linear information processing units with hierarchical architectures to be present that are exploited for feature learning and pattern classification [1, 2]. Learning methods based on representations of data can also be defined as representation learning [3]. Recent literature states that DL based representation learning involves a hierarchy of features or concepts, where the high-level concepts can be defined from the low-level ones and low-level concepts can be defined from high-level ones. In some articles DL has been described as a universal learning approach that is able to solve almost all kinds of problems in different application domains. In other words, DL is not task specific [4].

### A. Types of DL approaches:

Like machine learning, deep learning approaches can be categorized as follows: supervised, semi-supervised or partially supervised, and unsupervised. In addition, there is another category of learning called Reinforcement Learning (RL) or Deep RL (DRL) which are often discussed under the

Md Zahangir Alom[1*], Tarek M. Taha[1], Chris Yakopcic[1], Stefan Westberg[1] , Mst Shamima Nasrin[1], and Vijayan K. Asari[1] are with the University of Dayton, 300 College Park, Dayton, OH 45469 USA (e-mail: Emails: {[1*]alomm1, ttaha1, cyakopcic1, westbergs1, nasrinm1, vasari1}@udayton.edu).

Paheding Sidike[2], is with department of Earth and Atmospheric Sciences, Saint Louis University, St. Louis, MO, USA. He is currently working as Post-Doctoral research scientist on deep Learning, computer vision for remote sensing and hyper spectral imaging (e-mail: pehedings@slu.edu).

Brian C Van Esesn[3]and Abdul A S. Awwal[3] are with the Lawrence Livermore National Laboratory (LLNL), Livermore, CA 94550 USA. (e-mail: {vanessen1, awwal1}@llnl.gov).

scope of semi supervised or sometimes under unsupervised learning approaches.

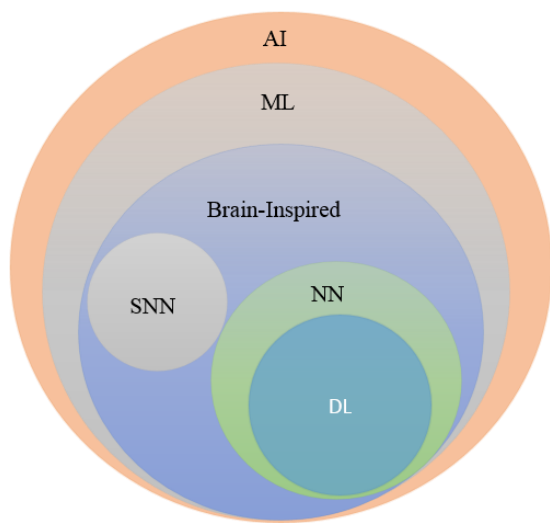

**Fig. 1**. AI: Artificial Intelligence, ML, NN, DL, and Spiking Neural Networks (SNN) according to [294].

*1) Supervised Learning*

Supervised learning is a learning technique that uses labeled data. In the case of supervised DL approaches, the environment has a set of inputs and corresponding outputs $(x_t, y_t) \sim \rho$. For example, if for input $x_t$, the intelligent agent predicts $\hat{y}_t = f(x_t)$, the agent will receive a loss value $l(y_t, \hat{y}_t)$. The agent will then iteratively modify the network parameters for better approximation of the desired outputs. After successful training, the agent will be able to get the correct answers to questions from the environment. There are different supervised learning approaches for deep leaning including Deep Neural Networks (DNN), Convolutional Neural Networks (CNN), Recurrent Neural Networks (RNN) including Long Short Term Memory (LSTM), and Gated Recurrent Units (GRU). These networks are described in Sections 2, 3, 4, and 5 respectively.

*2) Semi-supervised Learning*
Semi-supervised learning is learning that occurs based on partially labeled datasets (often also called reinforcement learning). Section 8 of this study surveys DRL approaches. In some cases, DRL and Generative Adversarial Networks (GAN) are used as semi-supervised learning techniques. Additionally, RNN including LSTM and GRU are used for semi-supervised learning as well. GAN is discussed in Section 7.

*3) Unsupervised learning*

Unsupervised learning systems are ones that can without the presence of data labels. In this case, the agent learns the internal representation or important features to discover unknown relationships or structure within the input data. Often clustering, dimensionality reduction, and generative techniques are considered as unsupervised learning approaches. There are several members of the deep learning family that are good at clustering and non-linear dimensionality reduction, including Auto Encoders (AE), Restricted Boltzmann Machines (RBM), and the recently developed GAN. In addition, RNNs, such as LSTM and RL, are also used for unsupervised learning in many application domains [243]. Sections 6 and 7 discuss RNNs and LSTMs in detail.

*4) Deep Reinforcement Learning (DRL)*

Deep Reinforcement Learning is a learning technique for use in unknown environments. DRL began in 2013 with Google Deep Mind [5, 6]. From then on, several advanced methods have been proposed based on RL. Here is an example of RL: if environment samples inputs: $x_t \sim \rho$, agent predict: $\hat{y}_t = f(x_t)$, agent receive cost: $c_t \sim P(c_t|x_t, \hat{y}_t)$ where $P$ is an unknown probability distribution, the environment asks an agent a question, and gives a noisy score as the answer. Sometimes this approach is called semi-supervised learning as well. There are many semi-supervised and un-supervised techniques that have been implemented based on this concept (in Section 8). In RL, we do not have a straight forward loss function, thus making learning harder compared to traditional supervised approaches. The fundamental differences between RL and supervised learning are: first, you do not have full access to the function you are trying to optimize; you must query them through interaction, and second, you are interacting with a state-based environment: input $x_t$ depends on previous actions.

Depending upon the problem scope or space, you can decide which type of RL needs to be applied for solving a task. If the problem has a lot of parameters to be optimized, DRL is the best way to go. If the problem has fewer parameters for optimization, a derivation free RL approach is good. An example of this is annealing, cross entropy methods, and SPSA. We conclude this section with a quote from Yann LeCun:

**"If intelligence was a cake, unsupervised learning would be the cake, supervised learning would be the icing, and reinforcement learning would be the carry." – Yann LeCun**

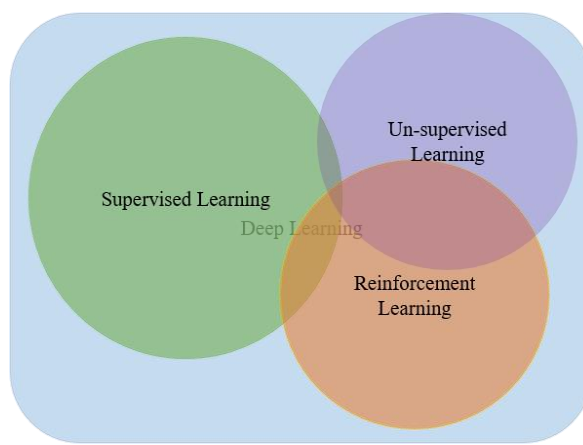

**Fig. 2.** Category of Deep Leaning approaches

## B. Feature Learning

A key difference between traditional ML and DL is in how features are extracted. Traditional ML approaches use handmade features by applying several feature extraction algorithms including Scale Invariant Feature Transform (SIFT), Speeded Up Robust Features (SURF), GIST, RANSAC, Histogram Oriented Gradient (HOG), Local Binary Pattern (LBP), Empirical mode decomposition (EMD) for speech analysis, and many more. Finally, the leaning algorithms including support vector machine (SVM), Random Forest (RF), Principle Component Analysis (PCA), Kernel PCA (KPCA), Linear Decrement Analysis (LDA), Fisher Decrement Analysis (FDA), and many more are applied for classification on the extracted features. Additionally, other boosting approaches are often used where several learning algorithms are applied on the features of a single task or dataset and a decision is made according to the multiple outcomes from the different algorithms.

TABLE I
DIFFERENT FEATURE LEARNING APPROACHES

| Approaches | Learning steps | | | | |
|---|---|---|---|---|---|
| Rule based | Input | Hand-design features | Output | | |
| Traditional Machine Learning | Input | Hand-design features | Mapping from features | Output | |
| Representation Learning | Input | Features | Mapping from features | Output | |
| Deep Learning | Input | Simple features | Complex features | Mapping from features | Output |

On the other hand, in the case of DL, the features are learned automatically and are represented hierarchically in multiple levels. This is the strong point of deep learning against traditional machine learning approaches. The following table shows the different feature-based learning approaches with different learning steps.

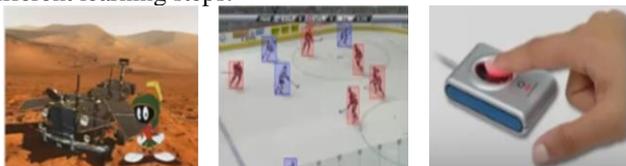

**Fig. 3.** Where to apply DL approaches

## C. When and where to apply DL

DL is employed in several situations where machine intelligence would be useful (see Fig. 3):
1. Absence of a human expert (navigation on Mars)
2. Humans are unable to explain their expertise (speech recognition, vision and language understanding)
3. The solution to the problem changes over time (tracking, weather prediction, preference, stock, price prediction)
4. Solutions need to be adapted to the particular cases (biometrics, personalization).
5. The problem size is too vast for our limited reasoning capabilities (calculation webpage ranks, matching ads to Facebook, sentiment analysis).

At present deep learning is being applied in almost all areas. As a result, this approach is often called a universal learning approach. Some example applications are shown in Fig. 4.

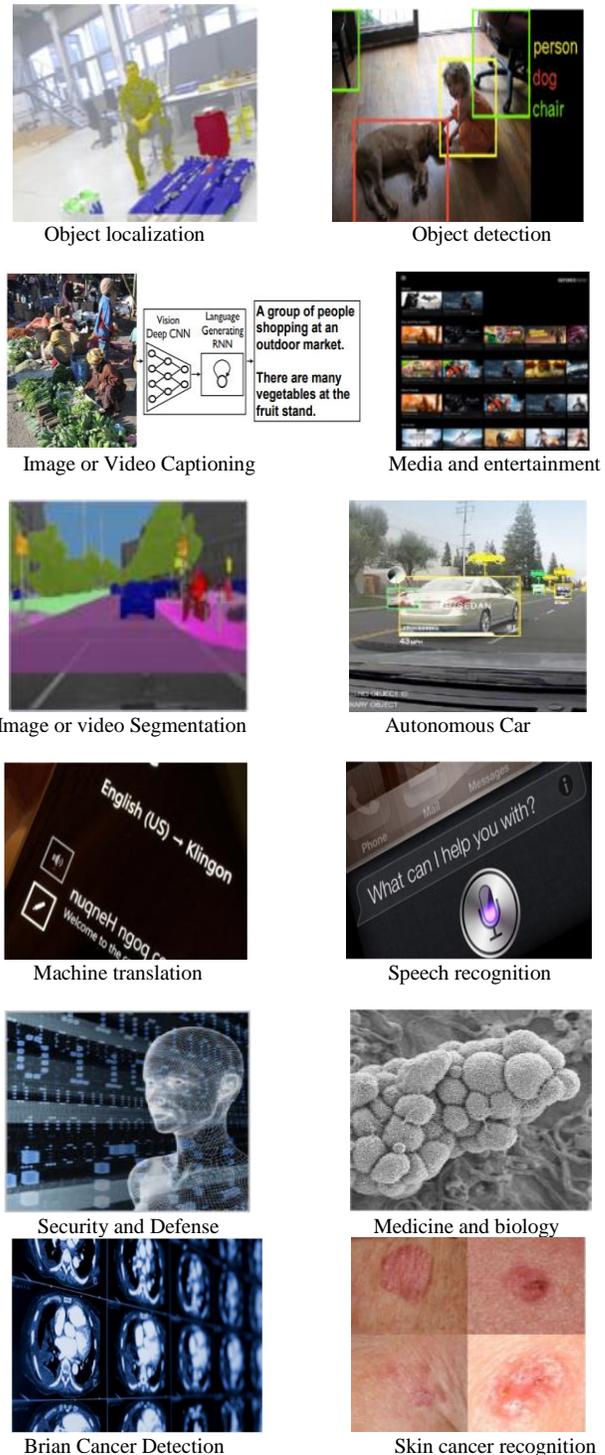

**Fig. 4.** Example images where DL is applied successfully and achieved state-of-the-art performance.

## D. State-of-the-art performance of DL

There are some outstanding successes in the fields of computer vision and speech recognition as discussed below:

*1) Image classification on ImageNet dataset*
One of the large-scale problems is named Large Scale Visual Recognition Challenge (LSVRC). DL CNN based techniques show state-of-the-art accuracy on the ImageNet task [11]. Russakovsky et al. recently published a paper on the ImageNet dataset and the state-of-the-art accuracies achieved during last few years [285]. The following graph shows the success story of deep learning techniques overtime on this challenge from 2012. ResNet-152 shows only 3.57% error, which is better than human error for this task at 5%.

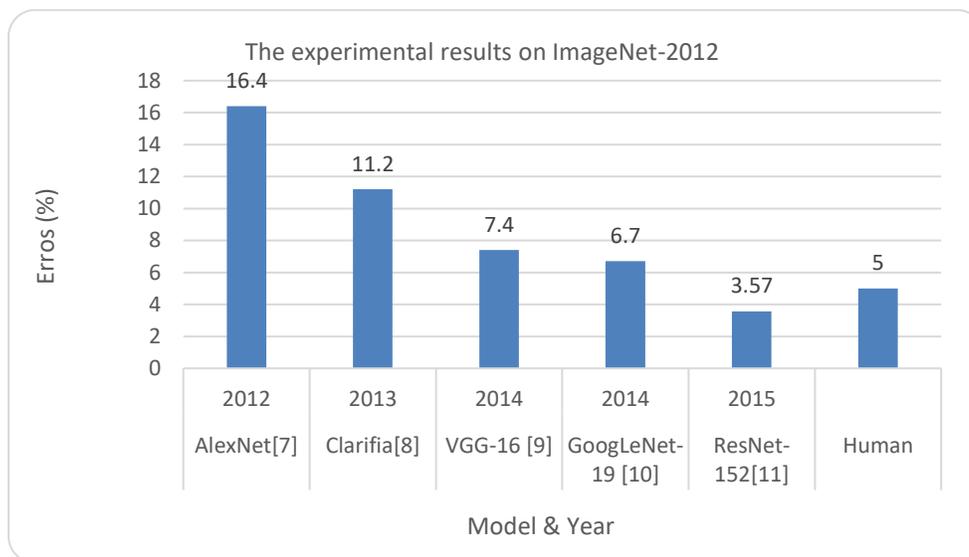

**Fig. 5.** Accuracy for ImageNet challenge with different DL models.

*2) Automatic Speech recognition*
The initial success in the field of speech recognition on the popular TIMIT dataset (common data set are generally used for evaluation) was with small scale recognition tasks. The TIMIT acoustic-Phonetic continuous speech Corpus contains 630 speakers from eight major dialects of American English, where each speaker reads 10 sentences. The graph below summarizes the error rates including these early results and is measured as percent phone error rate (PER) over the last 20 years. The bar graph clearly shows that the recently developed deep learning approaches (top of the graph) perform better compared to any other previous machine learning approaches on the TIMIT dataset.

*E. Why deep Learning*

*1) Universal learning approach*
This approach is sometimes called universal learning because it can be applied to almost any application domain.

*2) Robust*
 Deep learning approaches do not require the design of features ahead of time. Features are automatically learned that are optimal for the task at hand. As a result, the robustness to natural variations in the data is automatically learned.

*3) Generalization*
The same deep learning approach can be used in different applications or with different data types. This approach is often called transfer learning. In addition, this approach is helpful where the problem does not have sufficient available data. There are several papers that have been published based on this concept (discussed in more detail in Section 4).

*4) Scalability*
The deep learning approach is highly scalable. In a 2015 paper, Microsoft described a network known as ResNet [11]. This network contains 1202 layers and is often implemented at a supercomputing scale. There is a big initiative at Lawrence Livermore National Laboratory (LLNL) in developing frameworks for networks like this, which can implement thousands of nodes [24].

*F. Challenges of DL*
There are several challenges for deep learning:
- Big data analytics using Deep Learning
- Scalability of DL approaches
- Ability to generate data which is important where data is not available for learning the system (especially for computer vision task such as inverse graphics).
- Energy efficient techniques for special purpose devices including mobile intelligence, FPGAs, and so on.
- Multi-task and transfer learning (generalization) or multi-module learning. This means learning from different domains or with different models together.
- Dealing with causality in learning.

Most of the mentioned challenges have already been considered seriously by the deep learning community. Several papers have been published as solutions to all of those challenges. For the big data analytics challenge, there is a good survey that was conducted in 2014. In this paper, the authors explain details about how DL can deal with different criteria including volume, velocity, variety, and veracity of the big data problem. The authors also have shown different advantages of DL approaches when dealing with big data problems [25, 26, and 27]. Deep learning is a data driven technique. Fig. 7 clearly demonstrates that the performance of traditional ML approaches shows better performance for lesser amounts of input data. As the amount of data increases beyond a certain amount, the performance of traditional machine learning approaches becomes steady. In contrast, the performance of deep learning approaches increased with respect to the increment in the amount of data.

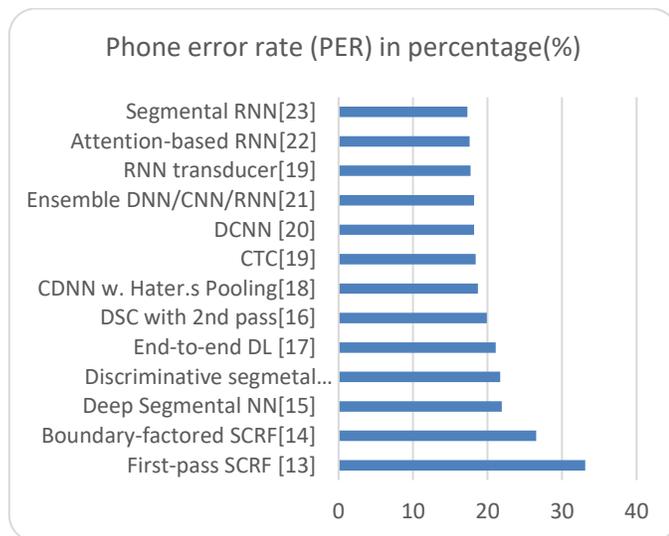

**Fig. 6.** Phone error rate (PER) for TIMIT dataset

Secondly, in most of the cases for solving large scale problems, the solution is being implemented on High Performance Computing (HPC) system (super-computing, cluster, sometime considered cloud computing) which offers immense potential for data-intensive business computing. As data explodes in velocity, variety, veracity and volume, it is getting increasingly difficult to scale compute performance using enterprise class servers and storage in step with the increase. Most of the papers considered all the demands and suggested efficient HPC with heterogeneous computing systems. In one example, Lawrence Livermore National Laboratory (LLNL) has developed a framework which is called Livermore Big Artificial Neural Networks (LBANN) for large-scale implementation (in super-computing scale) for DL which clearly supplants the issue of scalability of DL [24]

Thirdly, generative models are another challenge for deep learning. One examples is the GAN, which is an outstanding approach for data generation for any task which can generate data with the same distribution [28]. Fourthly, multi-task and transfer learning which we have discussed in Section 7. Fourthly, there is a lot of research that has been conducted on energy efficient deep learning approaches with respect to network architectures and hardwires. Section 10 discusses this issue

Can we make any uniform model that can solve multiple tasks in different application domains? As far as the multi-model system is concerned, there has been one paper published recently from Google titled "One Model To Learn Them All" [29]. This approach can learn from different application domains including ImageNet, multiple translation tasks, Image captioning (MS-COCO dataset), speech recognition corpus and English parsing task. We will be discussing most of the challenges and respective solutions through this survey. There are some other multi-task techniques that have been proposed in the last few years [30, 31, and 32]

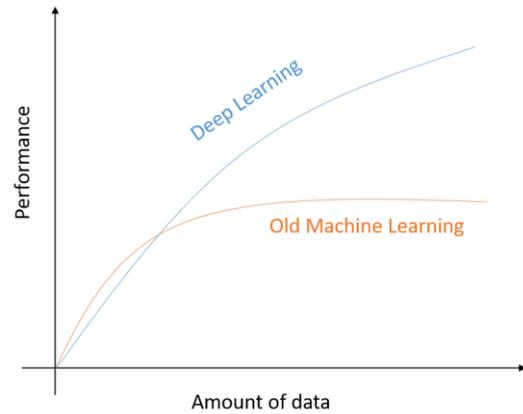

**Fig. 7.** The performance of deep learning with respect to the number of data.

Finally, a learning system with causality has been presented, which is a graphical model that defines how one may infer a causal model from data. Recently a DL based approach has been proposed for solving this type of problem [33]. However, there are other many challenging issues have been solved in the last few years which were not possible to solve efficiently before this revolution. For example: image or video captioning [34], style transferring from one domain to anther domain using GAN [35], text to image synthesis [36], and many more [37].
There are some surveys that have been conducted recently in this field [294,295]. These papers survey on deep learning and its revolution, but this they did not address the recently developed generative model called GAN [28]. In addition, they discuss little about RL and did not cover recent trends of DRL approaches [1, 39]. In most of the cases, the surveys that have been conducted are on different DL approaches individually. There is good survey which is based on Reinforcement Learning approaches [40, 41]. Another survey exists on transfer learning [42]. One surveys has been conducted on neural network hardware [43]. However, the main objective of this work is to provide an overall idea on deep learning and its related fields including deep supervised (e.g. DNN, CNN, and RNN), unsupervised (e.g. AE, RBM, GAN) (sometimes GAN also used for semi-supervised learning tasks) and DRL. In some cases, DRL is considered to be a semi-supervised or an un-supervised approach. In addition, we have considered the recently developing trends of this field and applications which are developed based on these techniques. Furthermore, we have included the framework and benchmark datasets which are often used for evaluating deep learning techniques. Moreover, the name of the conferences and journals are also included which are considered by this community for publishing their research articles.
The rest of the paper has been organized in the following ways: the detailed surveys of DNNs are discussed in Section II, Section III discusses on CNNs. Section IV describes different advanced techniques for efficient training of DL approaches. Section V. discusses on RNNs. AEs and RBMs are discussed in Section VI. GANs with applications are discussed in Section VII. RL is presented in the Section VIII. Section IX explains transfer learning. Section X. presents energy efficient approaches and hardwires for DL. The section XI discusses on deep learning frameworks and standard development kits (SDK). The benchmarks for different application domains with web links are given in Section XII. The conclusions are made in Section XIII.

## II. DEEP NEURAL NETWORK (DNN)

### A. The History of DNN

Below is a brief history of neural networks highlighting key events:

- 1943: McCulloch & Pitts show that neurons can be combined to construct a Turing machine (using ANDs, ORs, & NOTs) [44].
- 1958: Rosenblatt shows that perceptron's will converge if what they are trying to learn can be represented [45].
- 1969: Minsky & Papert show the limitations of perceptron's, killing research in neural networks for a decade [46].
- 1985: The backpropagation algorithm by GeoffreyHinton et al [47] revitalizes the field.
- 1988: Neocognitron: a hierarchical neural network capable of visual pattern recognition [48].
- 1998: CNNs with Backpropagation for document analysis by Yan LeCun [49].
- 2006: The Hinton lab solves the training problem for DNNs [50,51].
- 2012 : AlexNet by Alex Krizhevesky in 2012 [7].

**Fig. 8.** History of DL

Computational neurobiology has conducted significant research on constructing computational models of artificial neurons. Artificial neurons, which try to mimic the behavior of the human brain, are the fundamental component for building ANNs. The basic computational element (neuron) is called a node (or unit) which receives inputs from external sources, and has some internal parameters (including weights and biases that are learned during training) which produce outputs. This unit is called a perceptron. The basic block diagram of a perceptron for NNs is shown in the following diagram.

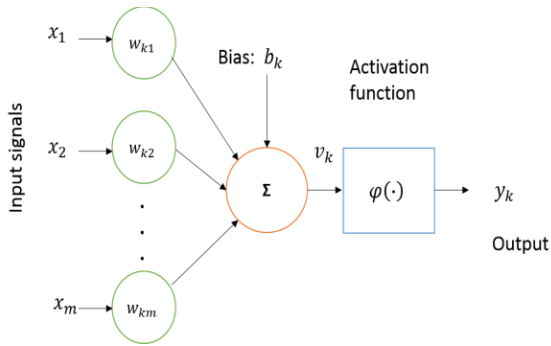

**Fig. 9.** Basic model of a neuron

Fig. 9 shows the basic nonlinear model of a neuron, where $x_1, x_2, x_3, \cdots x_m$ are input signals; $w_{k1}, w_{k2}, w_{k3}, \cdots w_{km}$ are synaptic weights; $v_k$ is the linear combination of input signals; $\varphi(\cdot)$ is the activation function (such as sigmoid), and $y_k$ is the output. The bias $b_k$ is added with a linear combiner of outputs $v_k$, which has the effect of applying an affine transformation, producing the outputs $y_k$. The neuron functionality can be represented mathematically as follows:

$$v_k = \sum_{j=1}^{m} w_{kj} x_j \quad (1)$$

$$y_k = \varphi(v_k + b_k) \quad (2)$$

ANNs or general NNs consist of Multilayer Perceptron's (MLP) which contain one or more hidden layers with multiple hidden units (neurons) in them. The NN model with MLP is shown in Fig. 10.

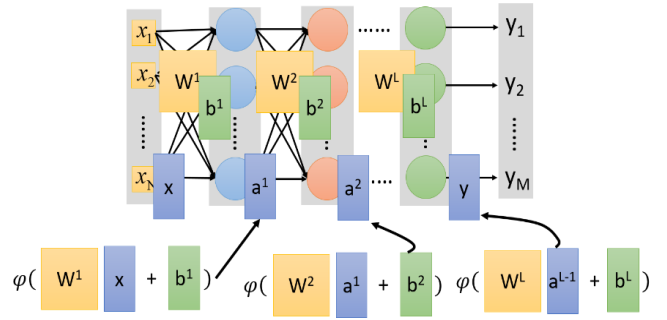

**Fig. 10.** Neural network model with multiple layers perceptron

The multilayer perceptron can be expressed mathematically (which is a composite function) as follows:

$$y = f(x) = \varphi(w^L \cdots \varphi(w^2 \varphi(w^1 x + b^1) + b^2) \cdots + b^L) \quad (3)$$

### B. Gradient descent

The gradient descent approach is a first order optimization algorithm which is used for finding the local minima of an objective function. This has been used for training ANNs in the last couple of decades successfully. Algorithm I explains the concept of gradient descent:

---
**Algorithm I.** Gradient descent
**Inputs:** loss function $\varepsilon$, learning rate $\eta$, dataset $X, y$ and the model $\mathcal{F}(\theta, x)$
**Outputs:** Optimum $\theta$ which minimizes $\varepsilon$
**REPEAT** until converge:
$$\tilde{y} = \mathcal{F}(\theta, x)$$
$$\theta = \theta - \eta \cdot \frac{1}{N} \sum_{i=1}^{N} \frac{\partial \varepsilon(y, \tilde{y})}{\partial \theta}$$
**End**
---

## C. Stochastic Gradient Descent (SGD)

Since a long training time is the main drawback for the traditional gradient descent approach, the SGD approach is used for training Deep Neural Networks (DNN) [52]. Algorithm II explains SGD in detail.

---
**Algorithm II. Stochastic** Gradient Descent (SGD)

**Inputs:** loss function $\varepsilon$, learning rate $\eta$, dataset $X, y$ and the model $\mathcal{F}(\theta, x)$

**Outputs:** Optimum $\theta$ which minimizes $\varepsilon$

**REPEAT** until converge:

    Shuffle $X, y$;

    **For** each batch of $x_i, y_i$ in $X, y$ **do**

    $\tilde{y}_i = \mathcal{F}(\theta, x_i)$;

    $\theta = \theta - \eta \cdot \frac{1}{N} \sum_{i=1}^{N} \frac{\partial \varepsilon(y_i, \tilde{y}_i)}{\partial \theta}$

**End**

---

## D. Back-propagation

DNN are trained with the popular Back-Propagation (BP) algorithm with SGD [53]. The pseudo code of the basic Back-propagation is given in Algorithm III. In the case of MLPs, we can easily represent NN models using computation graphs which are directive acyclic graphs. For that representation of DL, we can use the chain-rule to efficiently calculate the gradient from the top to the bottom layers with BP as shown in Algorithm III for a single path network. For example:

$$y = f(x) = \varphi(w^L \cdots \varphi(w^2 \varphi(w^1 x + b^1) + b^2) \cdots + b^L) \quad (4)$$

This is composite function for $L$ layers of a network. In case of $L = 2$, then the function can be written as

$$y = f(x) = f(g(x)) \quad (5)$$

According to the chain rule, the derivative of this function can be written as

$$\frac{\partial y}{\partial x} = \frac{\partial f(x)}{\partial x} = f'(g(x)) \cdot g'(x) \quad (6)$$

## E. Momentum

Momentum is a method which helps to accelerate the training process with the SGD approach. The main idea behind it is to use the moving average of the gradient instead of using only the current real value of the gradient. We can express this with the following equation mathematically:

$$v_t = \gamma \, v_{t-1} - \eta \, \nabla \mathcal{F}(\theta_{t-1}) \quad (7)$$
$$\theta_t = \theta_{t-1} + v_t \quad (8)$$

Here $\gamma$ is the momentum and $\eta$ is the learning rate for the $t^{th}$ round of training. Other popular approaches have been introduced during last few years which are explained in section IX under the scope of optimization approaches. The main advantage of using momentum during training is to prevent the network from getting stuck in local minimum. The values of momentum are $\gamma \in (0,1]$. It is noted that a higher momentum value overshoots its minimum, possibly making the network unstable. In general, $\gamma$ is set to 0.5 until the initial learning stabilizes and is then increased to 0.9 or higher [54].

---
**Algorithm III.** Back-propagation

**Input:** A network with $l$ layers, the activation function $\sigma_l$, the outputs of hidden layer $h_l = \sigma_l(W_l^T h_{l-1} + b_l)$ and the network output $\tilde{y} = h_l$

Compute the gradient: $\delta \leftarrow \frac{\partial \varepsilon(y_i, \tilde{y}_i)}{\partial y}$

**For** $i \leftarrow l$ to 0 **do**
    Calculate gradient for present layer:
$$\frac{\partial \varepsilon(y, \tilde{y})}{\partial W_l} = \frac{\partial \varepsilon(y, \tilde{y})}{\partial h_l} \frac{\partial h_l}{\partial W_l} = \delta \frac{\partial h_l}{\partial W_l}$$
$$\frac{\partial \varepsilon(y, \tilde{y})}{\partial b_l} = \frac{\partial \varepsilon(y, \tilde{y})}{\partial h_l} \frac{\partial h_l}{\partial b_l} = \delta \frac{\partial h_l}{\partial b_l}$$
    Apply gradient descent using $\frac{\partial \varepsilon(y, \tilde{y})}{\partial W_l}$ and $\frac{\partial \varepsilon(y, \tilde{y})}{\partial b_l}$
    Back-propagate gradient to the lower layer
$$\delta \leftarrow \frac{\partial \varepsilon(y, \tilde{y})}{\partial h_l} \frac{\partial h_l}{\partial h_{l-1}} = \delta \frac{\partial h_l}{\partial h_{l-1}}$$
**End**

---

## F. Learning rate ($\eta$)

The learning rate is an important component for training DNN (as explained in Algorithm I and II). The learning rate is the step size considered during training which makes the training process faster. However, selecting the value of the learning rate is sensitive. For example: if you choose a larger value for $\eta$, the network may start diverging instead of converging. On the other hand, if you choose a smaller value for $\eta$, it will take more time for the network to converge. In addition, it may easily get stuck in local minima. The typical solution for this problem is to reduce the learning rate during training [52].

There are three common approaches used for reducing the learning rate during training: constant, factored, and exponential decay. First, we can define a constant $\zeta$ which is applied to reduce the learning rate manually with a defined step function. Second, the learning rate can be adjusted during training with the following equation:

$$\eta_t = \eta_0 \, \beta^{t/\epsilon} \quad (9)$$

Where $\eta_t$ is the t[th] round learning rate, $\eta_0$ is the initial learning rate, and $\beta$ is the decay factor with a value between the range of (0,1).

The step function format for exponential decay is:

$$\eta_t = \eta_0 \, \beta^{\lfloor t/\epsilon \rfloor} \qquad (10)$$

The common practice is to use a learning rate decay of $\beta = 0.1$ to reduce the learning rate by a factor of 10 at each stage.

*G. Weight decay*

Weight decay is used for training deep learning models as a L2 regularization approach, which helps to prevent over fitting the network and model generalization. L2 regularization for $\mathcal{F}(\theta, x)$ can be define as:

(with 1 or 2 hidden layers). Hinton's revolutionary work on DBNs spearheaded a change in this in 2006 [50, 53].

Due to their composition, many layers of DNNs are more capable at representing highly varying nonlinear functions compared to shallow learning approaches [56, 57, and 58]. Moreover, DNNs are more efficient for learning because of the combination of feature extraction and classification layers. The following sections discuss in detail about different DL approaches with necessary components.

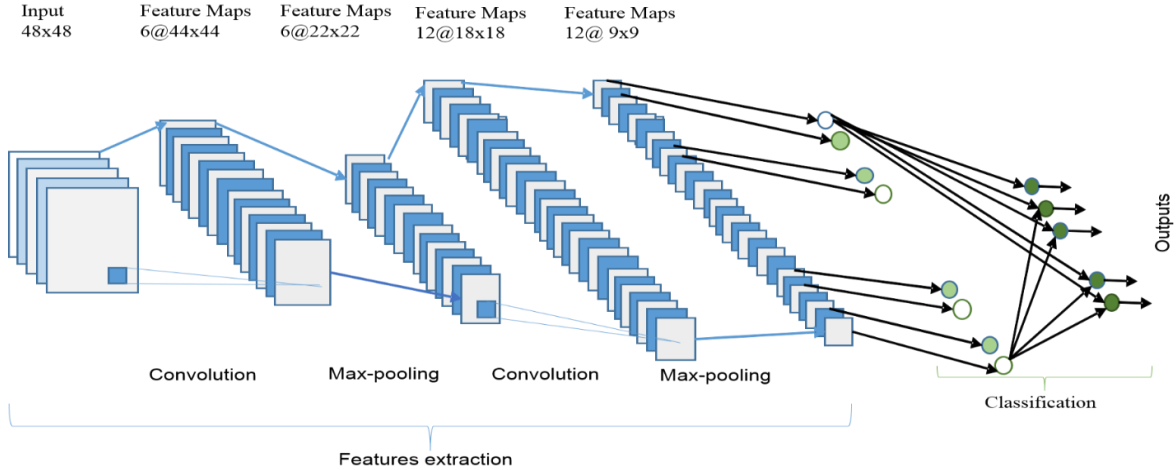

**Fig. 11.** The overall architecture of the CNN includes an input layer, multiple alternating convolution and max-pooling layers, one fully-connected layer and one classification layer.

$$\Omega = \|\theta\|^2 \qquad (11)$$

$$\hat{\varepsilon}(\mathcal{F}(\theta, x), y) = \varepsilon(\mathcal{F}(\theta, x), y) + \tfrac{1}{2}\lambda\,\Omega \qquad (12)$$

The gradient for the weight $\theta$ is:

$$\frac{\partial \tfrac{1}{2}\lambda\Omega}{\partial \theta} = \lambda \cdot \theta \qquad (13)$$

General practice is to use the value $\lambda = 0.0004$. A smaller $\lambda$ will accelerate training.

Other necessary components for efficient training including data preprocessing and augmentation, network initialization approaches, batch normalization, activation functions, regularization with dropout, and different optimization approaches (as discussed in Section 4).

In the last few decades, many efficient approaches have been proposed for better training of deep neural networks. Before 2006, attempts taken at training deep architectures failed: training a deep supervised feed-forward neural network tended to yield worse results (both in training and in test error) then shallow ones

### III. CONVOLUTIONAL NEURAL NETWORKS (CNN)

*A. CNN overview*

This network structure was first proposed by Fukushima in 1988 [48]. It was not widely used however due to limits of computation hardware for training the network. In the 1990s, LeCun *et al*. applied a gradient-based learning algorithm to CNNs and obtained successful results for the handwritten digit classification problem [49]. After that, researchers further improved CNNs and reported state-of-the-art results in many recognition tasks. CNNs have several advantages over DNNs, including being more similar to the human visual processing system, being highly optimized in structure for processing 2D and 3D images, and being effective at learning and extracting abstractions of 2D features. The max pooling layer of CNNs is effective in absorbing shape variations. Moreover, composed of sparse connections with tied weights, CNNs have significantly fewer parameters than a fully connected network of similar size. Most of all, CNNs are trained with the gradient-based learning algorithm, and suffer less from the diminishing gradient problem. Given that the gradient-based algorithm trains the whole network to minimize an error criterion directly, CNNs can produce highly optimized weights.

Fig. 11 shows the overall architecture of CNNs consist of two main parts: feature extractors and a classifier. In the feature extraction layers, each layer of the network receives the output from its immediate previous layer as its input, and passes its output as the input to the next layer. The CNN architecture consists of a combination of three types of layers: convolution, max-pooling, and classification. There are two types of layers in the low and middle-level of the network: convolutional layers and max-pooling layers. The even numbered layers are for convolutions and the odd numbered layers are for max-pooling operations. The output nodes of the convolution and max-pooling layers are grouped into a 2D plane called feature mapping. Each plane of a layer is usually derived of the combination of one or more planes of previous layers. The nodes of a plane are connected to a small region of each connected planes of the previous layer. Each node of the convolution layer extracts the features from the input images by convolution operations on the input nodes.

Higher-level features are derived from features propagated from lower level layers. As the features propagate to the highest layer or level, the dimensions of features are reduced depending on the size of kernel for the convolutional and max-pooling operations respectively. However, the number of feature maps usually increased for representing better features of the input images for ensuring classification accuracy. The output of the last layer of the CNN are used as the input to a fully connected network which is called classification layer. Feed-forward neural networks have been used as the classification layer as they have better performance [50, 58]. In the classification layer, the desired number of features are selected as inputs with respect to the dimension of the weight matrix of the final neural network. However, the fully connected layers are expensive in terms of network or learning parameters. Nowadays, there are several new techniques including average pooling and global average pooling that are used as an alternative of fully-connected networks. The score of the respective class is calculated in the top classification layer using a soft-max layer. Based on the highest score, the classifier gives output for the corresponding classes. Mathematical details on different layers of CNNs are discussed in the following section.

*1) Convolution Layer*

In this layer, feature maps from previous layers are convolved with learnable kernels. The output of the kernels go through a linear or non-linear activation function such as a(sigmoid, hyperbolic tangent, Softmax, rectified linear, and identity functions) to form the output feature maps. Each of the output feature maps can be combined with more than one input feature map. In general, we have that

$$x_j^l = f\left(\sum_{i \in M_j} x_i^{l-1} * k_{ij}^l + b_j^l\right) \quad (14)$$

where $x_j^l$ is the output of the current layer, $x_i^{l-1}$ is the previous layer output, $k_{ij}^l$ is the kernel for the present layer, and $b_j^l$ are biases for the current layer. $M_j$ represents a selection of input maps. For each output map, an additive bias $b$ is given. However, the input maps will be convolved with distinct kernels to generate the corresponding output maps. The output maps finally go through a linear or non-linear activation function (such as sigmoid, hyperbolic tangent, Softmax, rectified linear, or identity functions).

*2) Sub-sampling Layer*

The subsampling layer performs the down sampled operation on the input maps. This is commonly known as the pooling layer. In this layer, the number of input and output feature maps does not change. For example, if there are *N* input maps, then there will be exactly *N* output maps. Due to the down sampling operation the size of each dimension of the output maps will be reduced, depending on the size of the down sampling mask. For example: if a 2×2 down sampling kernel is used, then each output dimension will be the half of the corresponding input dimension for all the images. This operation can be formulated as

$$x_j^l = \text{down}(x_j^{l-1}) \quad (15)$$

where down(.) represents a sub-sampling function. Two types of operations are mostly performed in this layer: average pooling or max-pooling. In the case of the average pooling approach, the function usually sums up over N×N patches of the feature maps from the previous layer and selects the average value. On the other hand, in the case of max-pooling, the highest value is selected from the N×N patches of the feature maps. Therefore, the output map dimensions are reduced by n times. In some special cases, each output map is multiplied with a scalar. Some alternative sub-sampling layers have been proposed, such as fractional max-pooling layer and sub-sampling with convolution. These are explained in Section 4.6.

*3) Classification Layer*

This is the fully connected layer which computes the score of each class from the extracted features from a convolutional layer in the preceding steps. The final layer feature maps are represented as vectors with scalar values which are passed to the fully connected layers. The fully connected feed-forward neural layers are used as a soft-max classification layer. There are no strict rules on the number of layers which are incorporated in the network model. However, in most cases, two to four layers have been observed in different architectures including LeNet [49], AlexNet [7], and VGG Net [9]. As the fully connected layers are expensive in terms of computation, alternative approaches have been proposed during the last few years. These include the global average pooling layer and the average pooling layer which help to reduce the number of parameters in the network significantly.

In the backward propagation through the CNNs, the fully connected layers update following the general approach of fully connected neural networks (FCNN). The filters of the convolutional layers are updated by performing the full convolutional operation on the feature maps between the convolutional layer and its immediate previous layer. Fig. 12 shows the basic operations in the convolution and sub-sampling of an input image.

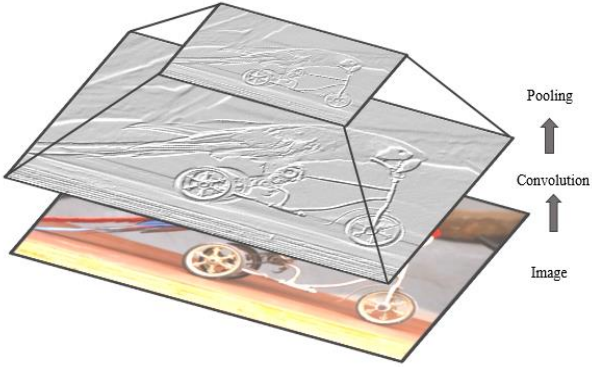

**Fig. 12**. Example of convolution and pooling operation.

*4) Network parameters and required memory for CNN*

The number of computational parameters is an important metric to measure the complexity of a deep learning model. The size of the output feature maps can be formulated as follows:

$$M = \frac{(N-F)}{S} + 1 \qquad (16)$$

Where $N$ refers to the dimensions of the input feature maps, $F$ refers to the dimensions of the filters or the receptive field, $M$ refers to the dimensions of output feature maps, and $S$ stands for the stride length. Padding is typically applied during the convolution operations to ensure the input and output feature map have the same dimensions. The amount of padding depends on the size of the kernel. Equation 17 is used for determining the number of rows and columns for padding.

$$P = (F - 1)/2 \qquad (17)$$

Here $P$ is the amount of padding and $F$ refers to the dimension of the kernels. Several criteria are considered for comparing the models. However, in most of the cases, the number of network parameters and the total amount of memory are considered. The number of parameters ($Parm_l$) of $l^{th}$ layer is calculated based on the following equation:

$$Parm_l = (F \times F \times FM_{l-1}) \times FM_l \qquad (18)$$

If bias is added with the weights, then the above equation can be written as follows:

$$Parm_l = (F \times (F + 1) \times FM_{l-1}) \times FM_l \qquad (19)$$

Here the total number of parameters of $l^{th}$ layer can be represented with $P_l$, $FM_l$ is for the total number of output feature maps, and $FM_{l-1}$ is the total number of input feature maps or channels. For example, let's assume the $l^{th}$ layer has $FM_{l-1} = 32$ input features maps, $FM_l = 64$ output feature maps, and the filter size is $F = 5$. In this case, the total number of parameters with bias for this layer is

$$Parm_l = (5 \times 5 \times 33) \times 64 = 528,000$$

Thus, the amount of memory ($Mem_l$) needs for the operations of the $l^{th}$ layer can be expressed as

$$Mem_l = (N_l \times N_l \times FM_l) \qquad (20)$$

*B. Popular CNN architectures*

We will now examine several popular state-of-the-art CNN architectures. In general, most deep convolutional neural networks are made of a key set of basic layers, including the convolution layer, the sub-sampling layer, dense layers, and the soft-max layer. The architectures typically consist of stacks of several convolutional layers and max-pooling layers followed by a fully connected and SoftMax layers at the end. Some examples of such models are LeNet [49], AlexNet [7], VGG Net [9], NiN [60] and all convolutional (All Conv) [61]. Other alternative and more efficient advanced architectures have been proposed including GoogLeNet with Inception units [10, 64], Residual Networks [11], DenseNet [62], and FractalNet [63]. The basic building components (convolution and pooling) are almost the same across these architectures. However, some topological differences are observed in the modern deep learning architectures. Of the many DCNN architectures, AlexNet [7], VGG [9], GoogLeNet [10, 64], Dense CNN [62] and FractalNet [63] are generally considered the most popular architectures because of their state-of-the-art performance on different benchmarks for object recognition tasks. Among all of these structures, some of the architectures are designed especially for large scale data analysis (such as GoogLeNet and ResNet), whereas the VGG network is considered a general architecture. Some of the architectures are dense in terms of connectivity, such DenseNet [62]. Fractal Network is an alternative of ResNet.

*1) LeNet (1998)*

Although LeNet was proposed in the 1990s, limited computation capability and memory capacity made the algorithm difficult to implement until about 2010 [49]. LeCun, however, proposed CNNs with the back-propagation algorithm and experimented on handwritten digits dataset to achieve state-of-the-art accuracies. His architecture is well known as LeNet-5 [49]. The basic configuration of LeNet-5 is (see Fig. 13): 2 convolution (conv) layers, 2 sub-sampling layers, 2 fully connected layers, and an output layer with Gaussian connection. The total number of weights and Multiply and Accumulates (MACs) are 431k and 2.3M respectively.

As computational hardware started improving in capability, CNNs stated becoming popular as an efficient learning approach in the computer vision and machine learning communities.

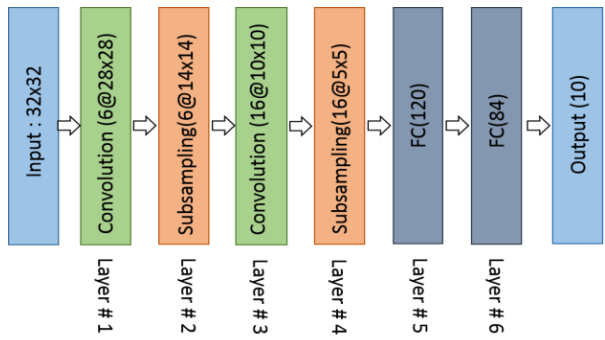

Fig. 13. Architecture of LeNet

*2) AlexNet (2012)*

In 2012, Alex Krizhevsky and others proposed a deeper and wider CNN model compared to LeNet and won the most difficult ImageNet challenge for visual object recognition called the ImageNet Large Scale Visual Recognition Challenge (ILSVRC) in 2012 [7]. AlexNet achieved state-of-the-art recognition accuracy against all the traditional machine learning and computer vision approaches. It was a significant breakthrough in the field of machine learning and computer vision for visual recognition and classification tasks and is the point in history where interest in deep learning increased rapidly.

The architecture of AlexNet is shown in Fig. 14. The first convolutional layer performs convolution and max pooling with Local Response Normalization (LRN) where 96 different receptive filters are used that are 11×11 in size. The max pooling operations are performed with 3×3 filters with a stride size of 2. The same operations are performed in the second layer with 5×5 filters. 3×3 filters are used in the third, fourth, and fifth convolutional layers with 384, 384, and 296 feature maps respectively. Two fully connected (FC) layers are used with dropout followed by a Softmax layer at the end. Two networks with similar structure and the same number of feature maps are trained in parallel for this model. Two new concepts, Local Response Normalization (LRN) and dropout, are introduced in this network. LRN can be applied in two different ways: first applying on single channel or feature maps, where an N×N patch is selected from same feature map and normalized based one the neighborhood values. Second, LRN can be applied across the channels or feature maps (neighborhood along the third dimension but a single pixel or location).

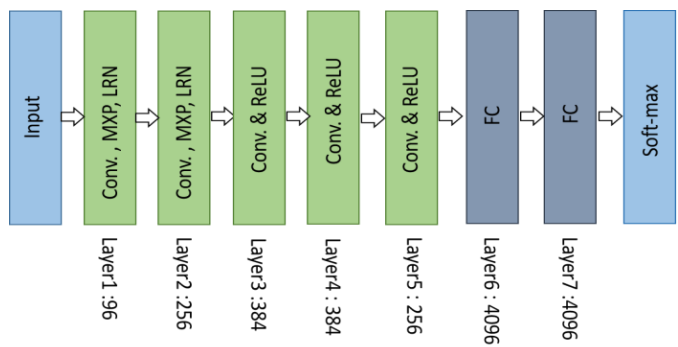

Fig. 14. Architecture of AlexNet: Convolution, max-pooling, LRN and fully connected (FC) layer

AlexNet has 3 convolution layers and 2 fully connected layers. When processing the ImageNet dataset, the total number of parameters for AlexNet can be calculated as follows for the first layer: input samples are 224×224×3, filters (kernels or masks) or a receptive field that has a size 11, the stride is 4, and the output of the first convolution layer is 55×55×96. According to the equations in section 3.1.4, we can calculate that this first layer has 290400 (55×55×96) neurons and 364 (11 ×11×3 = 363 + 1 bias) weights. The parameters for the first convolution layer are 290400×364 = 105,705,600. Table II shows the number of parameters for each layer in millions. The total number of weights and MACs for the whole network are 61M and 724M respectively.

*3) ZFNet / Clarifai (2013)*

In 2013, Matthew Zeiler and Rob Fergue won the 2013 ILSVRC with a CNN architecture which was an extension of AlexNet. The network was called ZFNet [8], after the authors' names. As CNNs are expensive computationally, an optimum use of parameters is needed from a model complexity point of view. The ZFNet architecture is an improvement of AlexNet, designed by tweaking the network parameters of the latter. ZFNet uses 7x7 kernels instead of 11x11 kernels to significantly reduce the number of weights. This reduces the number of network parameters dramatically and improves overall recognition accuracy.

*4) Network in Network (NiN)*

This model is slightly different from the previous models where a couple of new concepts are introduced [60]. The first concept is to use multilayer perception convolution, where convolutions are performed with a 1×1 filters that help to add more nonlinearity in the models. This helps to increase the depth of the network, which can then be regularized with dropout. This concept is used often in the bottleneck layer of a deep learning model.

The second concept is to use Global Average Pooling (GAP) as an alternative of fully connected layers. This helps to reduce the number of network parameters significantly. GAP changes the network structure significantly. By applying GAP on a large feature map, we can generate a final low dimensional feature vector without reducing the dimension of the feature maps.

*5) VGGNET (2014)*

The Visual Geometry Group (VGG), was the runner up of the 2014 ILSVRC [9]. The main contribution of this work is that it shows that the depth of a network is a critical component to achieve better recognition or classification accuracy in CNNs. The VGG architecture consists of two convolutional layers both of which use the ReLU activation function. Following the activation function is a single max pooling layer and several fully connected layers also using a ReLU activation function. The final layer of the model is a Softmax layer for classification. In VGG-E [9] the convolution filter size is changed to a 3x3 filter with a stride of 2. Three VGG-E [9] models, VGG-11,VGG-16, and VGG-19; were proposed the models had 11,16,and 19 layers respectively.

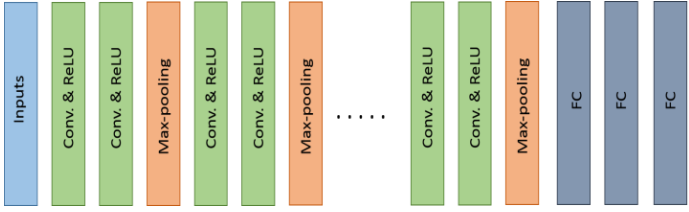

**Fig. 15.** Basic building block of VGG network: Convolution (Conv) and FC for fully connected layers

All versions of the VGG-E models ended the same with three fully connected layers. However, the number of convolution layers varied VGG-11 contained 8 convolution layers, VGG-16 had 13 convolution layers, and VGG-19 had 16 convolution layers. VGG-19, the most computational expensive model, contained 138Mweights and had 15.5M MACs.

*6) GoogLeNet (2014)*

GoogLeNet, the winner of ILSVRC 2014[10], was a model proposed by Christian Szegedy of Google with the objective of reducing computation complexity compared to the traditional CNN. The proposed method was to incorporate "Inception Layers" that had variable receptive fields, which were created by different kernel sizes. These receptive fields created operations that captured sparse correlation patterns in the new feature map stack.

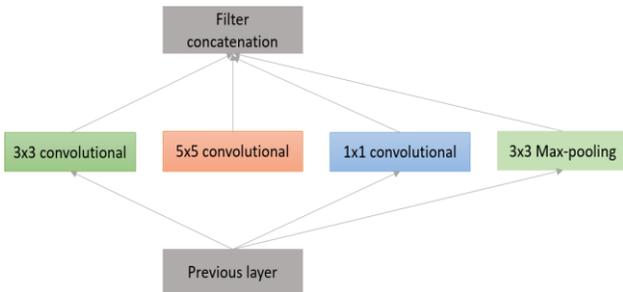

**Fig. 16.** Inception layer: naive version

The initial concept of the Inception layer can be seen in Fig. 16. GoogLeNet improved the state of the art recognition accuracy using a stack of Inception layers seen in Fig. 17. The difference between the naïve inception layer and final Inception Layer was the addition of 1x1 convolution kernels. These kernels allowed for dimensionality reduction before computationally expensive layers. GoogLeNet consisted of 22 layers in total, which was far greater than any network before it. However, the number of network parameters GoogLeNet used was much lower than its predecessor AlexNet or VGG. GoogLeNet had 7M network parameters when AlexNet had 60M and VGG-19 138M. The computations for GoogLeNet also were 1.53G MACs far lower than that of AlexNet or VGG.

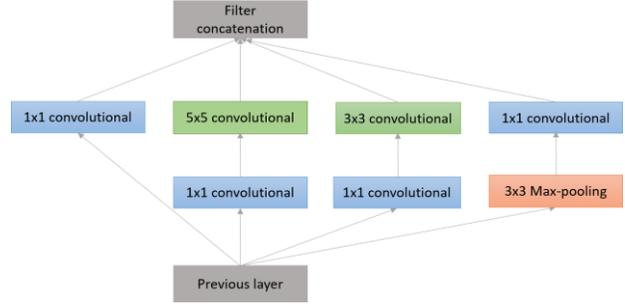

**Fig. 17.** Inception layer with dimension reduction

*7) Residual Network (ResNet in 2015)*

The winner of ILSVRC 2015 was the Residual Network architecture, ResNet[11]. Resnet was developed by Kaiming He with the intent of designing ultra-deep networks that did not suffer from the vanishing gradient problem that predecessors had. ResNet is developed with many different numbers of layers; 34, 50,101, 152, and even 1202. The popular ResNet50 contained 49 convolution layers and 1 fully connected layer at the end of the network. The total number of weights and MACs for the whole network are 25.5M and 3.9M respectively.

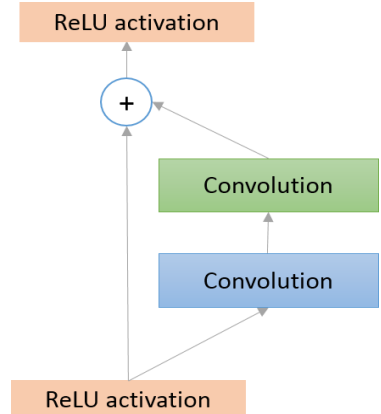

**Fig. 18.** Basic diagram of Residual block

The basic block diagram of the ResNet architecture is shown in Fig. 18. ResNet is a traditional feed forward network with a residual connection. The output of a residual layer can be defined based on the outputs of $(l-1)^{th}$ which comes from the previous layer defined as $x_{l-1}$. $\mathcal{F}(x_{l-1})$ is the output after performing various operations (e.g. convolution with different size of filters, Batch Normalization (BN) followed by an activation function such as a ReLU on $x_{l-1}$). The final output of residual unit is $x_l$ which can be defined with the following equation:

$$x_l = \mathcal{F}(x_{l-1}) + x_{l-1} \qquad (21)$$

The residual network consists of several basic residual blocks. However, the operations in the residual block can be varied depending on the different architecture of residual networks [11]. The wider version of residual network was proposed by Zagoruvko el at. In 2016 [66]. Another improved residual network approach known as aggregated residual transformation was proposed in 2016[67]. Recently, some other variants of residual models have been proposed based on the Residual Network architecture [68, 69, and 70]. Furthermore, there are several advanced architectures that have been proposed with the combination of Inception and Residual units. The basic conceptual diagram of Inception-Residual unit is shown in the following Fig.19.

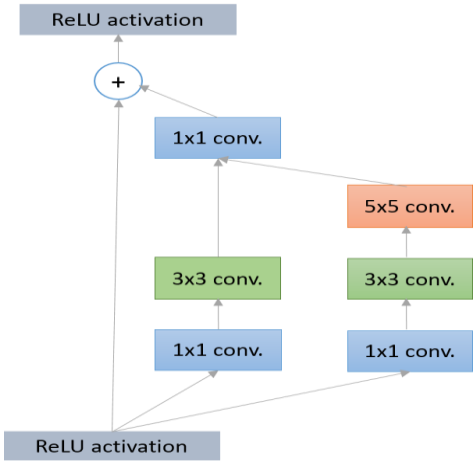

**Fig. 19.** Basic block diagram for Inception Residual unit

Mathematically, this concept can be represented as

$$x_l = \mathcal{F}(x_{l-1}^{3\times 3} \odot x_{l-1}^{5\times 5}) + x_{l-1} \qquad (22)$$

Where the symbol $\odot$ refers the concentration operations between two outputs from the 3×3 and 5×5 filters. After that the convolution operation is performed with 1×1 filters. Finally, the outputs are added with the inputs of this block of $x_{l-1}$. The concept of Inception block with residual connections is introduced in the Inception-v4 architecture [65]. The improved version of the Inception-Residual network known as PolyNet was recently proposed [70,290].

*8) Densely Connected Network (DenseNet)*

DenseNet developed by Gao Huang and others in 2017[62], which consists of densely connected CNN layers, the outputs of each layer are connected with all successor layers in a dense block [62]. Therefore, it is formed with dense connectivity between the layers rewarding it the name "DenseNet". This concept is efficient for feature reuse, which dramatically reduces network parameters. DenseNet consists of several dense blocks and transition blocks, which are placed between two adjacent dense blocks. The conceptual diagram of a dense block is shown in Fig. 20.

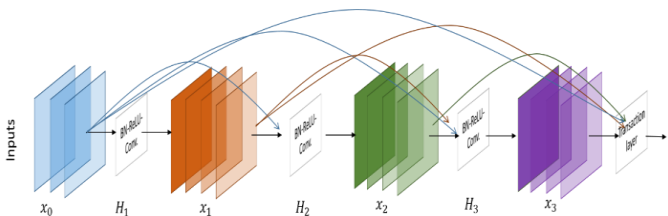

**Fig. 20.** A 4-layer Dense block with growth rate of $k = 3$.

Each layer takes all the preceding feature maps as input.

When deconstructing Fig. 20, the $l^{th}$ layer received all the feature maps from previous layers of $x_0, x_1, x_2 \cdots x_{l-1}$ as input:

$$x_l = H_l([x_0, x_1, x_2 \cdots x_{l-1}]) \qquad (23)$$

Where $[x_0, x_1, x_2 \cdots x_{l-1}]$ are the concatenated features for layers $0, \cdots\cdots, l-1$ and $H_l(\cdot)$ is considered as a single tensor. It performs three different consecutive operations: Batch-Normalization (BN) [110], followed by a ReLU [58] and a $3 \times 3$ convolution operation. In the transaction block, $1 \times 1$ convolutional operations are performed with BN followed by a $2 \times 2$ average pooling layer. This new model shows state-of-the-art accuracy with a reasonable number of network parameters for object recognitions tasks.

*9) FractalNet (2016)*

This architecture is an advanced and alternative architecture of ResNet model, which is efficient for designing large models with nominal depth, but shorter paths for the propagation of gradient during training [63]. This concept is based on drop-path which is another regularization approach for making large networks. As a result, this concept helps to enforce speed versus accuracy tradeoffs. The basic block diagram of FractalNet is shown in Fig. 21.

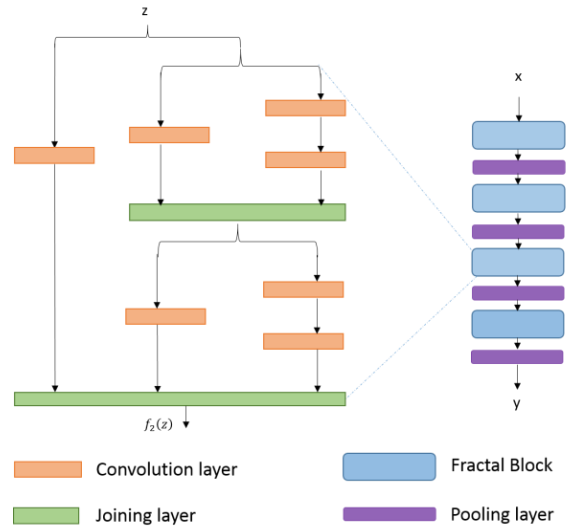

**Fig. 21.** The detailed FractalNet module on the left and FractalNet on the right

*C. CapsuleNet*

CNNs are an effect methodology for detecting features of an object and achieving good recognition performance compared to state of the art hand crafted feature detectors. There are limits to CNNs, which are that it does not take into account special relationships, perspective, size and orientation, of features, For example: if you have a face image, it does not matter the placement of different components (nose, eye, mouth etc.) of the faces neurons of a CNN will wrongly active and recognition as face without considering special relationships (orientation, size). Now, imagine a neuron which contains the likelihood with properties of features (perspective, orientation, size etc.). This special type of neurons, capsules, can detect face efficiently with distinct information. The capsule network consists of several layers of capsule nodes. The first version of capsule network (CapsNet) consisted of three layers of capsule nodes in an encoding unit.

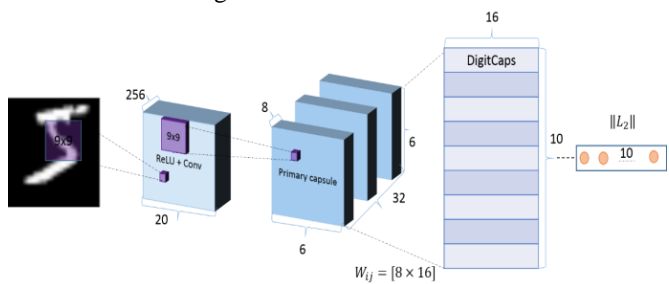

**Fig.22**. A CapsNet encoding unit with 3 layers. The instance of each class is represented with a vector of a capsule in DigitCaps layer that is used for calculating classification loss. The weights between primary capsule layer and DigitCaps layer are represented with $W_{ij}$.

This architecture for MNIST (28×28) images, the 256 9×9 kernels are applied with a stride 1, so the output is $(28 − 9 + 1 = 20)$ with 256 feature maps. Then the outputs are feed to the primary capsule layer which is a modified convolution layer that generates an 8-D vector instead of a scalar. In the first convolutional layer, 9×9 kernels are applied with stride 2, the output dimension is $((20 − 9)/2 + 1 = 6)$. The primary capsules are used 8×32 kernels which generates 32×8×6×6 (32 groups for 8 neurons with 6×6 size).

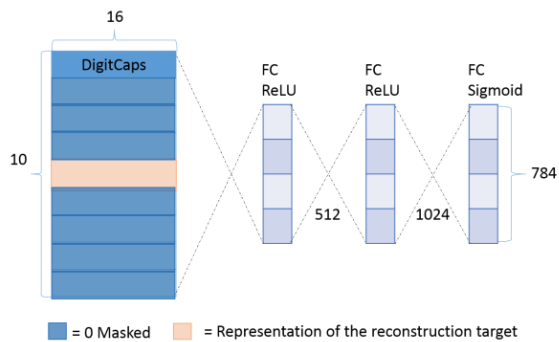

**Fig. 23**. The decoding unit where a digit is reconstructed from DigitCaps layer representation. The Euclidean distance is used minimizing error between input sample and reconstructed sample from sigmoid layer. True labels are used for reconstruction target during training.

The entire encoding and decoding processes of CapsNet is shown in Fig. 22 and Fig. 23 respectively. We used max pooling layer in CNN often that can handle translation variance. Even if a feature moves if it is still under a max pooling window it can be detected. As the capsule contains the weighted sum of features from the previous layer, therefore this approach is capable of detecting overlapped features which is important for segmentation and detection tasks.

In the traditional CNN, we have used a single cost function to evaluate the overall error which propagates backward during training. However, in this case if the weight between two neurons is zero then the activation of a neuron is not propagated from that neuron. The signal is routed with respect to the feature parameters rather than a one size fits all cost function in iterative dynamic routing with agreement. For details about this architecture, please see [293]. This new CNN architecture provides state-of-the-art accuracy for handwritten digits recognition on MNIST. However, from an application point of view, this architecture is more suitable for segmentation and detection tasks compare to classification tasks.

### D. Comparison on different models

The comparison of recently proposed models based on error, network parameters, and maximum number of connections are given in Table II.

### E. Other models

There are many other network architectures such as fast region based CNN [71] and Xception [72] which are popular in the computer vision community. In 2015 a new model was proposed using recurrent convolution layers named Recurrent Convolution Neural Network or RCNN[73]. The improved version of this network is a combination of the two most popular architectures in the Inception network and Recurrent Convolutional Network, Inception Convolutional Recurrent Neural Networks(IRCNN)[74]. IRCNN provided better accuracy compared RCNN and inception network with almost identical network parameters. Visual Phase Guided CNN (ViP CNN) is proposed with phase guided message passing structure (PMPS) to build connections between relational components, which show better speed up and recognition accuracy [75]. Look up based CNN[76] is a fast, compact, and accurate model enabling efficient inference. In 2016 the architecture known as fully convolutional network (FCN) was proposed for segmentation tasks where it is now commonly used. Other recently proposed CNN models include deep network with stochastic depth, deeply-supervised networks and ladder network [79, 80, and 81]

**The Question is, do deep nets really need to be deeper?** Some papers have been published base on the justification for deeper networks and concluded that "Deeper is better" [82, 83]. Now the question is which one is better width versus depth? On the one hand, there is controversy whether deep or wide networks are better some studies can be seen in the following papers [84, 85, 86]. As DL approaches are data driven techniques which require a lot of labeled samples for training for the supervised approach. Recently some frameworks have been developed for making efficient databases from labeled and un-labeled datasets [87, 88].

Hyper parameter optimization allows for variable levels of performance, which is helpful for creating models to pair with designing hardware for deep learning [89,90].

### F. Applications of CNNs

Most of the techniques that have been discussed above are evaluated on computer vision and image processing tasks. Here are some recently published papers that have been discussed, which are applied for different modalities of computer vision and image processing.

**TABLE II.** THE TOP-5% ERRORS WITH COMPUTATIONAL PARAMETERS AND MACS FOR DIFFERENT DEEP CNN MODELS.

| Methods | LeNet-5[48] | AlexNet [7] | OverFeat (fast)[8] | VGG-16[9] | GoogLeNet [10] | ResNet-50(v1)[11] |
|---|---|---|---|---|---|---|
| Top-5 errors | n/a | 16.4 | 14.2 | 7.4 | 6.7 | 5.3 |
| Input size | 28x28 | 227x227 | 231x231 | 224x224 | 224x224 | 224x224 |
| Number of Conv Layers | 2 | 5 | 5 | 16 | 21 | 50 |
| Filter Size | 5 | 3,5,11 | 3,7 | 3 | 1,3,5,7 | 1,3,7 |
| Number of Feature Maps | 1,6 | 3-256 | 3-1024 | 3-512 | 3-1024 | 3-1024 |
| Stride | 1 | 1,4 | 1,4 | 1 | 1,2 | 1,2 |
| Number of Weights | 26k | 2.3M | 16M | 14.7M | 6.0M | 23.5M |
| Number of MACs | 1.9M | 666M | 2.67G | 15.3G | 1.43G | 3.86G |
| Number of FC layers | 2 | 3 | 3 | 3 | 1 | 1 |
| Number of Weights | 406k | 58.6M | 130M | 124M | 1M | 1M |
| Number of MACs | 405k | 58.6M | 130M | 124M | 1M | 1M |
| Total Weights | 431k | 61M | 146M | 138M | 7M | 25.5M |
| Total MACs | 2.3M | 724M | 2.8G | 15.5G | 1.43G | 3.9G |

*1) CNNs for solving Graph problem*

Learning graph data structures is a common problem with various different applications in data mining and machine learning tasks. DL techniques have made a bridge in between the machine learning and data mining groups. An efficient CNN for arbitrary graph processing was proposed in 2016[91].

*2) Image processing and computer vision*

Most of the models, we have discussed above are applied on different application domains including image classification, detection, segmentation, localization, captioning, video classification and many more. There is a good survey on deep learning approaches for image processing and computer vision related tasks [92]. Single image super-resolution using CNN methods [93]. Image de-noising using block-matching CNN [94]. Photo aesthetic assessment using A-Lamp: Adaptive Layout-Aware Multi-Patch Deep CNN [95]. DCNN for hyper spectral imaging for segmentation using Markov Random Field (MRF) [96]. Image registration using CNN [97]. The Hierarchical Deep CNN for Fast Artistic Style Transfer [98]. Background segmentation using DCNN [99]. Handwritten character recognition using DCNN approaches [291]. Optical image classification using deep learning approaches [296]. Object recognition using cellular simultaneous recurrent networks and convolutional neural network [297].

*3) Speech processing*

CNN methods are also applied for speech processing: speech enhancement using multimodal deep CNN [100], and audio tagging using Convolutional Gated Recurrent Network (CGRN) [101].

*4) CNN for medical imaging*

A good survey on DL for medical imaging for classification, detection, and segmentation tasks [102]. There are some papers published after this survey. MDNet, which was developed for medical diagnosis with images and corresponding text description [103]. Cardiac Segmentation using short-Axis MRI [104]. Segmentation of optic disc and retina vasculature using CNN [105]. Brain tumor segmentation using random forests with features learned with fully convolutional neural network (FCNN) [106].

IV. ADVANCED TRAINING TECHNIQUES

What is missing in the previous section is the advanced training techniques or components which need to be considered carefully for efficient training of DL approaches. There are different advanced techniques to apply to train a deep learning model better. The techniques including input pre-processing, better initialization method, batch normalization, alternative convolutional approaches, advanced activation functions, alternative pooling techniques, network regularization approaches, and better optimization method for training. The following sections are discussed on individual advanced training techniques individually.

*A. Preparing dataset*

Presently different approaches have been applied before feeding the data to the network. The different operations to prepare a dataset are as follows; sample rescaling, mean subtraction, random cropping, flipping data with respective to the horizon or vertical axis, color jittering, PCA/ZCA whitening and many more.

*B. Network initialization*

The initialization of deep networks has a big impact on the overall recognition accuracy [53,54]. Previously, most of the networks have been initialized with random weights. For complex tasks with high dimensionality data training a DNN becomes difficult because weights should not be symmetrical due to the back-propagation process. Therefore, effective initialization techniques are important for training this type of DNN. However, there are many efficient techniques that have been proposed during last few years. In 1998, LeCun [107] and Y. Bengio in 2010 [108] proposed a simple but effective approach. In this method, the weights are scaled by the inverse of the square root of number of input neurons of the layer, which can be stated $1/\sqrt{N_l}$, where $N_l$ is the number of input neurons of $l^{th}$ layer. The deep network initialization approach of Xavier has been proposed based on the symmetric activation function with respect to the hypothesis of linearity. This approach is known as "Xavier" initialization approach.

Recently in 2016, Dmytro M. et al. proposed Layer-sequential unit-invariance(LSUV), which is a data-driven initialization approach and provides good recognition accuracy on several benchmark datasets including ImageNet [ 85]. One of the popular initialization approaches has proposed by Kiming He in 2015 [109]. The distribution of the weights of $l^{th}$ layer will be normal distribution with mean zero and variance $\frac{2}{n_l}$ which can be expressed as follows.

$$w_l \sim \mathcal{N}\left(0, \frac{2}{n_l}\right) \quad (24)$$

### C. Batch Normalization

Batch normalization helps accelerate DL processes by reducing internal covariance by shifting input samples. What that means is the inputs are linearly transformed to have zero mean and unit variance. For whitened inputs, the network converges faster and shows better regularization during training, which has an impact on the overall accuracy. Since the data whitening is performed outside of the network, there is no impact of whitening during training of the model. In the case of deep recurrent neural networks, the inputs of the $n^{th}$ layer are the combination of n-$1^{th}$ layer, which is not raw feature inputs. As the training progresses the effect of normalization or whitening reduce respectively, which causes the vanishing gradient problem. This can slow down entire training process and cause saturation. To better the training process during training batch normalization is then applied to the internal layers of the deep neural network. This approach ensures faster convergence in theory and during experiment on benchmarks. In batch normalization, the features of a layer are independently normalized with mean zero and variance one [110,111]. The algorithm of Batch normalization is given in Algorithm IV.

---

**Algorithm IV**: Batch Normalization (BN)

**Inputs:** Values of x over a mini-batch: $\mathcal{B} = \{x_{1,2,3,\ldots,m}\}$
**Outputs:** $\{y_i = BN_{\gamma,\beta}(x_i)\}$
$\mu_{\mathcal{B}} \leftarrow \frac{1}{m}\sum_{i=1}^{m} x_i$  // mini-batch mean
$\sigma_{\mathcal{B}}^2 \leftarrow \frac{1}{m}\sum_{i=1}^{m}(x_i - \mu_{\mathcal{B}})^2$  // mini-batch variance
$\hat{x}_i \leftarrow \frac{x_i - \mu_{\mathcal{B}}}{\sqrt{\sigma_{\mathcal{B}}^2 + \epsilon}}$  // normalize
$y_i = \gamma \hat{x}_i + \beta \equiv BN_{\gamma,\beta}(x_i)$  // Scaling and shifting

---

The parameters $\gamma$ and $\beta$ are used for the scale and shift factor for the normalization values, so normalization does not only depend on layer values. If you use normalization techniques, the following criterions are recommended to consider during implementation:
- Increase learning rate
- Dropout (batch normalization does the same job)
- $L_2$ weight regularization
- Accelerating the learning rate decay

- Remove Local Response Normalization (LRN) (if you used it)
- Shuffle training sample more thoroughly
- Use less distortion of images in the training set

### D. Alternative Convolutional methods

Alternative and computationally efficient convolutional techniques that reduces the cost of multiplications by factor of 2.5 have been proposed [112].

### E. Activation function

The traditional Sigmoid and Tanh activation functions have been used for implementing neural network approaches in the past few decades. The graphical and mathematical representation is shown in Fig. 24.

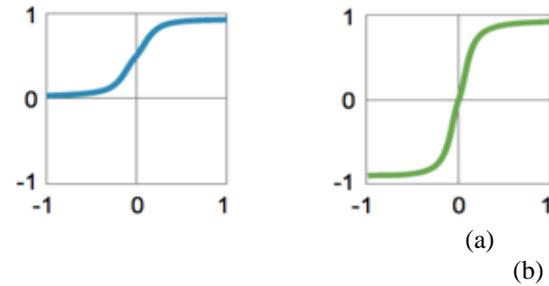

**Fig. 24.** Activation function: (a) sigmoid function and (b) Hyperbolic transient

**Sigmoid:**

$$y = \frac{1}{1+e^x} \quad (25)$$

**TanH:**

$$y = \frac{e^x - e^{-x}}{e^x + e^{-x}} \quad (26)$$

The popular activation function called Rectified Linear Unit (ReLU) proposed in 2010 solves the vanishing gradient problem for training deep learning approaches. The basic concept is simple keep all the values above zero and sets all negative values to zero that is shown in Fig. 25[58]. The ReLU activation was first used in AlexNet, which was a breakthrough deep CNN proposed in 2012 by Hinton [7].

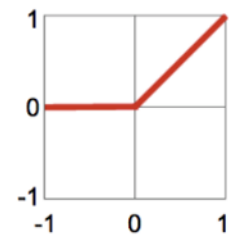

**Fig. 25.** Pictorial representation of Rectified Linear Unit (ReLU)

Mathematically we can express ReLU as follows:

$$y = \max(0, x) \quad (27)$$

As the activation function plays a crucial role in learning the weights for deep architectures. Many researchers focus here because there is much that can be done in this area. Meanwhile, there are several improved versions of ReLU that have been proposed, which provide even better accuracy compared to the ReLU activation function. An efficient improved version of ReLU activation function is called the parametric ReLU (PReLU) proposed by Kaiming He et al. in 2015. The Fig.26 shows the pictorial representation of Leaky ReLU and ELU activation functions. This technique can automatically learn the parameters adaptively and improve the accuracy at negligible extra computing cost [109].

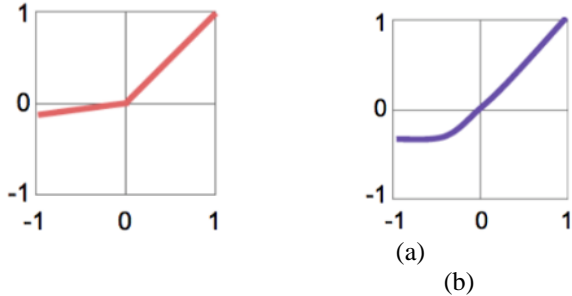

**Fig. 26.** Diagram for (a) Leaky ReLU (b) Exponential Linear Unit (ELU)

**Leaky ReLU:**

$$y = \max(ax, x) \quad (28)$$

Here $a$ is a constant, the value is 0.1.

**ELU:**

$$y = \begin{cases} x, & x \geq 0 \\ a(e^x - 1), & x < 0 \end{cases} \quad (29)$$

The recent proposal of the Exponential Linear Unit activation function, which allowed for a faster and more accurate version of the DCNN structure [113]. Furthermore, tuning the negative part of activation function creates the leaky ReLU with Multiple Exponent Linear Unit (MELU) that are proposed recently [114]. S shape Rectified Linear Activation units are proposed in 2015 [115]. A survey on modern activation functions was conducted in 2015 [116].

*F. Sub-sampling layer or pooling layer*

At present, two different techniques have been used for implementation of deep networks in the sub-sampling or pooling layer: average and max-pooling. The concept of average pooling layer was used for the first time in LeNet [49] and AlexNet used Max-pooling layers instead in 2012[7]. The conceptual diagram for max pooling and average pooling operation are shown in the Fig 27. The concept of special pyramid pooling has been proposed by He et al. in 2014 which is shown in Fig. 28 [117].

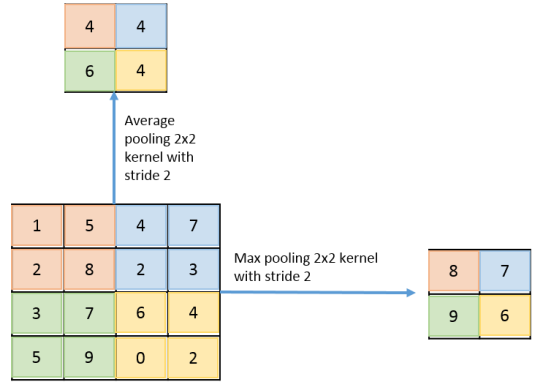

**Fig. 27.** Average and max pooling operations.

The multi-scale pyramid pooling was proposed in 2015 [118]. In 2015, Benjamin G. proposed a new architecture with Fractional max pooling, which provides state-of-the-art classification accuracy for CIFAR-10 and CIFAR-100 datasets. This structure generalizes the network by considering two important properties for sub-sampling layer or pooling layer. First, the non-overlapped max-pooling layer limits the generalize of the deep structure of the network, this paper proposed a network with 3x3 overlapped max-pooling with 2-stride instead of 2x2 as sub-sampling layer [119]. Another paper which has conducted research on different types of pooling approaches including mixed, gated, and tree as generalization of pooling functions [120].

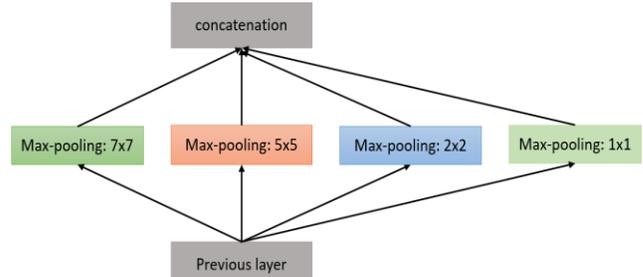

**Fig. 28.** Spatial pyramid pooling

*G. Regularization approaches for DL*

There are different regularization approaches that have been proposed in the past few years for deep CNN. The simplest but efficient approach called "dropout" was proposed by Hinton in 2012 [121]. In Dropout a randomly selected subset of activations are set to zero within a layer [122]. The dropout concept is shown in Fig. 29.

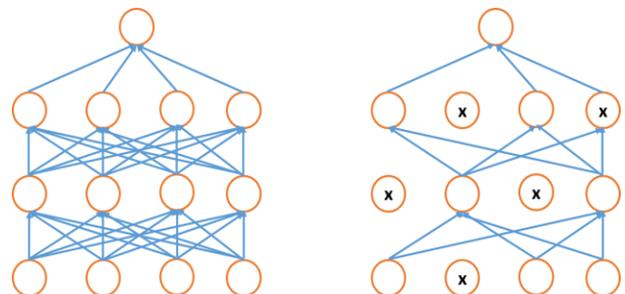

**Fig. 29.** Pictorial representation of the concept Dropout

Another regularization approach is called Drop Connect, in this case instead of dropping the activation, the subset of weights within the network layers are set to zero. As a result, each layer receives the randomly selected subset of units from the immediate previous layer [123]. Some other regularization approaches are proposed as well, details in [124].

*H. Optimization methods for DL*

There are different optimization methods such as SGD, Adagrad, AdaDelta, RMSprop, and Adam [125]. Some activation functions have been improved upon such as in the case of SGD where it was been proposed with an added variable momentum, which improved training and testing accuracy. In the case of Adagrad, the main contribution was to calculate adaptive learning rate during training. For this method the summation of the magnitude of the gradient is considered to calculate the adaptive learning rate. In the case with a large number of epochs, the summation of magnitude of the gradient becomes large. The result of this is the learning rate decreases radically, which causes the gradient to approach zero quickly. The main drawback to this approach is that it causes problems during training. Later, RMSprop was proposed considering only the magnitude of the gradient of the immediate previous iteration, which prevents the problems with Adagrad and provides better performance in some cases. The Adam optimization approach is proposed based on the momentum and the magnitude of the gradient for calculating adaptive learning rate similar RMSprop. Adam has improved overall accuracy and helps for efficient training with better convergence of deep learning algorithms [126]. The improved version of the Adam optimization approach has been proposed recently, which is called EVE. EVE provides even better performance with fast and accurate convergence [127].

## V. RECURRENT NEURAL NETWORKS (RNN)

*A. Introduction*

Human thoughts have persistence; Human don't throw a thing away and start their thinking from the scratch in a second. As you are reading this article, you are understanding each word or sentence based on the understanding of previous words or sentences. The traditional neural network approaches including DNNs and CNNs cannot deal with this type of problem. The standard Neural Networks and CNN are incapable due to the following reasons. First, these approaches only handle a fixed-size vector as input (e.g., an image or video frame) and produce a fixed-size vector as output (e.g., probabilities of different classes). Second, those models operate with a fixed number of computational steps (e.g. the number of layers in the model). The RNNs are unique as they allow operation over a sequence of vectors over time. This idea of RNNs were developed in 1980. The Hopfield Newark introduced this concept in 1982 but the idea was described shortly in 1974 [128]. The pictorial representation is shown in Fig. 30.

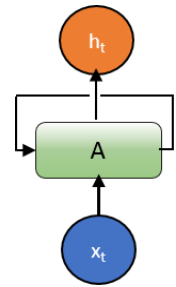

**Fig. 30.** The structure of basic RNNs with loop.

Different versions of RNNs have been proposed in Jordan and Elman. In The Elman architecture uses the output from a hidden layers as inputs alongside the normal inputs of hidden layers [129]. On the other hand, the outputs from output unit are used as inputs with the inputs of hidden layer in Jordan network [130]. Jordan in contrast uses inputs from the outputs of the output unit with the inputs to the hidden layer. Mathematically we can express these as:

Elman network [129]:

$$h_t = \sigma_h(w_h x_t + u_h h_{t-1} + b_h) \quad (30)$$

$$y_t = \sigma_y(w_y h_t + b_y) \quad (31)$$

Jordan network [130]

$$h_t = \sigma_h(w_h x_t + u_h y_{t-1} + b_h) \quad (32)$$

$$y_t = \sigma_y(w_y h_t + b_y) \quad (33)$$

Where $x_t$ is a vector of inputs, $h_t$ are hidden layer vectors, $y_t$ are the output vectors, $w$ and $u$ are weight matrices and $b$ is the bias vector.

A loop allows information to be passed from one step of the network to the next. A recurrent neural network can be thought of as multiple copies of the same network, each network passing a message to a successor. The diagram below shows what happens if we unroll the loop.

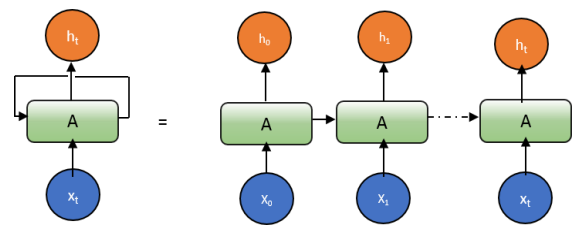

**Fig. 31.** An unrolled RNNs

The main problem with RNN approaches is they experience the vanishing gradient problem. For the first time, this problem is solved by Hochreiter el at. in 1992 [131]. A deep RNN consisting of 1000 subsequent layers was implemented and evaluated to solve deep learning tasks in 1993 [132]. There are several solutions that have been proposed for solving the vanishing gradient problem of RNN approaches in the past few decades. Two possible effective solutions to this problem are first to clip the gradient and scale the gradient if the norm is too large, and secondly create a better RNN model. One of better models was introduced by Felix A. el at. in 2000 named Long Short-Term Memory (LSTM) [133,134]. From the LSTM there have been different advanced approaches proposed in the last few years which are explained in the following sections.

The RNN approaches allowed sequences in the input, the output, or in the most general case both. For example: DL for Text Mining, building deep learning models on textual data requires representation of the basic text unit and word. Neural network structures that can hierarchically capture the sequential nature of text. In most of these cases RNNs or Recursive Neural Networks are used for language understanding [292]. In the language modeling, it tries to predict the next word or set of words or some cases sentences based on the previous ones [135]. RNNs are networks with loops in them, allowing information to persist. Another example: the RNNs are able to connect previous information to the present task: using previous video frames, understanding the present and trying to generate the future frames as well [142].

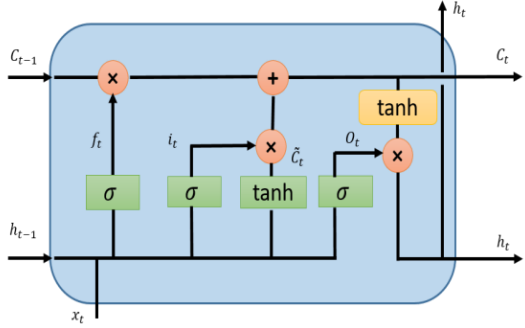

**Fig. 32.** Diagram for Long Short Term Memory (LSTM)

### B. Long Short Term Memory (LSTM)

The key idea of LSTMs is the cell state, the horizontal line running through the top of the Fig. 32. LSTMs remove or add information to the cell state called gates: input gate $(i_t)$, forget gate $(f_t)$ and output gate $(o_t)$ can be defined as :

$$f_t = \sigma(W_f.[h_{t-1}, x_t] + b_f) \quad (34)$$

$$i_t = \sigma(W_i.[h_{t-1}, x_t] + b_i) \quad (35)$$

$$\tilde{C}_t = tanh(W_C.[h_{C-1}, x_t] + b_C) \quad (36)$$

$$C_t = f_t * C_{t-1} + i_t * \tilde{C}_t \quad (37)$$

$$O_t = \sigma(W_O.[h_{t-1}, x_t] + b_O) \quad (38)$$

$$h_t = O_t * tanh(C_t) \quad (39)$$

LSTM models are popular for temporal information processing. Most of the papers that include LSTM models with some minor variance. Some of them are discussed in the following section. There is a slightly modified version of the network with "peephole connections" by Gers and Schimidhuber proposed in 2000. The concept of peepholes is included with almost all the gated in this model.

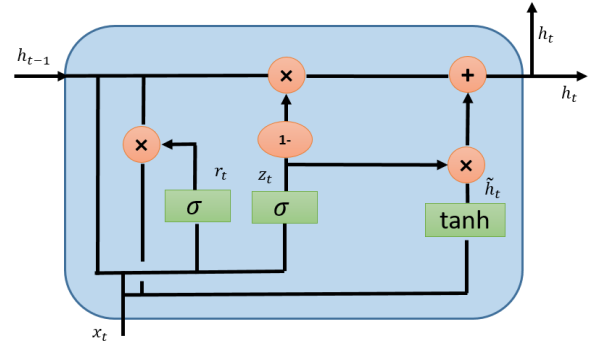

**Fig. 33.** Diagram for Gated Recurrent Unit (GRU)

### C. Gated Recurrent Unit (GRU)

GRU also came from LSTMs with slightly more variation by Cho, et al. in 2014. GRUs are now popular in the community who are working with recurrent networks. The main reason of the popularity is computation cost and simplicity of the model, which is shown in Fig. 33. GRUs are lighter versions of RNN approaches than standard LSTM in term of topology, computation cost and complexity [136]. This technique combines the forget and input gates into a single "update gate" and merges the cell state and hidden state along with some other changes. The simpler model of the GRU has been growing increasingly popular. Mathematically the GRU can be expressed with the following equations:

$$z_t = \sigma(W_z.[h_{t-1}, x_t]) \quad (40)$$

$$r_t = \sigma(W_r.[h_{t-1}, x_t]) \quad (41)$$

$$\tilde{h}_t = tanh(W.[r_t * h_{t-1}, x_t]) \quad (42)$$

$$h_t = (1 - z_t) * h_{t-1} + z_t * \tilde{h}_t \quad (43)$$

**The question is: which one is the best?** According to the different empirical studies there is no clear evidence of a winner. However, the GRU requires fewer network parameters, which makes the model faster. On the other hand, LSTM provides better performance, if you have enough data and computational power [137]. There is a variant LSTM named Deep LSTM [138]. Another variant that is bit different approach called "A clockwork RNN" [139]. There is an important empirical evaluation on a different version of RNN approaches including LSTM by Greff , et al. in 2015 and the final conclusion was all the LSTM variants were all about the same [140]. Another empirical evaluation is conducted on thousands of RNN architecture including LSTM, GRU and so on finding some that worked better than LSTMs on certain tasks [141]

### D. Convolutional LSTM (ConvLSTM)

The problem with fully connected (FC) LSTM and short FC-LSTM model is handling spatiotemporal data and its usage of full connections in the input-to-state and state-to-state transactions, where no spatial information has been encoded. The internal gates of ConvLSTM are 3D tensors, where the last two dimensions are spatial dimensions (rows and columns). The ConvLSTM determines the future state of a certain cell in the grid with respect to inputs and the past states of its local neighbors which can be achieved using convolution operations in the state-to-state or inputs-to-states transition show in Fig. 34.

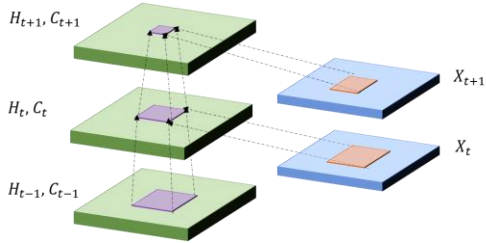

**Fig. 34.** Pictorial diagram for ConvLSTM [142]

ConvLSTM is providing good performance for temporal data analysis with video datasets [142]. Mathematically the ConvLSTM is expressed as follows where * represents the convolution operation and ∘ denotes for Hadamard product:

$$i_t = \sigma(w_{xi} \cdot \mathcal{X}_t + w_{hi} * \mathcal{H}_{t-1} + w_{hi} \circ \mathcal{C}_{t-1} + b_i) \quad (44)$$

$$f_t = \sigma(w_{xf} \cdot \mathcal{X}_t + w_{hf} * \mathcal{H}_{t-1} + w_{hf} \circ \mathcal{C}_{t-1} + b_f) \quad (45)$$

$$\widetilde{C}_t = \tanh(w_{xc} \cdot \mathcal{X}_t + w_{hc} * \mathcal{H}_{t-1} + b_C) \quad (46)$$

$$C_t = f_t \circ C_{t-1} + i_t * \widetilde{C}_t \quad (47)$$

$$o_t = \sigma(w_{xo} \cdot \mathcal{X}_t + w_{ho} * \mathcal{H}_{t-1} + w_{ho} \circ \mathcal{C}_t + b_o) \quad (48)$$

$$h_t = o_t \circ \tanh(C_t) \quad (49)$$

### E. Variant of architectures of RNN with respective to the applications

To incorporate the attention mechanism with RNNs, Word2Vec is used in most of the cases for word or sentence encoding. Word2vec is a powerful word embedding technique with a 2-layer predictive NN from raw text inputs. This approach is used in the different fields of applications including unsupervised learning with words, relationship learning between the different words, the ability to abstract higher meaning of the words based on the similarity, sentence modeling, language understanding and many more. There are different other word embedding approaches that have been proposed in the past few years which are used to solve difficult tasks and provide state-of-the-art performance including machine translation and language modeling, Image and video captioning and time series data analysis [143,144, and 288].

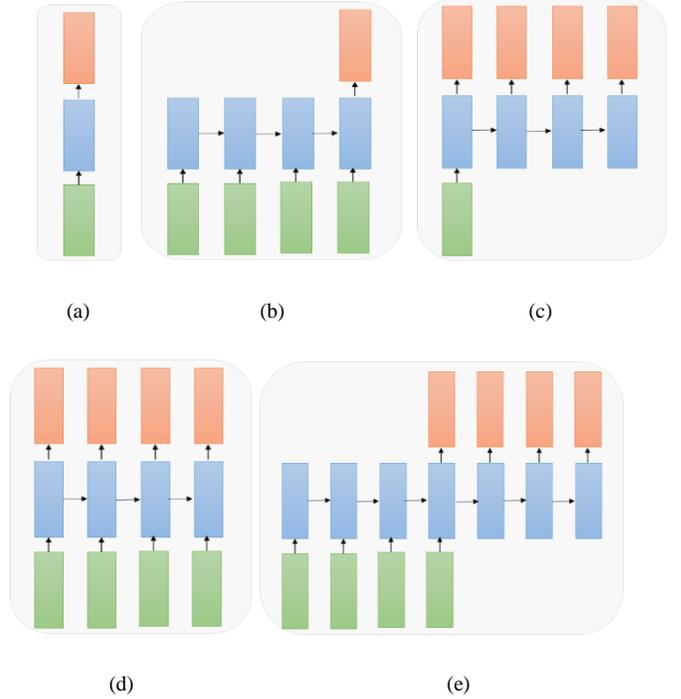

**Fig. 35.** Different structure of RNN with respect to the applications: (a) One to one (b) Many to one (c) One to many (d) Many to many and (e) Many to many

From the application point of view, RNNs can solve different types of problems which need different architectures of RNNs shown in Fig. 35. In Fig. 35, Input vectors are represented as green, RNN states are represented with blue and orange represents the output vector. These structures can be described as:

**One to One**: Standard mode for classification without RNN (e.g. image classification problem) shown Fig. 35 (a)
**Many to One**: Sequence of inputs and single output (e.g. the sentiment analysis where inputs are a set of sentences or words and output is positive or negative expression) shown Fig. 35 (b)
**One to Many**: Where a system takes an input and produces a sequence of outputs (Image Captioning problem: input is a single image and output is a set of words with context) shown Fig. 35 (c).
**Many to Many**: sequences of inputs and outputs (e.g. machine translation: machine takes a sequence of words from English and translates to a sequence of words in French) shown Fig. 35 (d).
**Many to Many**: sequence to sequence learning (e.g. video classification problem in which we take video frames as input and wish to label each frame of the video shown Fig. 35(e).

### F. Attention based models with RNN

Different attention based models have been proposed using RNN approaches. First initiative for RNNs with attention that automatically learns to describe the content of images is proposed by Xu, et al. in 2015 [145]. A dual state attention based RNN is proposed for effective time series prediction [146]. Another difficult task is Visual Question Answering (VQA) using GRUs where the inputs are an image and a natural language question about the image, the task is to provide an accurate natural language answer. The output is to be conditional on both image and textual inputs. A CNN is used to encode the image and a RNN is implemented to encode the sentence [147]. Another, outstanding concept is released from Google called Pixel Recurrent Neural Networks (Pixel RNN). This approach provides state-of-the-art performance for image completion tasks [148]. The new model called residual RNN is proposed, where the RNN is introduced with an effective residual connection in a deep recurrent network [149].

*G. RNN Applications*

RNNs including LSTM and GRU are applied on Tensor processing [150]. Natural Language Processing using RNN techniques including LSTMs and GRUs [151,152]. Convolutional RNNs based on multi-language identification system was been proposed in 2017 [153]. Time series data analysis using RNNs [154]. Recently, TimeNet was proposed based on pre-trained deep RNNs for time series classification (TSC) [155]. Speech and audio processing including LSTMs for large scale acoustic modeling [156,157]. Sound event prediction using convolutional RNNs [158]. Audio tagging using Convolutional GRUs [159]. Early heart failure detection is proposed using RNNs [160].

RNNs are applied in tracking and monitoring: data driven traffic forecasting systems are proposed using Graph Convolutional RNN (GCRNN) [161]. An LSTM based network traffic prediction system is proposed with a neural network based model [162]. Bidirectional Deep RNN is applied for driver action prediction [163]. Vehicle Trajectory prediction using an RNN [164]. Action recognition using an RNN with a Bag-of-Words [165]. Collection anomaly detection using LSTMs for cyber security [166].

## VI. AUTO-ENCODER (AE) AND RESTRICTED BOLTZMANN MACHINE (RBM)

This section will be discussing one of the un-supervised deep learning approaches the Auto Encoder [55] (e.g. variational auto-encoder (VAE) [167], denoising AE [59], sparse AE [168], stacked denoising AE [169], Split-Brain AE [170]). The applications of different AE are also discussed in the end of this chapter.

*A. Review of Auto-Encoder (AE)*

An AE is a deep neural network approach used for unsupervised feature learning with efficient data encoding and decoding. The main objective of auto encoder to learn and representation (encoding) of data, typically for data dimensionality reduction, compression, fusion and many more. This auto encoder technique consists of two parts: the encoder and the decoder. In the encoding phase, the input samples are mapped usually in the lower dimensional features space with a constructive feature representation. This approach can be repeated until the desired feature dimensional space is reached. Whereas in the decoding phase, we regenerate actual features from lower dimensional features with reverse processing. The conceptual diagram of auto-encoder with encoding and decoding phases is shown in Fig. 36.

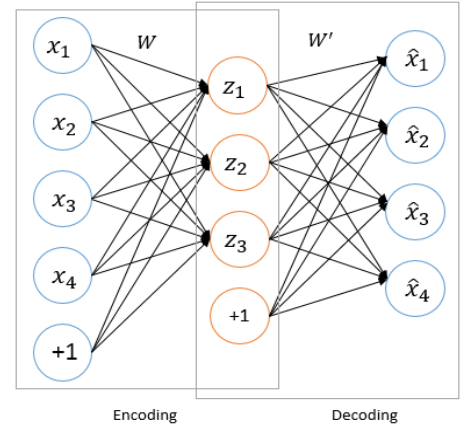

**Fig. 36.** Diagram for Auto encoder.

The encoder and decoder transition can be represented with $\emptyset$ and $\varphi$

$$\emptyset : \mathcal{X} \to \mathcal{F}$$
$$\varphi : \mathcal{F} \to \mathcal{X}$$

$$\emptyset, \varphi = argmin_{\emptyset,\varphi} \|X - (\emptyset, \varphi)X\|^2 \quad (50)$$

If we consider a simple auto encoder with one hidden layer, where the input is $x \in \mathbb{R}^d = \mathcal{X}$, which is mapped onto $\in \mathbb{R}^p = \mathcal{F}$, it can then be expressed as follows:

$$z = \sigma_1(Wx + b) \quad (51)$$

Where W is the weight matrix and b is bias. $\sigma_1$ represents an element wise activation function such as a sigmoid or a rectified linear unit (RLU). Let us consider $z$ is again mapped or reconstructed onto $x'$ which is the same dimension of $x$. The reconstruction can be expressed as

$$x' = \sigma_2(W'z + b') \quad (52)$$

This model is trained with minimizing the reconstruction errors, which is defined as loss function as follows:

$$\mathcal{L}(x, x') = \|x - x'\|^2 = \|x - \sigma_2(W'(\sigma_1(Wx + b)) + b')\|^2 \quad (53)$$

Usually the feature space of $\mathcal{F}$ has lower dimensions then the input feature space $\mathcal{X}$, which can be considered as the compressed representation of the input sample. In the case of multilayer auto encoder, the same operation will be repeated as required with in the encoding and decoding phases. A deep Auto encoder is constructed by extending the encoder and decoder of auto encoder with multiple hidden layers. The Gradient vanishing problem is still a big issue with the deeper model of AE: the gradient becomes too small as it passes back through many layers of a AE model. Different advanced AE models are discussed in the following sections.

*B. Variational auto encoders (VAEs)*

There are some limitations of using simple Generative Adversarial Networks (GAN) which are discussed in Section 7. The limitations are: first, images are generated using GAN from input noise. If someone wants to generate a specific image, then it is difficult to select the specific features (noise) randomly to produce desired images. It requires searching the entire distribution. Second, GANs differentiate between 'real' and 'fake' objects. For example: if you want to generate a dog, there is no constraint that the dog must be look like dog. Therefore, it produces same style images which the style looks like a dog but if we closely observed then it is not exactly. However, VAE is proposed to overcome those limitation of basic GANs, where the latent vector space is used to represent the images which follow a unit Gaussian distribution. [167,174].

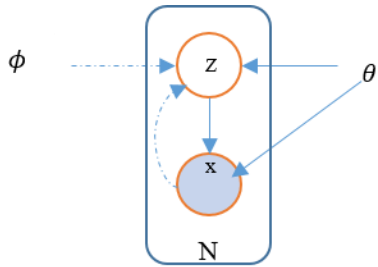

**Fig. 37.** Variational Auto-Encoder.

In this model, there are two losses, one is a mean squared error that determines, how good the network is doing for reconstructing image, and loss (the Kullback-Leibler (KL) divergence) of latent, which determines how closely the latent variable match is with unit Gaussian distribution. For example suppose $x$ is an input and the hidden representation is $z$. The parameters (weights and biases) are $\theta$. For reconstructing the phase the input is $z$ and the desired output is $x$. The parameters (weights and biases) are $\phi$. So, we can represent the encoder as $q_\theta(z|x)$ and decoder $p_\phi(x|z)$ respectively. The loss function of both networks and latent space can be represented as

$$l_i(\theta, \phi) = -E_{z \sim q_\theta(z|x_i)}[log p_\phi(x_i|z)] + KL(q_\theta(z|x_i)| p(z))$$
(54)

*C. Split-Brain Auto-encoder*

Recently Split-Brain AE was proposed from Berkeley AI Research (BAIR) lab, which is the architectural modification of traditional auto encoders for unsupervised representation learning. In this architecture, the network is split into disjoint sub-networks, where two networks try to predict the feature representation of an entire image [170].

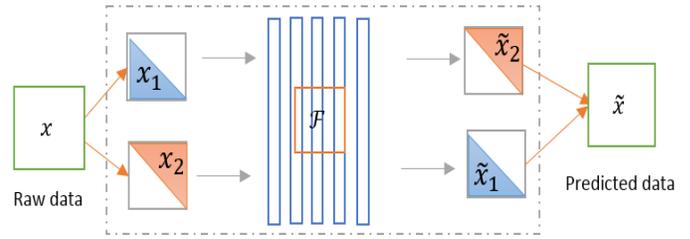

**Fig. 38.** Split-Brain Auto encoder

*D. Applications of AE*

AE is applied in Bio-informatics [102,171] and cyber security [172]. We can apply AE for unsupervised feature extraction and then apply Winner Take All (WTA) for clustering those samples for generating labels [173]. AE has been used as a encoding and decoding technique with or for other deep learning approaches including CNN, DNN, RNN and RL in the last decade. However, here are some other approaches recently published [174,175]

*E. Review of RBM*

Restricted Boltzmann Machines (RBM) are another unsupervised deep learning approach. The training phase can be modeled using a two-layer network called a "Restricted Boltzmann Machine" [176] in which stochastic binary pixels are connected to stochastic binary feature detectors using symmetrically weighted connections. RBM is an energy-based undirected generative model that uses a layer of hidden variables to model distribution over visible variables. The undirected model for the interactions between the hidden and visible variables is used to ensure that the contribution of the likelihood term to the posterior over the hidden variables are approximately factorial which greatly facilitates inference [177]. The conceptual diagram of RBM is shown in Fig. 39.

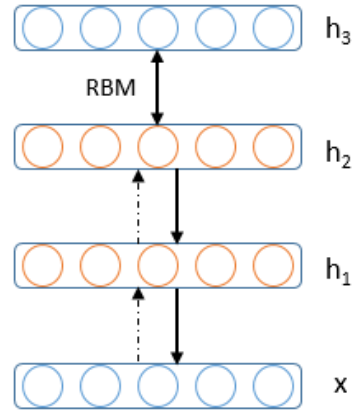

**Fig. 39.** Block diagram for RBM

Energy-based models mean that the probability distribution over the variables of interest is defined through an energy function. The energy function is composed from a set of observable variables s $V = \{v_i\}$ and a set of hidden variables = $\{h_i\}$, where $i$ is a node in the visible layer, $j$ is a node in the hidden layer. It is restricted in the sense that there are no visible-visible or hidden-hidden connections. The values corresponding to "visible" units of the RBM because their states are observed; the feature detectors correspond to "hidden" units. A joint configuration, (v,h) of the visible and hidden units has an energy (Hopfield, 1982) given by:

$$E(v,h) = -\sum_i a_i v_i - \sum_j b_j h_j - \sum_i \sum_j v_i w_{i,j} h_j \quad (55)$$

Where $v_i$ $h_j$ are the binary states of visible unit $i$ and hidden unit $j$, $a_i$, $b_j$ are their biases and $w_{ij}$ is the weight between them. The network assigns a probability to a possible pair of a visible and a hidden vector via this energy function:

$$p(v,h) = \frac{1}{Z} e^{-E(v,h)} \quad (56)$$

where the "partition function", $Z$, is given by summing over all possible pairs of visible and hidden vectors:

$$Z = \sum_{v,h} e^{-E(v,h)} \quad (57)$$

The probability that the network assigns to a visible vector, v, is given by summing over all possible hidden vectors:

$$p(v) = \frac{1}{Z} \sum_h e^{-E(v,h)} \quad (58)$$

The probability that the network assigns to a training sample can be raised by adjusting the weights and biases to lower the energy of that sample and to raise the energy of other samples, especially those have low energies and therefore make a big contribution to the partition function. The derivative of the log probability of a training vector with respect to a weight is surprisingly simple.

$$\frac{\partial \log p(v)}{\partial w_{ij}} = \langle v_i h_j \rangle_{data} - \langle v_i h_j \rangle_{model} \quad (59)$$

Where the angle brackets are used to denote expectations under the distribution specified by the subscript that follows. This leads to a simple learning rule for performing stochastic steepest ascent in the log probability of the training data:

$$w_{ij} = \varepsilon \left( \langle v_i h_j \rangle_{data} - \langle v_i h_j \rangle_{model} \right) \quad (60)$$

Where $\varepsilon$ is a learning rate. Given a randomly selected training image, $v$, the binary state, $h_j$, of each hidden unit, $j$ is set to 1 with probability

$$p(h_j = 1|v) = \sigma(b_j + \sum_i v_i w_{ij}) \quad (61)$$

Where $\sigma(x)$ is the logistic sigmoid function $1/(1 + e^{(-x)})$. $v_i h_j$ is then an unbiased sample. Because there are no direct connections between visible units in an RBM, it is also easy to get an unbiased sample of the state of a visible unit, given a hidden vector

$$p(v_i = 1|h) = \sigma(a_i + \sum_j h_j w_{ij}) \quad (62)$$

Getting an unbiased sample of $\langle v_i h_j \rangle_{model}$ is much more difficult. It can be done by starting at any random state of the visible units and performing alternating Gibbs sampling for a long time. One iteration of alternating Gibbs sampling consists of updating all the hidden units in parallel using Eq. (61) followed by updating all the visible units in parallel using following Eq. (62). A much faster learning procedure was proposed in Hinton (2002). This starts by setting the states of the visible units to a training vector. Then the binary states of the hidden units are all computed in parallel using Eq. (61). Once binary states have been chosen for the hidden units, a "reconstruction" is produced by setting each $v_i$ to 1 with a probability given by Eq. (62). The change in a weight is then given by

$$\Delta w_{ij} = \varepsilon \left( \langle v_i h_j \rangle_{data} - \langle v_i h_j \rangle_{recon} \right) \quad (63)$$

A simplified version of the same learning rule that uses the states of individual units instead of a pairwise products is used for the biases [178]. This approach is mainly used for pre-training a neural network in an un-supervised manner to generate initial weights. One of the most popular deep learning approaches called Deep Belief Network (DBN) is proposed based on this approach. Some of the examples of the applications with RBM and DBN for data encoding, news clustering, and cyber security are shown, for detail see [51, 179,289].

VII. GENERATIVE ADVERSARIAL NETWORKS (GAN)

At the beginning of this chapter, we started with a quote from Yann LeCun, "GAN is the best concept proposed in the last ten years in the field of deep learning (Neural networks)".

*A. Review on GAN*

The concept of generative models in machine learning started a long time before which is used for data modeling with conditional probability density function. Generally, this type of model is considered a probabilistic model with joint probability distribution over observation and target (label) values. However, we did not see big success of this generative model before. Recently deep learning based generative models have become popular and shown enormous success in different application domains.

Deep learning is a data driven technique that performs better as the number of input samples increased. Due to this reason, learning with reusable feature representations from a huge number of un-labels dataset has become an active research area. We mentioned in the introduction that Computer vision has different tasks, segmentation, classification, and detection, which requires large amounts of labelled data. This problem has been attempted to be solved be generating similar samples with a generative model.

Generative Adversarial Networks (GANs) are a deep learning approach recently developed by Goodfellow in 2014. GANs offer an alternative approach to maximum likelihood estimation techniques. GANs are an unsupervised deep learning approach where two neural networks compete against each other in a zero sum game. Each of the two networks gets better at its given task with each iteration. In the case of the image generation problem the generator starts with Gaussian noise to generate images and the discriminator determines how good the generated images are. This process continues until outputs of the generator become close to actual input samples. According to the Fig. 40, it can be considered that Discriminator (D) and Generator (G) two players playing min-max game with the function of V (D, G) which can be expressed as follows according to this paper [180,181].

development. GANs have been used to generate motion in video as well as generate artificial video [182].GANs have two different areas of deep learning that they fall into semi-supervised and unsupervised. Some research in these areas focuses on topology of the GAN architecture to improve functionality and the training approach. Deep convolution GAN (DCGAN) is a convolution based GAN approach proposed in 2015[183].This semi-supervised approach has shown promised results compared to its unsupervised counterpart. The regenerated results according the experiments of DCGAN are shown in the following figures [183]. Fig. 41 shows the output for generated bedroom images after one training pass through the dataset. Most of the figures included in this section are generated through experiments. Theoretically the model could learn to memorize training examples, but this is experimentally unlikely as we train with a small learning rate and mini batches with SGD. We are aware of no prior empirical evidence demonstrating memorization with SGD and a small learning rate [183].

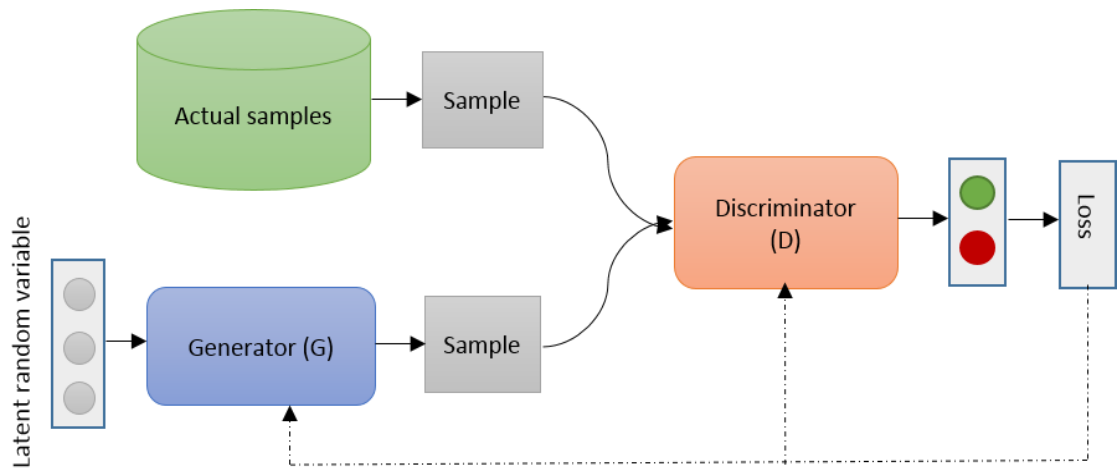

**Fig. 40.** Conceptual diagram for Generative Adversarial Networks (GAN)

$$min_G max_D V(D,G) = \mathbb{E}_{x \sim P_{data}(x)}[log(D(x))] + \mathbb{E}_{z \sim P_{data}(z)}[log(1 - D(G(z)))] \quad (64)$$

In practice, this equation may not provide sufficient gradient for learning G (which started from random Gaussian noise) at the early stages. In the early stages D can reject samples because they are clearly different compared to training samples. In this case, $log(1 - D(G(z)))$ will be saturated. Instead of training G to minimize $log(1 - D(G(z)))$ we can train G to maximize $log(G(z))$ objective function which provides much better gradients in early stages during learning. However, there were some limitations of convergence process during training with the first version. In the beginning state a GAN has some limitations regarding the following issues:
- The lack of a heuristic cost function (as pixel-wise approximate means square errors (MSE))
- Unstable to train (sometimes that can be cause of producing nonsensical outputs)

Research in the area of GANs has been on going with many improved versions being proposed [181]. GANs are able to produce photorealistic images for applications such as visualization of interior or industrial design, shoes, bags, and clothing items. GANs also so extensive use in the field of game

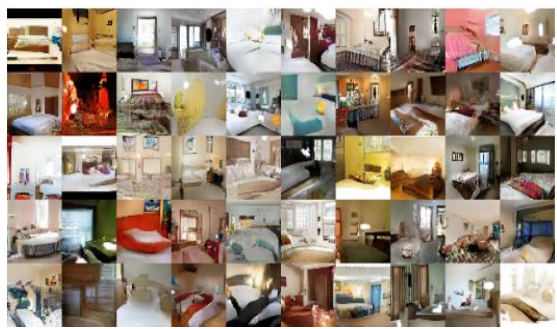

**Fig. 41.** Experimental outputs of bedroom images.

Fig. 42 represents generated bedroom images after five epochs of training. There appears to be evidence of visual under-fitting via repeated noise textures across multiple samples such as the base boards of some of the beds.

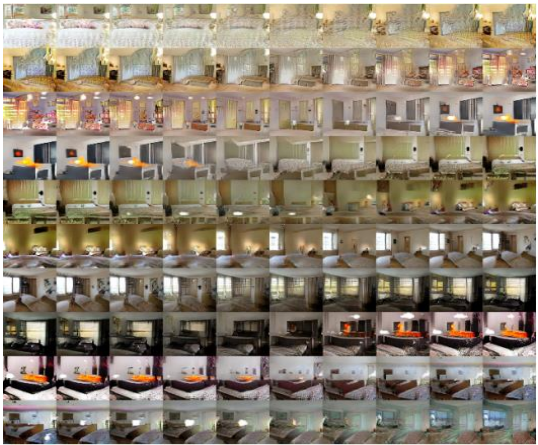

**Fig. 42.** Reconstructed bedroom images using DCGAN

In Fig. 42 the top rows interpolation between a series of 9 random points in Z and show that the space learned has smooth transitions. In every image the space plausibly looks like a bedroom. In the 6th row, you see a room without a window slowly transforming into a room with a giant window. In the 10th row, you see what appears to be a TV slowly being transformed into a window. The following Fig. 43 shows the effective application of latent space vectors. Latent space vectors can be turned into meaning output by first performing addition and subtraction operations followed by a decode. Fig. 43 shows that a man with glasses minus a man and add a women which results in a woman with glasses.

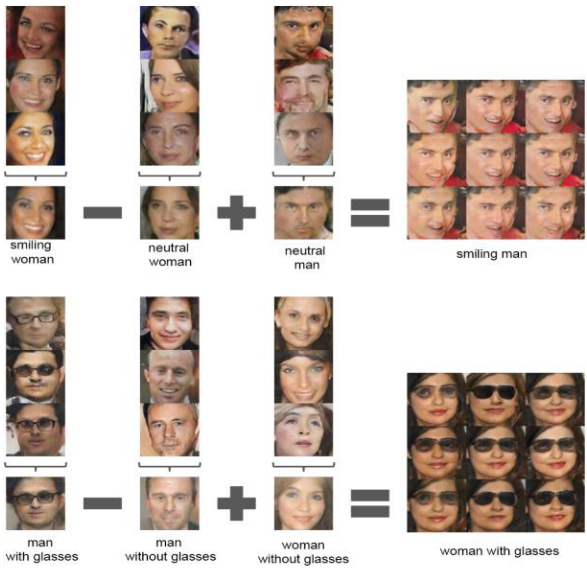

**Fig. 43.** Example of smile arithmetic and arithmetic for wearing glass using GAN

Fig. 44 shows a "turn" vector was created from four averaged samples of faces looking left versus looking right. By adding interpolations along this axis of random samples the pose can be reliably transformed. There are some interesting applications that have been proposed for GANs. For example natural indoor scenes are generated with improved GAN structures. These GANs learn surface normal and are combined with a Style GAN by Wang and Gupta[184]. In this implementation, authors considered style and structure of GAN named ($S^2$-GAN), which generates a surface normal map. this is an improved version of a GAN . In 2016 a information-theoretic extension to the GAN called "InfoGAN" was proposed. An infoGAN can learn with better representations in a completely unsupervised manner . The experimental results show that the unsupervised InfoGAN is competitive with representation learning with the fully supervised learning approach [185].

In 2016, another new architecture was proposed by Im et al., where the recurrent concept is included with the adversarial network during training [186]. Jun et. al. proposed iGANs which allowed image manipulation interactively on a natural image manifold. Image to image translation with conditional adversarial networks is proposed in 2017 [187]. Another improved version of GANs named Coupled Generative Adversarial Network (CoGAN) is a learned joint distribution of multi-domain images. The exiting approach does not need tuples of corresponding images in different domains in the training set [188]. Bidirectional Generative Adversarial Networks (BiGANs are learned with inverse feature mapping, and shown that the resulting learned feature representation is useful for auxiliary supervised discrimination tasks, competitive with contemporary approaches to un-supervised and self-supervised feature learning [189].

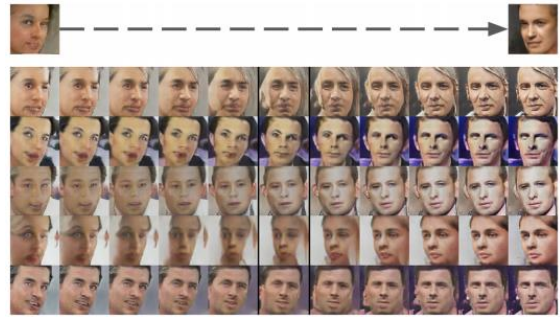

**Fig. 44.** Face generation in different angle using GAN

Recently Google proposed extended versions of GANs called Boundary Equilibrium Generative Adversarial Networks (BEGAN) with a simple but robust architecture. BEGAN has a better training procedure with fast and stable convergence. The concept of equilibrium helps to balance the power of the discriminator against generator. In addition, it can balance the trade-off between image diversity and visual quality [190]. Another similar work is called Wasserstein GAN (WGAN) algorithm that shows significant benefits over traditional GAN [191]. WGANs had two major benefits over traditional GANs. First a WGAN meaningfully correlates the loss metric with the generator's convergence and sample quality. Secondly WGANs have improved stability of the optimization process.
The improved version of WGAN is proposed with a new clipping technique, which penalizes the normal of the gradient of the critic with respect to its inputs [192]. There is promising architecture that has been proposed based on generative models where the images are represented with untrained DNN that give an opportunity for better understanding and visualization of DNNs [193]. Adversarial examples for generative models [194]. Energy-based GAN was proposed by Yann LeCun from Facebook in 2016 [195]. The training process is difficult for GANs, Manifold Matching GAN (MMGAN) proposed with better training process which is experimented on three different datasets and the experimental results clearly demonstrate the efficacy of MMGAN against other models [196]. GAN for geo-statistical simulation and inversion with efficient training approach [197]

Probabilistic GAN (PGAN) which is a new kind of GAN with a modified objective function. The main idea behind this method is to integrate a probabilistic model (A Gaussian Mixture Model) into the GAN framework that supports likelihood rather than classification [198]. A GAN with Bayesian Network model [199]. Variational Auto encode is a popular deep learning approach, which is trained with Adversarial Variational Bayes (AVB) which helps to establish a principle connection between VAE and GAN [200]. The f-GAN which is proposed based on the general feed forward neural network [201]. Markov model based GAN for texture synthesis [202]. Another generative model based on doubly stochastic MCMC method [203]. GAN with multi-Generator [204]

Is an unsupervised GAN capable of learning on a pixel level domain adaptation that transforms in the pixel space from one domain to another domain. This approach provides state-of-the-art performance against several unsupervised domain adaptation techniques with a large margin [205]. A new network is proposed called Schema Network, which is an object oriented generative physics simulator able to disentangle multiple causes of events reasoning through causes to achieve a goal that is learned from dynamics of an environment from data [206]. There is interesting research that has been conducted with a GAN that is to Generate Adversarial Text to Image Synthesis. In this paper, the new deep architecture is proposed for GAN formulation which can take the text description of an image and produce realistic images with respect to the inputs. This is an effective technique for text based image synthesis using a character level text encoder and class conditional GAN. GAN is evaluated on bird and flower dataset first then general text to image which is evaluated on MS COCO dataset [36].

B. Applications of GAN

This learning algorithm has been applied in different domain of applications that is discussed in the following sections:

*1) GAN for image processing*

GANs used for generating photo-realistic image using a super-resolution approach [207]. GAN for semantic segmentation with semi and weakly supervised approach [208]. Text Conditioned Auxiliary Classifier GAN (TAC-GAN) which is used for generating or synthesizing images from a text description [209]. Multi-style Generative network (MSG-Net) which retains the functionality of optimization based approaches with fast speed. This network matches image styles at multiple scale and puts the computational burden into training [210]. Most of the time, vision systems struggle with rain, snow, and fog. A single image de-raining system is proposed using a GAN recently [211].

*2) GAN for speech and audio processing*

An End-to-End Dialogue system using Generative Hierarchical Neural Network models [212]. In addition, GANs have been used in the field of speech analysis. Recently, GANs are used for speech enhancement which is called SEGAN that incorporates further speech-centric design to improve performance progressively [213]. GAN for symbolic-domain and music generation which performs comparably against Melody RNN [214].

*3) GAN for medical information processing*

GANs for Medical Imagining and medical information processing [102], GANs for medical image de-noising with Wasserstein distance and perceptual loss [215]. GANs can also be used for segmentation of Brain Tumors with conditional GANs (cGAN) [216]. A General medical image segmentation approach is proposed using a GAN called SegAN [217]. Before the deep learning revolution compressive sensing is one the hottest topic. However, Deep GAN is used for compresses sensing that automates MRI [218]. In addition, GANs can also be used in health record processing, due to the privacy issue the electronic health record (EHR) is limited to or is not publicly available like other datasets. GANs are applied for synthetic EHR data which could mitigate risk [219]. Time series data generation with Recurrent GAN (RGAN) and Recurrent Conditional GAN (RCGAN) [220]. LOGAN consists of the combination of a generative and discriminative model for detecting the over fitting and recognition inputs. This technique has been compared against state-of-the-art GAN technique including GAN, DCGAN, BEGAN and a combination of DCGAN with a VAE [221].

*4) Other applications*

A new approach called Bayesian Conditional GAN (BC-GAN) which can generate samples from deterministic inputs. This is simply a GAN with Bayesian framework that can handle supervised, semi-supervised and un-supervised learning problems [222,223]. In machine learning and deep learning community, online learning is an important approach. GANs are used for online learning in which it is being trained for finding a mixed strategy in a zero-sum game which is named Checkov GAN 1[224]. Generative moment matching networks based on statistical hypothesis testing called maximum mean discrepancy (MMD) [225]. One of the interesting ideas to replace the discriminator of GAN with two-sample based kernel MMD, which is called MMD-GAN. This approach significantly outperforms Generative moment matching network (GMMN) technique which is an alternative approach for generative model [226]

Pose estimation using a GAN [227]. Photo editing network using a GAN [228]. Anomaly detection [229]. DiscoGAN for learning cross-domain relation with GAN [230]. Single shot learning with GAN [231]. GAN is used for response generation and question answering system [232,233]. Last but not least is WaveNet a generative model that is used to generate audio waveform [286].

## VIII. DEEP REINFORCEMENT LEARNING (DRL)

In the previous sections, we have focused on supervised and unsupervised deep learning approaches including DNN, CNN, RNN including LSTM and GRU, AE, RBM, GAN etc. These types of deep learning approaches are used for prediction, classification, encoding, decoding, data generation, and many more application domains. However, this section demonstrates a survey on Deep Reinforcement Learning (DRL) based on the recently developed methods in this field of RL.

### A. Review on DRL

DRL is a learning approach which learns to act with general sense from unknown real environment (For details please read the following article [234]). RL can be applied in a different scope of field including fundamental Sciences for decision making, Machine learning from a computer science point of view, in the field of engineering and mathematics, optimal control, robotics control, power station control, wind turbines, and Neuroscience the reward strategy is widely studied in last couple of decades. It is also applied in economic utility or game theory for making better decisions and for investment choices. The psychological concept of classical conditioning is how animals learn. Reinforcement learning is a technique for what to do and how-to match a situation to an action. Reinforcement learning is different from supervised learning technique and other kinds of learning approaches studies recently including traditional machine learning, statistical pattern recognition, and ANN.

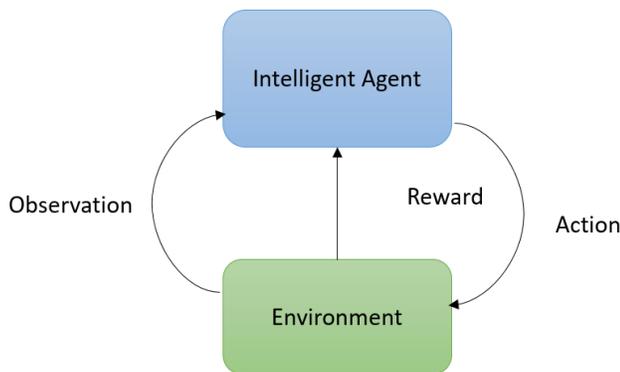

**Fig. 45.** Conceptual diagram for RL system.

Unlike the general supervised and unsupervised machine learning RL is defined not by characterizing learning methods, but by characterizing a learning problem. However, the recent success of DL has had huge impact on the success of DRL which is known as DRL. According to the learning strategy, the RL technique is learned through observation. For observing the environment, the promising DL techniques include CNN, RNN, LSTM, and GRU are used depending upon the observation space. As DL techniques encode data efficiently therefore the following step of action is performed more accurately. According to the action, the agent receives an appropriate reward respectively. As a result, the entire RL approach becomes more efficient to learn and interact in the environment with better performance.

However, the history of the modern DRL revolution began recently from Google Deep Mind in 2013 with Atari games with DRL. Where the agent was evaluated on more than fifty different games. In which the DRL based approaches perform better against the human expert in almost all of the games. In this case, the environment is observed on video frames which are processed using a CNN [235,236]. The success of DRL approaches depend on the level of difficulty of the task attempt to be solved. After a huge success of Alpha-Go and Atari from Google Deep mind, they proposed reinforcement learning environment based on StarCraft II in 2017, which is called SC2LE (StarCraft II Learning Environment) [237]. The SC2LE is a game with multi-agent with multiple player's interactions. This proposed approach has a large action space involving selection and control of hundreds of units. It contains many states to observe from raw feature space and it uses strategies over thousands of steps. The open source python based StarCraft II game engine has been provided free in online.

### B. Q- Learning

There are some fundamental strategies which are essential to know for working with DRL. First, the RL learning approach has a function that calculates the Quality of state-action combination which is called Q-Learning (Q-function).

$$Q: S \times A \to \mathbb{R}$$

The Q-function which is learned from the observation states $S$, action of the states $A$ and reward $\mathbb{R}$. This is an iterative approach to update the values. Q-learning is defined as a model-free reinforcement learning approach which is used to find an optimal action-selection policy for any given (finite) Markov Decision Process (MDP). MDP is a mathematical framework for modeling decision using state, action and rewards. Q-learning only needs to know about the states available and what are the possible actions in each state. Another improved version of Q-Learning known as Bi-directional Q-Learning. In this article, the Q-Learning is discussed, for details on bi-directional Q-Learning please see [238].

At each step s, choose the action which maximize the following function Q (s, a)
- Q is an estimated utility function – it tells us how good an action is given in a certain state
- r (s, a) immediate reward for making an action best utility (Q) for the resulting state

This can be formulated with recursive definition as follows:
$$Q(s,a) = r(s,a) + \gamma \, max_{a'}(Q(s',a')) \qquad (65)$$

This equation is called Bellman's equation, which is the core equation for RL. Here $r(s,a)$ is the immediate reward, $\gamma$ is the relative value of delay vs. immediate rewards [0, 1] $s'$ is the new state after action $a$. The $a$ and $a'$ are an action in sate $s$ and $s'$ respectively. The action is selected based on the following equation:

$$\pi(s) = argmax_a Q(s,a) \qquad (66)$$

In each state, a value is assigned called a Q-value. When we visit a state and we receive a reward accordingly. We use the reward to update the estimated value for that state. As the reward is stochastic, as a result we need to visit the states many times. In addition, it is not guaranteed that we will get same reward ($R_t$) in another episode. The summation of the future rewards in episodic tasks and environments are unpredictable, further in the future, we go further with the reward diversely as expressed.

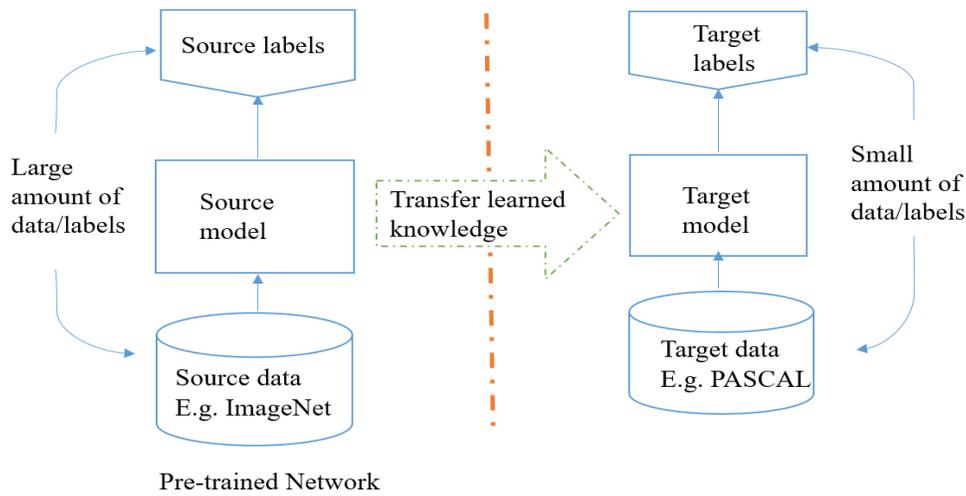

**Fig. 46.** Conceptual diagram for transfer learning: pretrained on ImageNet and transfer learning is used for retraining on PASAL dataset.

$$G_t = R_{t+1} + R_{t+2} + R_{t+3} + \ldots\ldots\ldots + R_T \quad (67)$$

The sum of discounted future rewards in both cases are some factor as scalar.

$$G_t = \gamma R_{t+1} + \gamma^2 R_{t+2} + \gamma^3 R_{t+3} + \ldots\ldots\ldots + \gamma^T R_T \quad (68)$$

Here $\gamma$ is a constant. The more we are in the future, the less we take the reward in to account

**Properties of Q-learning:**
- Convergence of Q-function: approximation will be converged to the true Q-function, but it must visit possible state-action pair infinitely many times.
- The state table size can be vary depending on the observation space and complexity.
- Unseen values are not considered during observation.

The way to fix these problems is to use a neural network (particularly DNN) as an approximation instead of the state table. The inputs of DNN are the state and action and the outputs are numbers between 0 and 1 that represent the utility encoding the states and actions properly. That is the place where the deep learning approaches contribute for making better decisions with respective to the state information. Most of the cases for observing the environment, we use several acquisition devices including camera or other sensing devices for observing the learning environment. For example: if you observed the setup for the challenge of Alpha-Go then it can be seen that the environment, action, and reward are learned based on the pixel values (pixel in action). For details see [235,236].

**Algorithm V: Q-Learning**

**Initialization:**
For each state-action pair $(s, a)$
initialize the table entry $\hat{Q}(s, a)$ to zero

**Steps:**
1. Observed the current state s
2. REPEAT:
   - Select an action $a$ and execute it
   - Received immediate reward $r$
   - Observe the new state $s'$
   - Update the table entry for $\hat{Q}(s, a)$ as follows:

   $$\hat{Q}(s, a) = r + \gamma\, max_{a'}(Q(s', a'))$$
   - $s = s'$

However, it is difficult to develop an agent which can interact or perform well in any observation environment. Therefore, most of the researchers in the field select their action space or environment before training the agent for that environment. The benchmark concept in this case is little bit different compared to supervised or unsupervised deep learning approach. Due to the variety of environments, the benchmark depends on what level of difficulty the environment has been considered compared to the previous or exiting researches? The difficulties depend on the different parameters, number of agents, way of interaction between the agents, number of players and so on.

Recently, another good learning approach has been proposed for DRL [234]. There are many papers published with different networks of DRL including Deep Q-Networks (DQN), Double DQN, Asynchronous methods, policy optimization strategy (including deterministic policy gradient, deep deterministic policy gradient, guided policy search, trust region policy optimization, combining policy gradient and Q-learning) are proposed [234]. Policy Gradient (DAGGER) Super human GO using supervised learning with policy gradient and Monte Carlo tree search with value function [239]. Robotics manipulation using guided policy search [240]. DRL for 3D games using policy gradients [241].

### C. Recent trends of DRL with applications

There is a survey published recently, where basic RL, DRL DQN, trust region policy optimization, and asynchronous advantage actor-critic are proposed. This paper also discusses the advantages of deep learning and focuses on visual understanding via RL and the current trend of research [243]. A network cohesion constrained based on online RL techniques is proposed for health care on mobile devices called mHealth. This system helps similar users to share information efficiently to improve and convert the limited user information into better learned policies [244]. Similar work with the group-driven RL is proposed for health care on mobile device for personalized mHealth Intervention. In this work, K-means clustering is applied for grouping the people and finally shared with RL policy for each group [245]. Optimal policy learning is a challenging task with RL for an agent. Option-Observation Initiation sets (OOIs) allow agents to learn optimal policies in challenging task of POMDPs which are learned faster than RNN [246]. 3D Bin Packing Problem (BPP) is proposed with DRL. The main objective is to place the number of cuboid-shaped item that can minimize the surface area of the bin [247].

The import component of DRL is the reward which is determine based on the observation and the action of the agent. The real-world reward function is not perfect at all times. Due to the sensor error the agent may get maximum reward whereas the actual reward should be smaller. This paper proposed a formulation based on generalized Markov Decision Problem (MDP) called Corrupt Reward MDP [248]. The truest region optimization based deep RL is proposed using recently developed Kronecker-factored approximation to the curvature (K-FAC) [249]. In addition, there is some research that has been conducted in the evaluation of physics experiments using the deep learning approach. This experiment focus agent to learn basic properties such as mass and cohesion of the objects in the interactive simulation environment [250].

Recently Fuzzy RL policies have been proposed that is suitable for continuous state and action space [251]. The important investigation and discussion are made for hyper-parameters in policy gradient for continuous control, general variance of algorithm. This paper also provides a guideline for reporting results and comparison against baseline methods [252]. Deep RL is also applied for high precision assembly tasks [253]. The Bellman equation is one of the main function of RL technique, a function approximation is proposed which ensures that the Bellman Optimality Equation always holds. Then the function is estimated to maximize the likelihood of the observed motion [254]. DRL based hierarchical system is used for could resource allocation and power management in could computing system [255]. A novel Attention-aware Face Hallucination (Attention-FC) is proposed where Deep RL is used for enhancing the quality of image on single patch for images which is applied on face images [256].

## IX. TRANSFER LEARNING

### A. What is transfer learning?

A good way to explain transfer learning is to look at the student teacher relationship. A teacher offers a course after gathering details knowledge regarding that subject. The information will be conveyed through a series of lectures over time. This can be considered that the teacher (expert) is transferring information (knowledge) to the students (learner). The same thing happens in case of deep learning, a network is trained with a big amount data and during the training the model learns the weights and bias. These weights can be transferred to other networks for testing or retraining a similar new model. The network can start with pre trained weights instead of training from scratch.

### B. What is a pre-trained models?

A pre-trained model is a model which is already trained on the same domains as the intended domain. For example for an image recognition task an Inception model already trained on ImageNet can be downloaded. The Inception model can then be used for a different recognition task, and instead of training it from scratch the weights can be left as is with some learned features. This method of training is useful when there is a lack of sample data. There are a lot of pre-trained models available (including VGG, ResNet, and Inception Net on different datasets) in model-zoo from the following link: https://github.com/BVLC/caffe/wiki/Model-Zoo.

### C. Why will you use pre-trained models?

There are a lot of reasons for using pre-trained models. Firstly it is requires a lot of expensive computation power to train big models on big datasets. Secondly it can take up to multiple weeks to train big models. Training new models with pre trained weights can speed up convergence as well as help the network generalization.

### D. How will you use pre-trained models?

We need to consider the following criterions with respective application domains and size of the dataset when using the pre-trained weights which is shown in Table III.

**TABLE III.** CRITERIONS NEED TO BE CONSIDERED FOR TRANSFER LEARNING.

|  | New dataset but **small** | New dataset but **large** |
|---|---|---|
| Pre-trained model on **similar** but new dataset | Freeze weights and train linear classifier from top level features | Fine-tune all the layers (pre-train for faster convergence and better generalization) |
| Pre-trained model on **different** but new dataset | Freeze weights and train linear classifier from non-top-level features | Fine-tune all the layers (pre-train for enhanced convergence speed) |

### E. Working with inference

Research groups working specifically on inference applications look into optimization approaches that include model compression. Model compression is important in the realm of mobile devices or special purpose hardware because it makes models more energy efficient as well as faster.

*F. Myth about Deep Learning*

There is a myth; do you need a million labelled samples for training a deep learning model? The answer is yes but in most cases the transfer leaning approach is used to train deep leaning approaches without having large amounts of label data. For example: the following Fig. 46 demonstrates the strategy for the transfer learning approach in details. Here the primary model has been trained with large amount of labeled data which is ImageNet and then the weights are used to train with the PASCAL dataset. The actual reality is:
- Possible to learn useful representations from unlabeled data.
- Transfer learning can help learned representation from the related task [257].

We can take a trained network for a different domain which can be adapted for any other domain for the target task [258, 589]. First training a network with a close domain for which it is easy to get labeled data using standard back propagation for example: ImageNet classification, pseudo classes from augmented data. Then cut of the top layers of network and replace with supervised objective for target domain. Finally, tune the network using back propagation with labels for target domain until validation loss starts to increase [258, 589]. There are some survey papers and books that are published on transfer learning [260,261]. Self-taught learning with transfer learning [262]. Boosting approach for transfer learning [263].

## X. ENERGY EFFICIENT APPROACHES AND HARDWIRES FOR DL

*A. Overview*

DNNs have been successfully applied and achieved better recognition accuracies in different application domains such as Computer vision, speech processing, natural language processing, big data problem and many more. However, most of the cases the training is being executed on Graphic Processing Units (GPU) for dealing with big volumes of data which is expensive in terms of power.

Recently researchers have been training and testing with deeper and wider networks to achieve even better classification accuracy to achieve human or beyond human level recognition accuracy in some cases. While the size of the neural network is increasing, it becomes more powerful and provides better classification accuracy. However, the storage consumption, memory bandwidth and computational cost are increasing exponentially. On the other hand, these types of massive scale implementation with large numbers of network parameters is not suitable for low power implementation, unmanned aerial vehicle (UAV), different medical devices, low memory system such as mobile devices, Field Programmable Gate Array (FPGA) and so on.

There is much research going on to develop better network structures or networks with lower computation cost, less numbers of parameters for low-power and low-memory systems without lowering classification accuracy. There are two ways to design efficient deep network structure:
- The first approach is to optimize the internal operational cost with an efficient network structure,
- Second design a network with low precision operations or a hardware efficient network.

The internal operations and parameters of a network structure can be reduced by using low dimensional convolution filters for convolution layers. [260].

There are lot of benefit of this approach, **first** the convolutional with rectification operations makes the decision more discriminative. **Second**, the main benefit of this approach is to reduce the number of computation parameters drastically. For example: if one layer has 5x5 dimensional filters which can be replaced with two 3x3 dimensional filters (without pooling layer in between then) for better feature learning; three 3x3 dimensional filter can be used as a replacement of 7x7 dimensional filters and so on. Benefits of using lower dimensional filter is that assuming both the present convolutional layer has C channels, for three layers for 3x3 filter the total number of parameters are weights: 3*(3*3*C*C) =$27C^2$ weights, whereas in case of 7x7 filters, the total number of parameters are (7*7*C*C) =$49C^2$ , which is almost double compared to the three 3x3 filter parameters. Moreover, placement of layers such as convolutional, pooling, drop-out in the network in different intervals has an impact on overall classification accuracy. There are some strategies that are mentioned to optimize the network architecture recently to design efficient deep learning models[89] [264]. According to the paper [89], **Strategy 1**: Replace 3x3 filter with 1x1 filters. The main reason to use lower dimension filter to reduce the overall number of parameter. By replacing 3x3 filters with 1x1 can be reduce 9x number of parameters.

**Strategy 2**: Decrease the number of input channels to 3x3 filters. For a layer, the size of the output feature maps are calculated which is related to the network parameters using $\frac{N-F}{S} + 1$, where N is input map's size, F is filter size, S is for strides. To reduce the number of parameters, it is not only enough to reduce the size of the filters but also it requires to control number of input channels or feature dimension.

**Strategy 3**: Down-sample late in the network so that convolution layers have activation maps: The outputs of present convolution layers can be at least 1x1 or often larger than 1x1. The output width and height can be controlled by some criterions: (1) the size of the input sample (e.g. 256x256) and (2) Choosing the post down sample layer. Most commonly pooling layers are such as average or max pooling layer are used, there is an alternative sub-sampling layer with convolution (3x3 filters) and stride with 2. If most of the earlier layers have larger stride, then most of layers will have small numbers of activation maps. On the other hand, if most of the layers have a stride of 1, and the stride larger than one applied in the end of the network, then many layers of network will have large activation maps. One intuition is the larger activation maps (due to delayed down-sampling) can lead to higher classification accuracy [89]. This intuition has been investigated by K. He and H. Sun applied delayed down-sampling into four different architecture of CNNs, and it is observed that each case delayed down-sampling led to higher classification accuracy [265].

*B. Binary or ternary connect Neural Networks*

The computation cost can be reduced drastically with low precision of multiplication and few multiplications with drop connection [266, 267]. These papers also introduced on Binary Connect Neural Networks (BNN) Ternary Connect Neural Networks (TNN). Generally, multiplication of a real-valued weight by a real-valued activation (in the forward propagations) and gradient calculation (in the backward propagations) are the main operations of deep neural networks. Binary connect or BNN is a technique that eliminates the multiplication operations by converting the weights used in the forward propagation to be binary, i.e. constrained to only two values (0 and 1 or -1 and 1). As a result, the multiplication operations can be performed by simple additions (and subtractions) and makes the training process faster. There are two ways to convert real values to its corresponding binary values such as deterministic and stochastic. In case of deterministic technique, straightforward thresholding technique is applied on weights. An alternative way to do that is stochastic approach where a matrix is converted to binary based on probability where the *"hard sigmoid"* function is used because it is computationally inexpensive. The experimental result shows significantly good recognition accuracy [268,269,270]. There are several advantages of BNN as follows:

- It is observed that the binary multiplication on GPU is almost seven times faster than traditional matrix multiplication on GPU
- In forward pass, BNNs drastically reduce memory size and accesses, and replace most arithmetic operation with bit-wise operations, which lead great increase of power efficiency
- Binarized kernels can be used in CNNs which can reduce around 60% complexity of dedicated hardware.
- It is also observed that memory accesses typically consume more energy compared to arithmetic operation and memory access cost increases with memory size. BNNs are beneficial with respect to both aspects.

There are some other techniques that have been proposed in last few years [271,272,273]. Another power efficient and hardware friendly network structure has been proposed for a CNN with XNOR operations. In XNOR based CNN implementations, both the filters and input to the convolution layer is binary. This result about 58x faster convolutional operations and 32x memory saving. In the same paper, Binary-Weight-Networks was proposed which saved around 32x memory saving. That makes it possible to implement state-of-the-art networks on CPU for real time use instead of GPU. These networks are tested on the ImageNet dataset and provide only 2.9% less classification accuracy than full-precision AlexNet (in top-1% measure). This network requires less power and computation time. This could make it possible to accelerate the training process of deep neural network dramatically for specialized hardware implementation [274]. For the first time, Energy Efficient Deep Neural Network (EEDN) architecture was proposed for neuromorphic system in 2016. In addition, they released a deep learning framework called EEDN, which provides close accuracy to the state-of-the art accuracy almost all the popular benchmarks except ImageNet dataset [275,276].

## XI. HARDWARE FOR DL

Along with the algorithmic development of DL approaches, there are many hardware architectures have been proposed in past few years. The details about present trends of hardware for deep learning have been published recently [277]. MIT proposed "Eyeriss" as a hardware for deep convolutional neural networks (DCNN) [278]. There is another architecture for machine learning called "Dadiannao" [279]. In 2016, an efficient hardware that works for inference was released and proposed by Stanford University called Efficient Inference Engine (EIE) [281]. Google developed a hardware named Tensor Processing Unit (TPU) for deep learning and was released in 2017[280]. IBM released a neuromorphic system called "TrueNorth" in 2015 [275].

Deep learning approaches are not limited to the HPC platform, there are a lot of application already developed which run on mobile devices. Mobile platforms provide data that is relevant to everyday activities of the user, which can make a mobile system more efficient and robust by retraining the system with collected data. There is some research ongoing to develop hardware friendly algorithms for DL [282,283,284].

## XII. FRAMEWORKS AND SDK

Most of the time people use different deep learning frameworks and Standard Development Kits (SDKs) for implementing deep learning approaches which are listed below:

*A. Frameworks*

- Tensorflow : https://www.tensorflow.org/
- Caffe : http://caffe.berkeleyvision.org/
- KERAS : https://keras.io/
- Theano : http://deeplearning.net/software/theano/
- Torch : http://torch.ch/
- PyTorch : http://pytorch.org/
- Lasagne : https://lasagne.readthedocs.io/en/latest/
- DL4J ( DeepLearning4J) : https://deeplearning4j.org/
- Chainer :  http://chainer.org/
- DIGITS : https://developer.nvidia.com/digits
- CNTK (Microsoft)
  : https://github.com/Microsoft/CNTK
- MatConvNet : http://www.vlfeat.org/matconvnet/
- MINERVA : https://github.com/dmlc/minerva
- MXNET : https://github.com/dmlc/mxnet
- OpenDeep : http://www.opendeep.org/
- PuRine : https://github.com/purine/purine2
- PyLerarn2
  : http://deeplearning.net/software/pylearn2/
- TensorLayer:
  https://github.com/zsdonghao/tensorlayer
- LBANN :  https://github.com/LLNL/lbann

*B. SDKs*

- cuDNN : https://developer.nvidia.com/cudnn
- TensorRT : https://developer.nvidia.com/tensorrt
- DeepStreamSDK : https://developer.nvidia.com/deepstream-sdk
- cuBLAS : https://developer.nvidia.com/cublas
- cuSPARSE : http://docs.nvidia.com/cuda/cusparse/
- NCCL : https://devblogs.nvidia.com/parallelforall/fast-multi-gpu-collectives-nccl/

### XIII. BENCHMARK DATABASES

Here is the list of benchmark datasets that are used often to evaluate deep learning approaches in different domains of application:

*A. Image classification or detection or segmentation*

List of datasets are used in the field of image processing and computer vision:

- MNIST : http://yann.lecun.com/exdb/mnist/
- CIFAR 10/100 : https://www.cs.toronto.edu/~kriz/cifar.html
- SVHN/ SVHN2 : http://ufldl.stanford.edu/housenumbers/
- CalTech 101/256 : http://www.vision.caltech.edu/Image_Datasets/Caltech101/
- STL-10 : https://cs.stanford.edu/~acoates/stl10/
- NORB : http://www.cs.nyu.edu/~ylclab/data/norb-v1.0/
- SUN-dataset : http://groups.csail.mit.edu/vision/SUN/
- ImageNet : http://www.image-net.org/
- National Data Science Bowl Competition : http://www.datasciencebowl.com/
- COIL 20/100 : http://www.cs.columbia.edu/CAVE/software/softlib/coil-20.php http://www.cs.columbia.edu/CAVE/software/softlib/coil-100.php
- MS COCO DATASET : http://mscoco.org/
- MIT-67 scene dataset : http://web.mit.edu/torralba/www/indoor.html
- Caltech-UCSD Birds-200 dataset :http://www.vision.caltech.edu/visipedia/CUB-200-2011.html
- Pascal VOC 2007 dataset : http://host.robots.ox.ac.uk/pascal/VOC/voc2007/
- H3D Human Attributes dataset : https://www2.eecs.berkeley.edu/Research/Projects/CS/vision/shape/poselets/
- Face recognition dataset: http://vis-www.cs.umass.edu/lfw/
- For more data-set visit : https://www.kaggle.com/
- http://homepages.inf.ed.ac.uk/rbf/CVonline/Imagedbase.htm
- Recently Introduced Datasets in Sept. 2016:
- Google Open Images (~9M images) – https://github.com/openimages/dataset
- Youtube-8M (8M videos: https://research.google.com/youtube8m/

*B. Text classification*

- Reuters-21578 Text Categorization Collection :
- 
  http://kdd.ics.uci.edu/databases/reuters21578/reuters21578.html
- Sentiment analysis from Stanford : http://ai.stanford.edu/~amaas/data/sentiment/
- Movie sentiment analysis from cornel :
- 
  http://www.cs.cornell.edu/people/pabo/movie-review-data/

*C. Language modeling*

- free eBooks: https://www.gutenberg.org/
- Brown and stanford corpus on present americal english:
  - https://en.wikipedia.org/wiki/Brown_Corpus
- Google 1Billion word corpus: https://github.com/ciprian-chelba/1-billion-word-language-modeling-benchmark

*D. Image Captioning*

- Flickr 8k: http://nlp.cs.illinois.edu/HockenmaierGroup/8k-pictures.html
- Flickr 30k :
- Common Objects in Context (COCO) : http://cocodataset.org/#overview
- http://sidgan.me/technical/2016/01/09/Exploring-Datasets

*E. Machine translation*

- Pairs of sentences in English and French: https://www.isi.edu/natural-language/download/hansard/
- European Parliament Proceedings parallel Corpus 196-2011 : http://www.statmt.org/europarl/
- The statistics for machine translation: http://www.statmt.org/

*F. Question Answering*

- Stanford Question Answering Dataset (SQuAD): https://rajpurkar.github.io/SQuAD-explorer/
- Dataset from DeepMind : https://github.com/deepmind/rc-data
- Amazon dataset: http://jmcauley.ucsd.edu/data/amazon/qa/
- http://trec.nist.gov/data/qamain...

- http://www.ark.cs.cmu.edu/QA-data/
- http://webscope.sandbox.yahoo.co...
- http://blog.stackoverflow.com/20..

G. *Speech Recognition*

- TIMIT : https://catalog.ldc.upenn.edu/LDC93S1
- Voxforge: http://voxforge.org/
- Open Speech and Language Resources: http://www.openslr.org/12/

H. *Document summarization*

- https://archive.ics.uci.edu/ml/datasets/Legal+Case+Reports
- http://www-nlpir.nist.gov/related_projects/tipster_summac/cmp_lg.html
- https://catalog.ldc.upenn.edu/LDC2002T31

I. *Sentiment analysis:*

- IMDB dataset: http://www.imdb.com/

In addition, there is another alternative solution in data programming that labels subsets of data using weak supervision strategies or domain heuristics as labeling functions even if they are noisy and may conflict samples [87].

XIV. JOURNAL AND CONFERENCES

In general, researchers publish their primary version of research on the ArXiv ( https://arxiv.org/ ). Most of the conferences have been accepting papers on Deep learning and its related field. Popular conferences are listed below:

A. *Conferences*
- Neural Information Processing System (NIPS)
- International Conference on Learning Representation (ICLR): What are you doing for Deep Learning?
- International Conference on Machine Learning(ICML)
- Computer Vision and Pattern Recognition (CVPR): What are you doing with Deep Learning?
- International Conference on Computer Vision (ICCV)
- European Conference on Computer Vision (ECCV)
- British Machine Vision Conference (BMVC)

B. *Journal*
- Journal of Machine Learning Research (JMLR)
- IEEE Transaction of Neural Network and Learning System (
- IEEE Transactions on Pattern Analysis and Machine Intelligence (TPAMI)
- Computer Vision and Image Understanding (CVIU)
- Pattern Recognition Letter
- Neural Computing and Application

C. *Tutorials on deep learning*
- http://deeplearning.net/tutorial/
- http://deeplearning.stanford.edu/tutorial/
- http://deeplearning.net/tutorial/deeplearning.pdf
- Courses on Reinforcement Learning: http://rll.berkeley.edu/deeprlcourse/

D. *Books on deep learning*
- *https://github.com/HFTrader/DeepLearningBookhttps://github.com/janishar/mit-deep-learning-book-pdf*
- http://www.deeplearningbook.org/

XV. CONCLUSIONS AND FUTURE WORKS

In this report, we have provide an in-depth review of deep learning and its applications over past few years. We have reviewed different state-of-the-art deep learning models in different categories of learning including supervised, un-supervised, and Reinforcement Learning (RL), as well as their applications in different domains. In addition, we have explained in detail the different supervised deep learning techniques including DNN, CNN, and RNN. We have also reviewed un-supervised deep learning techniques including AE, RBM, and GAN. In the same section, we have considered and explained unsupervised learning techniques which are proposed based on LSTM and RL. In Section 8, we presented a survey on Deep Reinforcement Learning (DRL) with the fundamental learning technique called Q-Learning. Furthermore, we have conducted a survey on energy efficient deep learning approaches, transfer learning with DL, and hardware development trends of DL. Moreover, we have discussed some DL frameworks and benchmark datasets, which are often used for the implementation and evaluation of deep learning approaches. Finally, we have included the relevant journals and conferences, where the DL community has been publishing their valuable research articles.

REFERENCES

[1] Jump, Schmidhuber, J. (2015). "Deep Learning in Neural Networks: An Overview". Neural Networks. **61**: 85–117.
[2] Bengio, Yoshua; LeCun, Yann; Hinton, Geoffrey (2015). "Deep Learning". Nature. **521**: 436–444. doi:10.1038/nature14539.
[3] Bengio, Y.; Courville, A.; Vincent, P. (2013). "Representation Learning: A Review and New Perspectives". IEEE Transactions on Pattern Analysis and Machine Intelligence. **35** (8): 1798–1828
[4] Bengio, Yoshua. "Learning deep architectures for AI." *Foundations and trends® in Machine Learning* 2.1 (2009): 1-127.
[5] Mnih, Volodymyr, et al. "Human-level control through deep reinforcement learning." *Nature* 518.7540 (2015): 529-533.
[6] Mnih, Volodymyr, et al. "Playing Atari with deep reinforcement learning." *arXiv preprint arXiv:1312.5602* (2013).
[7] Krizhevsky, A., Sutskever, I., and Hinton, G. E. ImageNet classification with deep convolutional neural networks. In NIPS, pp. 1106–1114, 2012.

[8] Zeiler, M. D. and Fergus, R. Visualizing and understanding convolutional networks. CoRR, abs/1311.2901, 2013. Published in Proc. ECCV, 2014.
[9] Simonyan, Karen, and Andrew Zisserman. " deep convolutional networks for large-scale image recognition." *arXiv preprint arXiv:1409.1556*(2014).
[10] Szegedy, Christian, et al. "Going deeper with convolutions." Proceedings of the IEEE conference on computer vision and pattern recognition. 2015.
[11] He, Kaiming, et al. "Deep residual learning for image recognition." Proceedings of the IEEE conference on computer vision and pattern recognition. 2016.
[12] Canziani, Alfredo, Adam Paszke, and Eugenio Culurciello. "An analysis of deep neural network models for practical applications." *arXiv preprint arXiv:1605.07678* (2016).
[13] G. Zweig, "Classification and recognition with direct segment models," in Proc. ICASSP. IEEE, 2012, pp. 4161– 4164.
[14] Y. He and E. Fosler-Lussier, "Efficient segmental conditional random fields for phone recognition," in Proc. INTERSPEECH, 2012, pp. 1898–1901.
[15] O. Abdel-Hamid, L. Deng, D. Yu, and H. Jiang, "Deep segmental neural networks for speech recognition." in Proc. INTERSPEECH, 2013, pp. 1849–1853.
[16] H. Tang, W. Wang, K. Gimpel, and K. Livescu, "Discriminative segmental cascades for feature-rich phone recognition," in Proc. ASRU, 2015.
[17] Song, William, and Jim Cai. "End-to-end deep neural network for automatic speech recognition." (2015): 1. (Errors: 21.1)
[18] Deng, Li, Ossama Abdel-Hamid, and Dong Yu. "A deep convolutional neural network using heterogeneous pooling for trading acoustic invariance with phonetic confusion." *Acoustics, Speech and Signal Processing (ICASSP), 2013 IEEE International Conference on*. IEEE, 2013.
[19] Graves, A.-R. Mohamed, and G. Hinton, "Speech recognition with deep recurrent neural networks," in Proc. ICASSP. IEEE, 2013, pp. 6645–6649.
[20] Zhang, Ying, et al. "Towards end-to-end speech recognition with deep convolutional neural networks." *arXiv preprint arXiv:1701.02720* (2017).
[21] Deng, Li, and John Platt. "Ensemble deep learning for speech recognition." (2014).
[22] J. K. Chorowski, D. Bahdanau, D. Serdyuk, K. Cho, and Y. Bengio, "Attention-based models for speech recognition," in Advances in Neural Information Processing Systems, 2015, pp. 577–585.
[23] Lu, Liang, et al. "Segmental recurrent neural networks for end-to-end speech recognition." *arXiv preprint arXiv:1603.00223* (2016).
[24] Van Essen, Brian, et al. "LBANN: Livermore big artificial neural network HPC toolkit." *Proceedings of the Workshop on Machine Learning in High-Performance Computing Environments*. ACM, 2015.
[25] Chen, Xue-Wen, and Xiaotong Lin . "Big Data Deep Learning: Challenges and Perspectives" IEEE Access in date of publication May 16, 2014.
[26] Zhou, Zhi-Hua, et al. "Big data opportunities and challenges: Discussions from data analytics perspectives [discussion forum]." *IEEE Computational Intelligence Magazine* 9.4 (2014): 62-74.
[27] Najafabadi, Maryam M., et al. "Deep learning applications and challenges in big data analytics." *Journal of Big Data* 2.1 (2015): 1.
[28] Goodfellow, Ian, et al. "Generative adversarial nets." *Advances in neural information processing systems*. 2014.
[29] Kaiser, Lukasz, et al. "One Model To Learn Them All." *arXiv preprint arXiv:1706.05137* (2017).
[30] Collobert, Ronan, and Jason Weston. "A unified architecture for natural language processing: Deep neural networks with multitask learning." *Proceedings of the 25th international conference on Machine learning*. ACM, 2008.
[31] Johnson, Melvin, et al. "Google's multilingual neural machine translation system: enabling zero-shot translation." *arXiv preprint arXiv:1611.04558* (2016).
[32] Argyriou, Andreas, Theodoros Evgeniou, and Massimiliano Pontil. "Multi-task feature learning." *Advances in neural information processing systems*. 2007.
[33] Singh, Karamjit, et al. "Deep Convolutional Neural Networks for Pairwise Causality." *arXiv preprint arXiv:1701.00597* (2017).
[34] Yu, Haonan, et al. "Video paragraph captioning using hierarchical recurrent neural networks." *Proceedings of the IEEE conference on computer vision and pattern recognition*. 2016.
[35] Kim, Taeksoo, et al. "Learning to discover cross-domain relations with generative adversarial networks." *arXiv preprint arXiv:1703.05192* (2017).
[36] Reed, Scott, et al. "Generative adversarial text to image synthesis." *arXiv preprint arXiv:1605.05396* (2016).
[37] Deng, Li, and Dong Yu. "Deep learning: methods and applications." *Foundations and Trends® in Signal Processing* 7.3–4 (2014): 197-387.
[38] Gu, Jiuxiang, et al. "Recent advances in convolutional neural networks." *arXiv preprint arXiv:1512.07108* (2015).
[39] Sze, Vivienne, et al. "Efficient processing of deep neural networks: A tutorial and survey." *arXiv preprint arXiv:1703.09039* (2017).
[40] Li, Yuxi. "Deep reinforcement learning: An overview." *arXiv preprint arXiv:1701.07274* (2017).
[41] Kober, Jens, J. Andrew Bagnell, and Jan Peters. "Reinforcement learning in robotics: A survey." *The International Journal of Robotics Research* 32.11 (2013): 1238-1274.
[42] Pan, Sinno Jialin, and Qiang Yang. "A survey on transfer learning." *IEEE Transactions on knowledge and data engineering*22.10 (2010): 1345-1359.
[43] Schuman, Catherine D., et al. "A Survey of Neuromorphic Computing and Neural Networks in Hardware." *arXiv preprint arXiv:1705.06963* (2017).
[44] McCulloch, Warren S., and Walter Pitts. "A logical calculus of the ideas immanent in nervous activity." *The bulletin of mathematical biophysics* 5.4 (1943): 115-133.
[45] Rosenblatt, Frank. "The perceptron: A probabilistic model for information storage and organization in the brain." *Psychological review* 65.6 (1958): 386.
[46] Minsky, Marvin, and Seymour Papert. "Perceptrons." (1969).
[47] Ackley, David H., Geoffrey E. Hinton, and Terrence J. Sejnowski. "A learning algorithm for Boltzmann machines." *Cognitive science* 9.1 (1985): 147-169.
[48] Fukushima, Kunihiko. "Neocognitron: A hierarchical neural network capable of visual pattern recognition." *Neural networks* 1.2 (1988): 119-130.
[49] LeCun, Yann, et al. "Gradient-based learning applied to document recognition." *Proceedings of the IEEE* 86.11 (1998): 2278-2324.
[50] Hinton, Geoffrey E., Simon Osindero, and Yee-Whye Teh. "A fast learning algorithm for deep belief nets." *Neural computation* 18.7 (2006): 1527-1554.
[51] Hinton, Geoffrey E., and Ruslan R. Salakhutdinov. "Reducing the dimensionality of data with neural networks." *science* 313.5786 (2006): 504-507.
[52] Bottou, Léon. "Stochastic gradient descent tricks." *Neural networks: Tricks of the trade*. Springer Berlin Heidelberg, 2012. 421-436.
[53] Rumelhart, David E., Geoffrey E. Hinton, and Ronald J. Williams. "Learning representations by back-propagating errors." *Cognitive modeling* 5.3 (1988): 1.
[54] Sutskever, Ilya, et al. "On the importance of initialization and momentum in deep learning." *International conference on machine learning*. 2013.
[55] Yoshua Bengio, Pascal Lamblin, Dan Popovici and Hugo Larochelle, Greedy Layer-Wise Training of Deep Network, in J. Platt et al. (Eds), Advances in Neural Information Processing Systems 19 (NIPS 2006), pp. 153-160, MIT Press, 2007
[56] Erhan, Dumitru, et al. "The difficulty of training deep architectures and the effect of unsupervised pre-training." *Artificial Intelligence and Statistics*. 2009.
[57] Mohamed, Abdel-rahman, George E. Dahl, and Geoffrey Hinton. "Acoustic modeling using deep belief networks,"Audio, Speech, and Language Processing, IEEE Transactions on 20.1 (2012): 14-22
[58] V. Nair and G. Hinton, Rectified linear units improve restricted boltzmann machines. Proceedings of the 27th International Conference on Machine Learning (ICML-10). 2010.
[59] P. Vincent, H. Larochelle, Y. Bengio, and P.-A. Manzagol, "Extracting and composing robust features with denoising autoencoders," Proceedings of the Twenty-fifth International Conference on Machine Learning, pp. 1096–1103, 2008.
[60] Lin, Min, Qiang Chen, and Shuicheng Yan. "Network in network." *arXiv preprint arXiv:1312.4400* (2013).
[61] Springenberg, Jost Tobias, et al. "Striving for simplicity: The all convolutional net." *arXiv preprint arXiv:1412.6806* (2014).
[62] Huang, Gao, et al. "Densely connected convolutional networks." *arXiv preprint arXiv:1608.06993* (2016).

[63] Larsson, Gustav, Michael Maire, and Gregory Shakhnarovich. "FractalNet: Ultra-Deep Neural Networks without Residuals." *arXiv preprint arXiv:1605.07648* (2016).
[64] Szegedy, Christian, Sergey Ioffe, and Vincent Vanhoucke. "Inception-v4, inception-resnet and the impact of residual connections on learning." *arXiv preprint arXiv:1602.07261* (2016).
[65] Szegedy, Christian, et al. "Rethinking the inception architecture for computer vision." *arXiv preprint arXiv:1512.00567* (2015).
[66] Zagoruyko, Sergey, and Nikos Komodakis. "Wide Residual Networks." *arXiv preprint arXiv:1605.07146* (2016).
[67] Xie, S., Girshick, R., Dollár, P., Tu, Z., & He, K. (2016). Aggregated residual transformations for deep neural networks. *arXiv preprint arXiv:1611.05431*
[68] Veit, Andreas, Michael J. Wilber, and Serge Belongie. "Residual networks behave like ensembles of relatively shallow networks." *Advances in Neural Information Processing Systems*. 2016.
[69] Abdi, Masoud, and Saeid Nahavandi. "Multi-Residual Networks: Improving the Speed and Accuracy of Residual Networks." *arXiv preprint arXiv:1609.05672* (2016).
[70] Zhang, Xingcheng, et al. "Polynet: A pursuit of structural diversity in deep networks." *arXiv preprint arXiv:1611.05725* (2016).
[71] Ren, Shaoqing, et al. "Faster R-CNN: Towards real-time object detection with region proposal networks." *Advances in neural information processing systems*. 2015.
[72] Chollet, François. "Xception: Deep Learning with Depthwise Separable Convolutions." *arXiv preprint arXiv:1610.02357* (2016).
[73] Liang, Ming, and Xiaolin Hu. "Recurrent convolutional neural network for object recognition." *Proceedings of the IEEE Conference on Computer Vision and Pattern Recognition*. 2015.
[74] Alom, Md Zahangir, et al. "Inception Recurrent Convolutional Neural Network for Object Recognition." *arXiv preprint arXiv:1704.07709* (2017).
[75] Li, Yikang, et al. "ViP-CNN: Visual Phrase Guided Convolutional Neural Network."
[76] Bagherinezhad, Hessam, Mohammad Rastegari, and Ali Farhadi. "LCNN: Lookup-based Convolutional Neural Network." *arXiv preprint arXiv:1611.06473* (2016).
[77] Long, Jonathan, Evan Shelhamer, and Trevor Darrell. "Fully convolutional networks for semantic segmentation." *Proceedings of the IEEE Conference on Computer Vision and Pattern Recognition*. 2015.
[78] Bansal, Aayush, et al. "Pixelnet: Representation of the pixels, by the pixels, and for the pixels." *arXiv preprint arXiv:1702.06506*(2017).
[79] Huang, Gao, et al. "Deep networks with stochastic depth." *arXiv preprint arXiv:1603.09382* (2016).
[80] Lee, Chen-Yu, et al. "Deeply-Supervised Nets." *AISTATS*. Vol. 2. No. 3. 2015.
[81] Pezeshki, Mohammad, et al. "Deconstructing the ladder network architecture." *arXiv preprint arXiv:1511.06430* (2015).
[82] Ba, Jimmy, and Rich Caruana. "Do deep nets really need to be deep?." *Advances in neural information processing systems*. 2014.
[83] Urban, Gregor, et al. "Do deep convolutional nets really need to be deep and convolutional?." *stat* 1050 (2016): 4.
[84] Romero, Adriana, et al. "Fitnets: Hints for thin deep nets." *arXiv preprint arXiv:1412.6550* (2014).
[85] Mishkin, Dmytro, and Jiri Matas. "All you need is a good init." *arXiv preprint arXiv:1511.06422* (2015).
[86] Pandey, Gaurav, and Ambedkar Dukkipati. "To go deep or wide in learning?." *AISTATS*. 2014.
[87] Ratner, Alexander, et al. "Data Programming: Creating Large Training Sets, Quickly." *arXiv preprint arXiv:1605.07723* (2016).
[88] Aberger, Christopher R., et al. "Empty-Headed: A Relational Engine for Graph Processing." *arXiv preprint arXiv:1503.02368* (2015).
[89] Iandola, Forrest N., et al. "SqueezeNet: AlexNet-level accuracy with 50x fewer parameters and< 1MB model size." *arXiv preprint arXiv:1602.07360* (2016).
[90] Han, Song, Huizi Mao, and William J. Dally. "Deep compression: Compressing deep neural network with pruning, trained quantization and huffman coding." *CoRR, abs/1510.00149* 2 (2015).
[91] Niepert, Mathias, Mohamed Ahmed, and Konstantin Kutzkov. "Learning Convolutional Neural Networks for Graphs." *arXiv preprint arXiv:1605.05273* (2016).
[92] https://github.com/kjw0612/awesome-deep-vision
[93] Jia, Xiaoyi, et al. "Single Image Super-Resolution Using Multi-Scale Convolutional Neural Network." *arXiv preprint arXiv:1705.05084* (2017).
[94] Ahn, Byeongyong, and Nam Ik Cho. "Block-Matching Convolutional Neural Network for Image Denoising." *arXiv preprint arXiv:1704.00524* (2017).
[95] Ma, Shuang, Jing Liu, and Chang Wen Chen. "A-Lamp: Adaptive Layout-Aware Multi-Patch Deep Convolutional Neural Network for Photo Aesthetic Assessment." *arXiv preprint arXiv:1704.00248*(2017).
[96] Cao, Xiangyong, et al. "Hyperspectral Image Segmentation with Markov Random Fields and a Convolutional Neural Network." *arXiv preprint arXiv:1705.00727* (2017).
[97] de Vos, Bob D., et al. "End-to-End Unsupervised Deformable Image Registration with a Convolutional Neural Network." *arXiv preprint arXiv:1704.06065* (2017).
[98] Wang, Xin, et al. "Multimodal Transfer: A Hierarchical Deep Convolutional Neural Network for Fast Artistic Style Transfer." *arXiv preprint arXiv:1612.01895* (2016).
[99] Babaee, Mohammadreza, Duc Tung Dinh, and Gerhard Rigoll. "A deep convolutional neural network for background subtraction." *arXiv preprint arXiv:1702.01731* (2017).
[100] Hou, Jen-Cheng, et al. "Audio-Visual Speech Enhancement based on Multimodal Deep Convolutional Neural Network." *arXiv preprint arXiv:1703.10893* (2017).
[101] Xu, Yong, et al. "Convolutional gated recurrent neural network incorporating spatial features for audio tagging." *arXiv preprint arXiv:1702.07787* (2017).
[102] Litjens, Geert, et al. "A survey on deep learning in medical image analysis." *arXiv preprint arXiv:1702.05747* (2017).
[103] Zhang, Zizhao, et al. "MDNet: a semantically and visually interpretable medical image diagnosis network." *arXiv preprint arXiv:1707.02485* (2017).
[104] Tran, Phi Vu. "A fully convolutional neural network for cardiac segmentation in short-axis MRI." *arXiv preprint arXiv:1604.00494*(2016).
[105] Tan, Jen Hong, et al. "Segmentation of optic disc, fovea and retinal vasculature using a single convolutional neural network." *Journal of Computational Science* 20 (2017): 70-79.
[106] Moeskops, Pim, et al. "Automatic segmentation of MR brain images with a convolutional neural network." *IEEE transactions on medical imaging* 35.5 (2016): 1252-1261.
[107] LeCun, Y., L. Bottou, and G. Orr. "Efficient BackProp in Neural Networks: Tricks of the Trade (Orr, G. and Müller, K., eds.)." *Lecture Notes in Computer Science* 1524.
[108] Glorot, Xavier, and Yoshua Bengio. "Understanding the difficulty of training deep feedforward neural networks." International conference on artificial intelligence and statistics. 2010.
[109] He, Kaiming, et al. "Delving deep into rectifiers: Surpassing human-level performance on imagenet classification." *Proceedings of the IEEE international conference on computer vision*. 2015.
[110] Ioffe, Sergey, and Christian Szegedy. "Batch normalization: Accelerating deep network training by reducing internal covariate shift." *International Conference on Machine Learning*. 2015.
[111] Laurent, César, et al. "Batch normalized recurrent neural networks." Acoustics, Speech and Signal Processing (ICASSP), 2016 IEEE International Conference on. IEEE, 2016.
[112] Lavin, Andrew. "Fast algorithms for convolutional neural networks." *arXiv preprint arXiv , ICLR 2016*
[113] Clevert, Djork-Arné, Thomas Unterthiner, and Sepp Hochreiter. "Fast and accurate deep network learning by exponential linear units (elus)." *arXiv preprint arXiv:1511.07289* (2015).
[114] Li, Yang, et al. "Improving Deep Neural Network with Multiple Parametric Exponential Linear Units." *arXiv preprint arXiv:1606.00305* (2016).
[115] Jin, Xiaojie, et al. "Deep Learning with S-shaped Rectified Linear Activation Units." *arXiv preprint arXiv:1512.07030* (2015).
[116] Xu, Bing, et al. "Empirical evaluation of rectified activations in convolutional network." *arXiv preprint arXiv:1505.00853* (2015)
[117] He, Kaiming, et al. "Spatial pyramid pooling in deep convolutional networks for visual recognition." *European Conference on Computer Vision*. Springer, Cham, 2014.
[118] Yoo, Donggeun, et al. "Multi-scale pyramid pooling for deep convolutional representation." *Proceedings of the IEEE Conference on Computer Vision and Pattern Recognition Workshops*. 2015.
[119] Graham, Benjamin. "Fractional max-pooling." *arXiv preprint arXiv:1412.6071* (2014).
[120] Lee, Chen-Yu, Patrick W. Gallagher, and Zhuowen Tu. "Generalizing pooling functions in convolutional neural networks: Mixed, gated, and

tree." *International Conference on Artificial Intelligence and Statistics*. 2016.
[121] Hinton, Geoffrey E., et al. "Improving neural networks by preventing co-adaptation of feature detectors." *arXiv preprint arXiv:1207.0580* (2012).
[122] Srivastava, Nitish, et al. "Dropout: a simple way to prevent neural networks from overfitting." *Journal of Machine Learning Research* 15.1 (2014): 1929-1958.
[123] Wan, Li, et al. "Regularization of neural networks using dropconnect." *Proceedings of the 30th International Conference on Machine Learning (ICML-13)*. 2013.
[124] Bulò, Samuel Rota, Lorenzo Porzi, and Peter Kontschieder. "Dropout distillation." *Proceedings of The 33rd International Conference on Machine Learning*. 2016.
[125] Ruder, Sebastian. "An overview of gradient descent optimization algorithms." *arXiv preprint arXiv:1609.04747* (2016).
[126] Ngiam, Jiquan, et al. "On optimization methods for deep learning." Proceedings of the 28th International Conference on Machine Learning (ICML-11). 2011.
[127] Koushik, Jayanth, and Hiroaki Hayashi. "Improving Stochastic Gradient Descent with Feedback." *arXiv preprint arXiv:1611.01505* (2016). (ICLR-2017)
[128] Sathasivam, Saratha, and Wan Ahmad Tajuddin Wan Abdullah. "Logic learning in Hopfield networks." *arXiv preprint arXiv:0804.4075* (2008).
[129] Elman, Jeffrey L. "Finding structure in time." *Cognitive science* 14.2 (1990): 179-211.
[130] Jordan, Michael I. "Serial order: A parallel distributed processing approach." *Advances in psychology* 121 (1997): 471-495.
[131] S. Hochreiter, Y. Bengio, P. Frasconi, and J. Schmidhuber. Gradient flow in recurrent nets: the difficulty of learning long-term dependencies. In S. C. Kremer and J. F. Kolen, editors, A Field Guide to Dynamical Recurrent Neural Networks. IEEE Press, 2001.
[132] *Schmidhuber, Jürgen* . Habilitation thesis: System modeling and optimization *in 1993*. Page 150 ff demonstrates credit assignment across the equivalent of 1,200 layers in an unfolded RNN
[133] Gers, Felix A., and Jürgen Schmidhuber. "Recurrent nets that time and count." Neural Networks, 2000. IJCNN 2000, Proceedings of the IEEE-INNS-ENNS International Joint Conference on. Vol. 3. IEEE, 2000.
[134] Gers, Felix A., Nicol N. Schraudolph, and Jürgen Schmidhuber. "Learning precise timing with LSTM recurrent networks." *Journal of machine learning research* 3.Aug (2002): 115-143.
[135] Mikolov, Tomas, et al. "Recurrent neural network based language model." *Interspeech*. Vol. 2. 2010.
[136] Chung, Junyoung, et al. "Empirical evaluation of gated recurrent neural networks on sequence modeling." *arXiv preprint arXiv:1412.3555* (2014).
[137] Jozefowicz, Rafal, Wojciech Zaremba, and Ilya Sutskever. "An empirical exploration of recurrent network architectures." *Proceedings of the 32nd International Conference on Machine Learning (ICML-15)*. 2015.
[138] Yao, Kaisheng, et al. "Depth-gated LSTM." *arXiv preprint arXiv:1508.03790*(2015).
[139] Koutnik, Jan, et al. "A clockwork rnn." *International Conference on Machine Learning*. 2014.
[140] Greff, Klaus, et al. "LSTM: A search space odyssey." *IEEE transactions on neural networks and learning systems* (2016).
[141] Karpathy, Andrej, and Li Fei-Fei. "Deep visual-semantic alignments for generating image descriptions." *Proceedings of the IEEE Conference on Computer Vision and Pattern Recognition*. 2015.
[142] Xingjian, S. H. I., et al. "Convolutional LSTM network: A machine learning approach for precipitation nowcasting." Advances in neural information processing systems. 2015.
[143] Mikolov, Tomas, et al. "Efficient estimation of word representations in vector space." *arXiv preprint arXiv:1301.3781* (2013).
[144] Goldberg, Yoav, and Omer Levy. "word2vec Explained: deriving Mikolov et al.'s negative-sampling word-embedding method." *arXiv preprint arXiv:1402.3722* (2014).
[145] Xu, Kelvin, et al. "Show, attend and tell: Neural image caption generation with visual a attention." *International Conference on Machine Learning*. 2015.
[146] Qin, Yao, et al. "A Dual-Stage Attention-Based Recurrent Neural Network for Time Series Prediction." *arXiv preprint arXiv:1704.02971* (2017).
[147] Xiong, Caiming, Stephen Merity, and Richard Socher. "Dynamic memory networks for visual and textual question answering." *International Conference on Machine Learning*. 2016.
[148] Oord, Aaron van den, Nal Kalchbrenner, and Koray Kavukcuoglu. "Pixel recurrent neural networks." *arXiv preprint arXiv:1601.06759*(2016).
[149] Xue, Wufeng, et al. "Direct Estimation of Regional Wall Thicknesses via Residual Recurrent Neural Network." *International Conference on Information Processing in Medical Imaging*. Springer, Cham, 2017.
[150] Tjandra, Andros, et al. "Gated Recurrent Neural Tensor Network." *Neural Networks (IJCNN), 2016 International Joint Conference on*. IEEE, 2016.
[151] Wang, Shuohang, and Jing Jiang. "Learning natural language inference with LSTM." *arXiv preprint arXiv:1512.08849* (2015).
[152] Sutskever, Ilya, Oriol Vinyals, and Quoc VV Le. "Sequence to sequence learning with neural networks." Advances in Neural Information Processing Systems. 2014.
[153] Lakhani, Vrishabh Ajay, and Rohan Mahadev. "Multi-Language Identification Using Convolutional Recurrent Neural Network." *arXiv preprint arXiv:1611.04010* (2016).
[154] Längkvist, Martin, Lars Karlsson, and Amy Loutfi. "A review of unsupervised feature learning and deep learning for time-series modeling." *Pattern Recognition Letters* 42 (2014): 11-24.
[155] Malhotra, Pankaj, et al. "TimeNet: Pre-trained deep recurrent neural network for time series classification." *arXiv preprint arXiv:1706.08838* (2017).
[156] Soltau, Hagen, Hank Liao, and Hasim Sak. "Neural speech recognizer: Acoustic-to-word LSTM model for large vocabulary speech recognition." *arXiv preprint arXiv:1610.09975* (2016).
[157] Sak, Haşim, Andrew Senior, and Françoise Beaufays. "Long short-term memory recurrent neural network architectures for large scale acoustic modeling." *Fifteenth Annual Conference of the International Speech Communication Association*. 2014.
[158] Adavanne, Sharath, Pasi Pertilä, and Tuomas Virtanen. "Sound event detection using spatial features and convolutional recurrent neural network." *arXiv preprint arXiv:1706.02291* (2017).
[159] Chien, Jen-Tzung, and Alim Misbullah. "Deep long short-term memory networks for speech recognition." *Chinese Spoken Language Processing (ISCSLP), 2016 10th International Symposium on*. IEEE, 2016.
[160] Choi, Edward, et al. "Using recurrent neural network models for early detection of heart failure onset." *Journal of the American Medical Informatics Association* 24.2 (2016): 361-370.
[161] Li, Yaguang, et al. "Graph Convolutional Recurrent Neural Network: Data-Driven Traffic Forecasting." *arXiv preprint arXiv:1707.01926* (2017).
[162] Azzouni, Abdelhadi, and Guy Pujolle. "A Long Short-Term Memory Recurrent Neural Network Framework for Network Traffic Matrix Prediction." *arXiv preprint arXiv:1705.05690* (2017).
[163] Olabiyi, Oluwatobi, et al. "Driver Action Prediction Using Deep (Bidirectional) Recurrent Neural Network." *arXiv preprint arXiv:1706.02257* (2017).
[164] Kim, ByeoungDo, et al. "Probabilistic Vehicle Trajectory Prediction over Occupancy Grid Map via Recurrent Neural Network." *arXiv preprint arXiv:1704.07049* (2017).
[165] Richard, Alexander, and Juergen Gall. "A bag-of-words equivalent recurrent neural network for action recognition." *Computer Vision and Image Understanding* 156 (2017): 79-91.
[166] Bontemps, Loïc, James McDermott, and Nhien-An Le-Khac. "Collective Anomaly Detection Based on Long Short-Term Memory Recurrent Neural Networks." *International Conference on Future Data and Security Engineering*. Springer International Publishing, 2016.
[167] Kingma, Diederik P., and Max Welling. "Stochastic gradient VB and the variational auto-encoder." *Second International Conference on Learning Representations, ICLR*. 2014.
[168] Ng, Andrew. "Sparse autoencoder." *CS294A Lecture notes*72.2011 (2011): 1-19.
[169] Vincent, Pascal, et al. "Stacked denoising autoencoders: Learning useful representations in a deep network with a local denoising criterion." *Journal of Machine Learning Research* 11.Dec (2010): 3371-3408.
[170] Zhang, Richard, Phillip Isola, and Alexei A. Efros. "Split-brain autoencoders: Unsupervised learning by cross-channel prediction." *arXiv preprint arXiv:1611.09842* (2016).
[171] *Chicco, Davide; Sadowski, Peter; Baldi, Pierre (1 January 2014).* "Deep Autoencoder Neural Networks for Gene Ontology Annotation Predictions". *Proceedings of the 5th ACM Conference*

> REPLACE THIS LINE WITH YOUR PAPER IDENTIFICATION NUMBER (DOUBLE-CLICK HERE TO EDIT) <    37on Bioinformatics, Computational Biology, and Health Informatics - BCB '14. ACM: 533–540.
[172] Alom, Md Zahangir and Tarek M. Taha. " Network Intrusion Detection for Cyber Security using Unsupervised Deep Learning Approaches " *Aerospace and Electronics Conference (NAECON), National*. IEEE, 2017.
[173] Song, Chunfeng, et al. "Auto-encoder based data clustering." *Iberoamerican Congress on Pattern Recognition*. Springer Berlin Heidelberg, 2013.
[174] Lu, Jiajun, Aditya Deshpande, and David Forsyth. "CDVAE: Co-embedding Deep Variational Auto Encoder for Conditional Variational Generation." *arXiv preprint arXiv:1612.00132* (2016).
[175] Ahmad, Muhammad, Stanislav Protasov, and Adil Mehmood Khan. "Hyperspectral Band Selection Using Unsupervised Non-Linear Deep Auto Encoder to Train External Classifiers." *arXiv preprint arXiv:1705.06920* (2017).
[176] Freund, Yoav, and David Haussler. "Unsupervised learning of distributions of binary vectors using two layer networks." (1994).
[177] Larochelle, Hugo, and Yoshua Bengio. "Classification using discriminative restricted Boltzmann machines." *Proceedings of the 25th international conference on Machine learning*. ACM, 2008.
[178] R. Salakhutdinov and G. E. Hinton. Deep Boltzmann machines. In AISTATS, volume 1, page 3, 2009.
[179] Alom, Md Zahangir, VenkataRamesh Bontupalli, and Tarek M. Taha. "Intrusion detection using deep belief networks." *Aerospace and Electronics Conference (NAECON), 2015 National*. IEEE, 2015.
[180] Goodfellow, Ian, et al. "Generative adversarial nets." *Advances in neural information processing systems*. 2014.
[181] T. Salimans, I. Goodfellow, W. Zaremba, V. Che- ung, A. Radford, and X. Chen. Improved techniques for training gans. *arXiv preprint arXiv:1606.03498*, 2016.
[182] Vondrick, Carl, Hamed Pirsiavash, and Antonio Torralba. "Generating videos with scene dynamics." *Advances In Neural Information Processing Systems*. 2016.
[183] Radford, Alec, Luke Metz, and Soumith Chintala. "Unsupervised representation learning with deep convolutional generative adversarial networks." *arXiv preprint arXiv:1511.06434* (2015).
[184] X. Wang and A. Gupta. Generative image modeling using style and structure adversarial networks. In *Proc. ECCV*, 2016.
[185] Chen, Xi, et al. "InfoGAN: Interpretable representation learning by information maximizing generative adversarial nets." *Advances in Neural Information Processing Systems*. 2016.
[186] D. J. Im, C. D. Kim, H. Jiang, and R. Memisevic. Generating images with recurrent adversarial net- works. *http://arxiv.org/abs/ 1602.05110*, 2016.
[187] Isola, Phillip, et al. "Image-to-image translation with conditional adversarial networks." *arXiv preprint* (2017).
[188] Liu, Ming-Yu, and Oncel Tuzel. "Coupled generative adversarial networks." *Advances in neural information processing systems*. 2016.
[189] Donahue, Jeff, Philipp Krähenbühl, and Trevor Darrell. "Adversarial feature learning." *arXiv preprint arXiv:1605.09782* (2016).
[190] Berthelot, David, Tom Schumm, and Luke Metz. "Began: Boundary equilibrium generative adversarial networks." *arXiv preprint arXiv:1703.10717*(2017).
[191] Martin Arjovsky, Soumith Chintala, and Léon Bottou. Wasserstein gan. *arXiv preprint arXiv:1701.07875*, 2017.
[192] Gulrajani, Ishaan, et al. "Improved training of wasserstein gans." *arXiv preprint arXiv:1704.00028* (2017).
[193] He, Kun, Yan Wang, and John Hopcroft. "A powerful generative model using random weights for the deep image representation." *Advances in Neural Information Processing Systems*. 2016.
[194] Kos, Jernej, Ian Fischer, and Dawn Song. "Adversarial examples for generative models." *arXiv preprint arXiv:1702.06832* (2017).
[195] Zhao, Junbo, Michael Mathieu, and Yann LeCun. "Energy-based generative adversarial network." *arXiv preprint arXiv:1609.03126* (2016).
[196] Park, Noseong, et al. "MMGAN: Manifold Matching Generative Adversarial Network for Generating Images." *arXiv preprint arXiv:1707.08273* (2017).
[197] Laloy, Eric, et al. "Efficient training-image based geostatistical simulation and inversion using a spatial generative adversarial neural network." *arXiv preprint arXiv:1708.04975* (2017).
[198] Eghbal-zadeh, Hamid, and Gerhard Widmer. "Probabilistic Generative Adversarial Networks." *arXiv preprint arXiv:1708.01886* (2017).
[199] Fowkes, Jaroslav, and Charles Sutton. "A Bayesian Network Model for Interesting Itemsets." *Joint European Conference on Machine Learning and Knowledge Disco in Databases*. Springer International Publishing, 2016.
[200] Mescheder, Lars, Sebastian Nowozin, and Andreas Geiger. "Adversarial variational bayes: Unifying variational autoencoders and generative adversarial networks." *arXiv preprint arXiv:1701.04722* (2017).
[201] Nowozin, Sebastian, Botond Cseke, and Ryota Tomioka. "f-gan: Training generative neural samplers using variational divergence minimization." *Advances in Neural Information Processing Systems*. 2016.
[202] Li, Chuan, and Michael Wand. "Precomputed real-time texture synthesis with markovian generative adversarial networks." *European Conference on Computer Vision*. Springer International Publishing, 2016.
[203] Du, Chao, Jun Zhu, and Bo Zhang. "Learning Deep Generative Models with Doubly Stochastic Gradient MCMC." *IEEE Transactions on Neural Networks and Learning Systems* (2017).
[204] Hoang, Quan, et al. "Multi-Generator Gernerative Adversarial Nets." *arXiv preprint arXiv:1708.02556* (2017).
[205] Bousmalis, Konstantinos, et al. "Unsupervised pixel-level domain adaptation with generative adversarial networks." *arXiv preprint arXiv:1612.05424* (2016).
[206] Kansky, Ken, et al. "Schema Networks: Zero-shot Transfer with a Generative Causal Model of Intuitive Physics." *arXiv preprint arXiv:1706.04317* (2017).
[207] Ledig, Christian, et al. "Photo-realistic single image super-resolution using a generative adversarial network." *arXiv preprint arXiv:1609.04802* (2016).
[208] Souly, Nasim, Concetto Spampinato, and Mubarak Shah. "Semi and Weakly Supervised Semantic Segmentation Using Generative Adversarial Network." *arXiv preprint arXiv:1703.09695* (2017).
[209] Dash, Ayushman, et al. "TAC-GAN-Text Conditioned Auxiliary Classifier Generative Adversarial Network." *arXiv preprint arXiv:1703.06412* (2017).
[210] Zhang, Hang, and Kristin Dana. "Multi-style Generative Network for Real-time Transfer." *arXiv preprint arXiv:1703.06953* (2017).
[211] Zhang, He, Vishwanath Sindagi, and Vishal M. Patel. "Image De-raining Using a Conditional Generative Adversarial Network." *arXiv preprint arXiv:1701.05957* (2017).
[212] Serban, Iulian Vlad, et al. "Building End-To-End Dialogue Systems Using Generative Hierarchical Neural Network Models." *AAAI*. 2016.
[213] Pascual, Santiago, Antonio Bonafonte, and Joan Serrà. "SEGAN: Speech Enhancement Generative Adversarial Network." *arXiv preprint arXiv:1703.09452* (2017).
[214] Yang, Li-Chia, Szu-Yu Chou, and Yi-Hsuan Yang. "MidiNet: A convolutional generative adversarial network for symbolic-domain music generation." *Proceedings of the 18th International Society for Music Information Retrieval Conference (ISMIR'2017), Suzhou, China*. 2017.
[215] Yang, Qingsong, et al. "Low Dose CT Image Denoising Using a Generative Adversarial Network with Wasserstein Distance and Perceptual Loss." *arXiv preprint arXiv:1708.00961* (2017).
[216] Rezaei, Mina, et al. "Conditional Adversarial Network for Semantic Segmentation of Brain Tumor." *arXiv preprint arXiv:1708.05227*(2017)
[217] Xue, Yuan, et al. "SegAN: Adversarial Network with Multi-scale $L\_1$ Loss for Medical Image Segmentation." *arXiv preprint arXiv:1706.01805* (2017).
[218] Mardani, Morteza, et al. "Deep Generative Adversarial Networks for Compressed Sensing Automates MRI." *arXiv preprint arXiv:1706.00051* (2017).
[219] Choi, Edward, et al. "Generating Multi-label Discrete Electronic Health Records using Generative Adversarial Networks." *arXiv preprint arXiv:1703.06490* (2017).
[220] Esteban, Cristóbal, Stephanie L. Hyland, and Gunnar Rätsch. "Real-valued (Medical) Time Series Generation with Recurrent Conditional GANs." *arXiv preprint arXiv:1706.02633* (2017).
[221] Hayes, Jamie, et al. "LOGAN: Evaluating Privacy Leakage of Generative Models Using Generative Adversarial Networks." *arXiv preprint arXiv:1705.07663* (2017).
[222] Gordon, Jonathan, and José Miguel Hernández-Lobato. "Bayesian Semisupervised Learning with Deep Generative Models." *arXiv preprint arXiv:1706.09751* (2017).
[223] Abbasnejad, M. Ehsan, et al. "Bayesian Conditional Generative Adverserial Networks." *arXiv preprint arXiv:1706.05477* (2017).
[224] Grnarova, Paulina, et al. "An Online Learning Approach to Generative Adversarial Networks." *arXiv preprint arXiv:1706.03269* (2017).

[225] Li, Yujia, Kevin Swersky, and Rich Zemel. "Generative moment matching networks." *Proceedings of the 32nd International Conference on Machine Learning (ICML-15)*. 2015.
[226] Li, Chun-Liang, et al. "MMD GAN: Towards Deeper Understanding of Moment Matching Network." *arXiv preprint arXiv:1705.08584* (2017).
[227] Nie, Xuecheng, et al. "Generative Partition Networks for Multi-Person Pose Estimation." *arXiv preprint arXiv:1705.07422* (2017).
[228] Saeedi, Ardavan, et al. "Multimodal Prediction and Personalization of Photo Edits with Deep Generative Models." *arXiv preprint arXiv:1704.04997* (2017).
[229] Schlegl, Thomas, et al. "Unsupervised Anomaly Detection with Generative Adversarial Networks to Guide Marker Disco ." *International Conference on Information Processing in Medical Imaging*. Springer, Cham, 2017.
[230] Kim, Taeksoo, et al. "Learning to discover cross-domain relations with generative adversarial networks." *arXiv preprint arXiv:1703.05192* (2017).
[231] Mehrotra, Akshay, and Ambedkar Dukkipati. "Generative Adversarial Residual Pairwise Networks for One Shot Learning." *arXiv preprint arXiv:1703.08033* (2017).
[232] Sordoni, Alessandro, et al. "A neural network approach to context-sensitive generation of conversational responses." *arXiv preprint arXiv:1506.06714* (2015).
[233] Yin, Jun, et al. "Neural generative question answering." *arXiv preprint arXiv:1512.01337* (2015).
[234] Li, Yuxi. "Deep reinforcement learning: An overview." *arXiv preprint arXiv:1701.07274* (2017).
[235] Goodfellow, Ian, Yoshua Bengio, and Aaron Courville. *Deep learning*. MIT press, 2016.
[236] David Silver, Aja Huang, Chris J Maddison, Arthur Guez, Laurent Sifre, George Van Den Driessche, Julian Schrittwieser, Ioannis Antonoglou, Veda Panneershelvam, Marc Lanc- tot, et al. Mastering the game of Go with deep neural networks and tree search. *Nature*, 529(7587):484–489, 2016.
[237] Vinyals, Oriol, et al. "StarCraft II: A New Challenge for Reinforcement Learning." arXiv preprint arXiv:1708.04782 (2017).
[238] Koenig, Sven, and Reid G. Simmons. *Complexity analysis of real-time reinforcement learning applied to finding shortest paths in deterministic domains*. No. CMU-CS-93-106. CARNEGIE-MELLON UNIV PITTSBURGH PA SCHOOL OF COMPUTER SCIENCE, 1992.
[239] Schulman, John, et al. "Trust region policy optimization." Proceedings of the 32nd International Conference on Machine Learning (ICML-15). 2015.
[240] Levine, Sergey, et al. "End-to-end training of deep visuomotor policies." *Journal of Machine Learning Research* 17.39 (2016): 1-40.
[241] Mnih, Volodymyr, et al. "Asynchronous methods for deep reinforcement learning." *International Conference on Machine Learning*. 2016.
[242] Kober, Jens, J. Andrew Bagnell, and Jan Peters. "Reinforcement learning in robotics: A survey." *The International Journal of Robotics Research* 32.11 (2013): 1238-1274.
[243] Arulkumaran, Kai, et al. "A brief survey of deep reinforcement learning." *arXiv preprint arXiv:1708.05866* (2017).
[244] Zhu, Feiyun, et al. "Cohesion-based Online Actor-Critic Reinforcement Learning for mHealth Intervention." *arXiv preprint arXiv:1703.10039* (2017).
[245] Zhu, Feiyun, et al. "Group-driven Reinforcement Learning for Personalized mHealth Intervention." *arXiv preprint arXiv:1708.04001* (2017).
[246] Steckelmacher, Denis, et al. "Reinforcement Learning in POMDPs with Memoryless Options and Option-Observation Initiation Sets." *arXiv preprint arXiv:1708.06551* (2017).
[247] Hu, Haoyuan, et al. "Solving a new 3d bin packing problem with deep reinforcement learning method." *arXiv preprint arXiv:1708.05930* (2017).
[248] Everitt, Tom, et al. "Reinforcement Learning with a Corrupted Reward Channel." *arXiv preprint arXiv:1705.08417* (2017).
[249] Wu, Yuhuai, et al. "Scalable trust-region method for deep reinforcement learning using Kronecker-factored approximation." *arXiv preprint arXiv:1708.05144* (2017).
[250] Denil, Misha, et al. "Learning to perform physics experiments via deep reinforcement learning." *arXiv preprint arXiv:1611.01843* (2016).
[251] Hein, Daniel, et al. "Particle swarm optimization for generating interpretable fuzzy reinforcement learning policies." *Engineering Applications of Artificial Intelligence* 65 (2017): 87-98.
[252] Islam, Riashat, et al. "Reproducibility of Benchmarked Deep Reinforcement Learning Tasks for Continuous Control." *arXiv preprint arXiv:1708.04133* (2017).
[253] Inoue, Tadanobu, et al. "Deep reinforcement learning for high precision assembly tasks." *arXiv preprint arXiv:1708.04033* (2017).
[254] Li, Kun, and Joel W. Burdick. "Inverse Reinforcement Learning in Large State Spaces via Function Approximation." *arXiv preprint arXiv:1707.09394* (2017).
[255] Liu, Ning, et al. "A Hierarchical Framework of Cloud Resource Allocation and Power Management Using Deep Reinforcement Learning." *Distributed Computing Systems (ICDCS), 2017 IEEE 37th International Conference on*. IEEE, 2017.
[256] Cao, Qingxing, et al. "Attention-aware face hallucination via deep reinforcement learning." *arXiv preprint arXiv:1708.03132* (2017).
[257] Chen, Tianqi, Ian Goodfellow, and Jonathon Shlens. "Net2net: Accelerating learning via knowledge transfer." *arXiv preprint arXiv:1511.05641* (2015).
[258] Ganin, Yaroslav, and Victor Lempitsky. "Unsupervised domain adaptation by backpropagation." *arXiv preprint arXiv:1409.7495* (2014).
[259] Ganin, Yaroslav, et al. "Domain-adversarial training of neural networks." *Journal of Machine Learning Research* 17.59 (2016): 1-35.
[260] Pan, Sinno Jialin, and Qiang Yang. "A survey on transfer learning." *IEEE Transactions on knowledge and data engineering* 22.10 (2010): 1345-1359.
[261] McKeough, Anne. Teaching for transfer: Fostering generalization in learning. Routledge, 2013.
[262] Raina, Rajat, et al. "Self-taught learning: transfer learning from unlabeled data." *Proceedings of the 24th international conference on Machine learning*. ACM, 2007
[263] Dai, Wenyuan, et al. "Boosting for transfer learning." *Proceedings of the 24th international conference on Machine learning*. ACM, 2007.
[264] Han, Song, Huizi Mao, and William J. Dally. "Deep compression: Compressing deep neural networks with pruning, trained quantization and huffman coding." *arXiv preprint arXiv:1510.00149* (2015).
[265] Qiu, Jiantao, et al. "Going deeper with embedded FPGA platform for convolutional neural network." *Proceedings of the 2016 ACM/SIGDA International Symposium on Field-Programmable Gate Arrays*. ACM, 2016.
[266] He, Kaiming, and Jian Sun. "Convolutional neural networks at constrained time cost." *Proceedings of the IEEE Conference on Computer Vision and Pattern Recognition*. 2015.
[267] 13. Lin, Zhouhan, et al. "Neural networks with few multiplications." *arXiv preprint arXiv:1510.03009* (2015).
[268] 14. Courbariaux, Matthieu, Jean-Pierre David, and Yoshua Bengio. "Training deep neural networks with low precision multiplications." *arXiv preprint arXiv:1412.7024* (2014).
[269] Courbariaux, Matthieu, Yoshua Bengio, and Jean-Pierre David. "Binaryconnect: Training deep neural networks with binary weights during propagations." *Advances in Neural Information Processing Systems*. 2015.
[270] Hubara, Itay, Daniel Soudry, and Ran El Yaniv. "Binarized Neural Networks." *arXiv preprint arXiv:1602.02505* (2016).
[271] Kim, Minje, and Paris Smaragdis. "Bitwise neural networks." *arXiv preprint arXiv:1601.06071* (2016).
[272] Dettmers, Tim. "8-Bit Approximations for Parallelism in Deep Learning." *arXiv preprint arXiv:1511.04561* (2015).
[273] Gupta, Suyog, et al. "Deep learning with limited numerical precision." *CoRR, abs/1502.02551* 392 (2015).
[274] Rastegari, Mohammad, et al. "XNOR-Net: ImageNet Classification Using Binary Convolutional Neural Networks." *arXiv preprint arXiv:1603.05279* (2016).
[275] Merolla, Paul A., et al. "A million spiking-neuron integrated circuit with a scalable communication network and interface." *Science* 345.6197 (2014): 668-673.
[276] Esser, Steven K., et al. "Convolutional networks for fast, energy-efficient neuromorphic computing "Proceedings of the National Academy of Science (2016): 201604850.
[277] Schuman, Catherine D., et al. "A Survey of Neuromorphic Computing and Neural Networks in Hardware." *arXiv preprint arXiv:1705.06963* (2017).
[278] Chen, Yu-Hsin, et al. "Eyeriss: An energy-efficient reconfigurable accelerator for deep convolutional neural networks." *IEEE Journal of Solid-State Circuits* 52.1 (2017): 127-138.
[279] Chen, Yunji, et al. "Dadiannao: A machine-learning supercomputer." *Proceedings of the 47th Annual IEEE/ACM*

*International Symposium on Microarchitecture*. IEEE Computer Society, 2014.

[280] Jouppi, Norman P., et al. "In-datacenter performance analysis of a tensor processing unit." *arXiv preprint arXiv:1704.04760* (2017).

[281] Han, Song, et al. "EIE: efficient inference engine on compressed deep neural network." *Proceedings of the 43rd International Symposium on Computer Architecture*. IEEE Press, 2016.

[282] Zhang, Xiangyu, et al. "Efficient and accurate approximations of nonlinear convolutional networks." *Proceedings of the IEEE Conference on Computer Vision and Pattern Recognition*. 2015.

[283] Novikov, Alexander, et al. "Tensorizing neural networks." *Advances in Neural Information Processing Systems*. 2015.

[284] Zhu, Chenzhuo, et al. "Trained ternary quantization." *arXiv preprint arXiv:1612.01064* (2016).

[285] Russakovsky, Olga, et al. "Imagenet large scale visual recognition challenge." *International Journal of Computer Vision* 115.3 (2015): 211-252.

[286] Oord, Aaron van den, et al. "Wavenet: A generative model for raw audio." *arXiv preprint arXiv:1609.03499* (2016).

[287] Zhang, Xingcheng, et al. "Polynet: A pursuit of structural diversity in deep networks." *2017 IEEE Conference on Computer Vision and Pattern Recognition (CVPR)*. IEEE, 2017.

[288] Kunihiko Fukushima, "Neural network model for selective attention in visual pattern recognition and associative recall," Appl. Opt. **26**, 4985-4992 (1987)

[289] Alom, Md Zahangir, et al. "Handwritten Bangla Digit Recognition Using Deep Learning." *arXiv preprint arXiv:1705.02680* (2017)

[290] Alom, Md Zahangir, et al. "Improved Inception-Residual Convolutional Neural Network for Object Recognition." *arXiv preprint arXiv:1712.09888* (2017).

[291] Alom, Md Zahangir, et al. "Handwritten Bangla Character Recognition Using The State-of-Art Deep Convolutional Neural Networks." *arXiv preprint arXiv:1712.09872* (2017).

[292] Socher, Richard, et al. "Parsing natural scenes and natural language with recursive neural networks." *Proceedings of the 28th international conference on machine learning (ICML-11)*. 2011.

[293] Sabour, Sara, Nicholas Frosst, and Geoffrey E. Hinton. "Dynamic routing between capsules." *Advances in Neural Information Processing Systems*. 2017.

[294] Sze, Vivienne, et al. "Efficient processing of deep neural networks: A tutorial and survey." *Proceedings of the IEEE* 105.12 (2017): 2295-2329.

[295] Rawat, Waseem, and Zenghui Wang. "Deep convolutional neural networks for image classification: A comprehensive review." *Neural computation* 29.9 (2017): 2352-2449.

[296] Alom, Md Zahangir, et al. "Optical beam classification using deep learning: a comparison with rule-and feature-based classification." *Optics and Photonics for Information Processing XI*. Vol. 10395. International Society for Optics and Photonics, 2017.

[297] Alom, Md Zahangir, et al. "Object recognition using cellular simultaneous recurrent networks and convolutional neural network." *Neural Networks (IJCNN), 2017 International Joint Conference on*. IEEE, 2017.